\documentclass{article}
\usepackage[utf8]{inputenc}
\usepackage{iclr2027_conference,times}
\iclrfinalcopy
\usepackage{amsmath,amsfonts,amssymb,amsthm,xcolor,url,booktabs,graphicx,enumitem}
\newtheorem{proposition}{Proposition}
\usepackage[colorlinks=true,allcolors=blue]{hyperref}
\usepackage{natbib}
\newcommand{\Hone}{\ensuremath{H_{1}}}
\newcommand{\Htwo}{\ensuremath{H_{2}}}
\newcommand{\Hthree}{\ensuremath{H_{3}}}

\title{SFT Leaves a Reproducible Signature in\\Activation Geometry: Calibrating\\Post-Training Interpretability Against\\Run-to-Run Noise}
\author{Xupeng Chen$^{1,*}$ \quad Chenqian Le$^{1}$ \quad Jiayi Cheng$^{1}$ \quad Runhao Li$^{2}$ \quad Michael Yue$^{3}$ \\
{\normalfont $^{1}$New York University \quad $^{2}$University of Southern California \quad $^{3}$Columbia University} \\
{\normalfont $^{*}$Corresponding author: \texttt{xc1490@nyu.edu}}}
\date{}

\usepackage{longtable,array}
\newcommand{\coderepo}{\url{https://github.com/james20141606/sft-signature-activation-geometry}}
\begin{document}
\maketitle
\lhead{Preprint}

\begin{abstract}
A model diff credits post-training with whatever separates a base checkpoint
from one post-trained descendant, presuming a rerun would land close by.
On the sparse co-activation graph of \texttt{Qwen3-1.7B} it does not: two
LoRA-SFT reruns sit over half as far apart as either sits from the base. We score each base-to-post contrast against its
own recipe's run-to-run band, and show that $k$
runs certify no better than $p\le1/(k+1)$ under run exchangeability, however large the effect, so
the band's $z$ is a descriptive scale. SFT leaves a large, reproducible
signature: base$\to$SFT \emph{displaces} the graph by $17.8$ pairwise-distance
SDs beyond the mean distance between four independent LoRA-SFT runs, and the pre-registered
estimator replicates on Llama-3.1-8B with Llama-Scope. At $29$
steps each, four SFT and four DPO runs occupy \emph{disjoint}
displacement ranges, a gap $5.0\times$ and $5.6\times$ the two arms' spreads, without
any band. Every band is scored with base and run graphs built on one GPU
architecture, because graphs are not bit-identical across architectures.
Three other readings stop short of a verdict: DPO's band counts ($3/4$, $1/4$,
$3/4$ contrasts outside at $12$, $29$, $60$ steps) show no consistent dose
ordering while its displacement stays flat; light GRPO is a secondary
non-detection, not an absence, and we run no equivalence test; and the graph
concentration read as an RL signature is already in the zero-RL base, and a
dictionary-width control removes an apparent scaling law in it. A pre-registered causal sweep at three edge budgets on $4{,}816$ held-out prompts found
the displaced sources load-bearing ($12\times$ a matched random control in mean, $3.5\times$ in median at $10{,}000$ edges), but
their stage-specificity interaction reversed at two budgets and the third failed its faithfulness gate,
so we claim calibrated structural displacement, not a changed mechanism.
Model diffs should be read beside their recipe's rerun band, within-band diffs
as non-detections.
\textbf{Scope:} the full detection protocol covers one backbone
(\texttt{Qwen3-1.7B}), one SFT recipe and general-text probe states; the
matched-base control replicates across three families.
\end{abstract}

\section{Introduction}
\label{sec:intro}
Model diffing reads what post-training did off one base checkpoint and one post-trained descendant, such as Gemma-2-2B and its chat model~\citep{minder2025crosscoders} or Qwen2.5-Math-7B and its reasoning-distilled variant~\citep{transcoder_diffing2026}; each is one run of its recipe, so the reading presumes a second run would land close to it. Diff the sparse co-activation graph of \texttt{Qwen3-1.7B} before and after $29$ steps of LoRA-SFT and the graph moves: across five SFT runs (the audited one and four reseeded reruns), its $10{,}000$ strongest edges shift by an edge-Jaccard distance of $0.057$--$0.061$. With only the seed changed, two SFT runs already sit $0.034$--$0.038$ apart, more than half the base-to-SFT distance. LoRA-SFT's diff exceeds every rerun distance by a wide margin, on two backbones; DPO's diff sits at the edge of its rerun band at every step count, so its band verdict follows no dose ordering; and light GRPO is not resolved from its reruns.

Interpretability work reads post-training off exactly this kind of diff: a statistic on the post-trained checkpoint, sometimes differenced against its base, is reported as what SFT, preference optimization or RL did to the model~\citep{rl_circuit_intensity2026,transcoder_diffing2026,shi2026rl_generalize}, often against the expectation that post-training reweights what the base already computes~\citep{yue2025,wang2025,lin2024} while policy entropy collapses~\citep{cui2025entropy_mechanism}. The feature-level ``narrowing'' prediction we audit is our own extrapolation of that picture to sparse features; those papers make behavioral claims. Two comparisons are usually missing: the same statistic on the base, and the spread between runs of the same recipe. Without them, base properties and run-to-run variation are reported as effects of training; with both in place, an RL signature in graph concentration turns out to be a base property and a scaling law in it a dictionary-width artifact.

\begin{figure}[t]
\centering
\includegraphics[width=\textwidth]{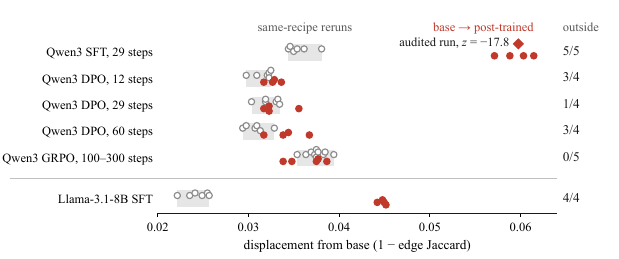}
\caption{\textbf{LoRA-SFT moves the co-activation graph further from its base than its own reruns sit apart, on two backbones; DPO's contrasts sit at their band's edge, light GRPO's inside it.} Rows are the arms of Table~\ref{tab:main}; shading spans the seed-pair distances (open circles; a range, not an interval); filled points are base-to-post contrasts (diamond: audited SFT checkpoint); counts are contrasts beyond every seed pair. DPO's band counts ($3/4$, $1/4$, $3/4$ contrasts outside at $12/29/60$ steps) follow no dose ordering. All graphs in a row were built on one GPU architecture (H200).}
\label{fig:bands}
\end{figure}

\paragraph{We build the missing comparisons.}
A \emph{matched-base control} credits post-training only with the change in a statistic from the base, computed with the same SAE and probes. A \emph{seed-noise null} scores each base-to-post displacement against the spread of independently seeded runs of the recipe that produced it, with the same estimator, so both are measured alike (Fig.~\ref{fig:bands}). A \emph{positive control} (the instrument must detect some stage, so that its non-detections mean something), a random-dictionary control and a precision rule complete the audit (Secs.~\ref{sec:method}--\ref{sec:results-rewiring}). The seeded LoRA runs are re-seeded before model construction, so their bands include adapter initialization and not data order alone; the full-parameter GRPO runs share their starting weights.

\paragraph{What survives, and what does not.}
SFT's signature survives: LoRA-SFT displaces the co-activation graph far outside its own four-run band ($z=-17.8$, a descriptive scale) and again on Llama-3.1-8B with Llama-Scope, and against the GRPO-family band it is detected under a random dictionary as under the trained SAE. At the identical $29$ optimizer steps, four SFT and four DPO runs occupy disjoint displacement ranges, a comparison that needs no band. Three readings do not survive as verdicts: DPO's contrasts sit at the edge of their band at every dose while its displacement stays flat, so its band counts follow no dose ordering (Table~\ref{tab:main}) and we report them per dose; the concentration that single-checkpoint studies read as an RL signature is already far above a permutation null in the zero-RL base, with the GRPO graph at the same level; and light GRPO is not resolved from its run-noise band, a non-detection and not an equivalence result, although it did learn its task (accuracy gain on $100$ held-out items, Sec.~\ref{sec:results-rewiring}). We draw no behavioral/internal dissociation from the cross-state comparison. Nor do we read displacement as mechanism: a registered causal sweep found the stage-specificity interaction reversed at two of three budgets.

\paragraph{Contributions.}
\begin{itemize}[leftmargin=*,itemsep=1pt,topsep=2pt]
\item \textbf{A rerun-band protocol for model diffing}: each base-to-post displacement is scored against same-recipe reruns with the same estimator, single-checkpoint statistics are credited only with their change from the base, and a precision rule fixes what can be scored (Sec.~\ref{sec:method}); the supplement recomputes the headline statistics without a GPU. The permutation rank floor applies: $k$ runs certify no better than $p\le1/(k+1)$ under run exchangeability, whatever the number of seed pairs (Proposition~\ref{prop:band}), so $z$ is a descriptive scale.
\item \textbf{A reproducible SFT signature}: base$\to$SFT displaces the co-activation graph beyond its own run-to-run band on two backbones and two SAE families, is disjoint from DPO at an equal step count without any band, and its dropped edges concentrate on boilerplate-to-hub edges (Sec.~\ref{sec:results-rewiring}).
\item \textbf{Four ways a signature reading fails, each with the control that catches it}: base-set concentration, a TopK-pinned sparsity metric, a width-driven ``scaling law'', and a DPO band verdict that changes across doses, and with the GPU that built the graphs, while displacement stays flat (Secs.~\ref{sec:results-h1}--\ref{sec:results-rewiring}).
\end{itemize}

\section{Related Work}
\label{sec:related}
\paragraph{What post-training changes inside a model.} The comparison our band calibrates, one base against one post-trained run, is how this literature measures fine-tuning. Mechanistic studies mostly find reuse of circuitry the pretrained model already has. \citet{jain2024mechanistic} describe fine-tuning on procedurally defined tasks as learning a thin wrapper over existing capabilities, \citet{prakash2024entity} find entity tracking improved by strengthening the same circuit, and \citet{lee2024dpo} show DPO lowers toxicity by bypassing rather than removing the responsible directions. For RL-era post-training the picture is less settled. \citet{shi2026rl_generalize} train SFT and RL from one base and find RL's feature changes restrained next to SFT's; \citet{rojasnunez2026forgetting} find SFT disrupts head-level circuits more than Dr.GRPO~\citep{liu2025drgrpo}; \citet{rl_circuit_intensity2026} report RL raising activation intensity and diversity, with DPO departing from that trend; \citet{sinha2026mechanistic} compare six alignment algorithms across three families. Our question comes before theirs: how large must such a difference be before it exceeds what a second run of the same recipe produces?

\paragraph{Model diffing with sparse dictionaries.} Sparse autoencoders~\citep{bricken2023monosemanticity,templeton2024scaling,gao2024scaling}, transcoders~\citep{dunefsky2024transcoders} and attribution graphs~\citep{ameisen2025circuit,lindsey2025biology,marks2024sparse_circuits} give a basis in which two checkpoints can be compared feature by feature. Diffing work builds on them: crosscoders~\citep{lindsey2024crosscoders} separate base-only from chat-only latents once their sparsity artifacts are corrected~\citep{minder2025crosscoders}, stage-wise fine-tuning of a base SAE isolates the features a fine-tune changes~\citep{bricken2024stagewise}, dedicated-feature crosscoders localize RL-induced tool use~\citep{crosscoder_tooluse2026}, and transcoder adapters diff a base model against its reasoning-distilled variant~\citep{transcoder_diffing2026}. Closest to our SFT result, \citet{minder2025narrow} show that narrow fine-tuning leaves biases in activation differences, read on the first tokens of random text, strong enough to recover the fine-tuning domain, and that mixing pretraining data into the fine-tuning set largely removes them. Our SFT set is narrow in their sense ($900$ function-calling rows), so its signature may be such a bias; what we add is the rerun band its size is judged against, and the same test applied to DPO and GRPO. We use the plainest member of the family, a base-trained SAE applied to every checkpoint with a co-activation graph over its codes, because a calibration result should not depend on a dictionary fitted to the models being compared; a random-dictionary control~\citep{korznikov2026sanity} checks that it does not depend on the trained dictionary either. Following \citet{five_stage_causal_sae2026}, we treat attribution mass as correlational and do not read displacement as a changed mechanism.

\paragraph{Run-to-run variability as the reference.} Representations are not reproducible across seeds by default. Networks trained from different initializations learn partly different units~\citep{li2016convergent}; BERT reproductions that differ only in seed support different conclusions~\citep{sellam2022multiberts}; fine-tuning outcomes vary with initialization and data order alone~\citep{dodge2020finetuning}. The instruments vary too: SAEs trained on identical data share roughly $30\%$ of their features~\citep{paulo2025different}, individual features are seed-unstable while their span is not~\citep{unstable_subspaces2026}, apparent dictionary stability can be an initialization artifact~\citep{brzozowski2026ablating}, and SAE benchmark scores carry more reseed noise than assumed~\citep{chanin2026saebench}. \citet{dead_salmons2026} argue that interpretability findings should be tested against explicit alternatives because the standard tools explain even random networks. These works measure seed variation in outcomes, units or dictionaries. We use the seed variation of the post-training run itself, read through the diffing estimator, as the reference a base-to-post difference must exceed.

\paragraph{Behavioral and weight-space accounts of RL post-training.} Pass@$k$ analyses argue that RLVR sharpens sampling within the base model's support rather than expanding it~\citep{yue2025}; a single training example recovers much of the RLVR gain~\citep{wang2025}; base models align in context without weight updates~\citep{lin2024}; and RLHF trades output diversity for generalization relative to SFT~\citep{kirk2024diversity}. Weight- and state-space accounts point the same way: RLVR weight deltas are close to rank one~\citep{relex_rank1_rlvr2026}, post-training can be read as reshaping the state distribution~\citep{posttraining_states_not_tokens2026}, and base models already hold the reasoning mechanisms that thinking models learn to deploy~\citep{venhoff2026base_reason}. None of these measures sparse features, and none says how large an internal footprint RL should leave. The seed-calibrated audit supplies that reference for one regime, light GRPO on a $1.7$B model.

\section{Problem Setup}
\label{sec:method}
\paragraph{Runs.} A recipe $A$ (objective, data, hyperparameters, step count) maps the base model $M_0$ and a seed $s$ to a checkpoint $M^A_s$. The base is \texttt{Qwen3-1.7B}~\citep{yang2025qwen3}, post-trained on BFCL function calling~\citep{patil2024bfcl}: four LoRA-SFT runs (rank-$16$, $\alpha{=}32$ on the attention projections, $0.37\%$ of parameters, $29$ steps), four LoRA-DPO runs~\citep{rafailov2023dpo} at $12$ steps plus four retrained runs read at $29$ and $60$ steps, and six full-parameter GRPO-family runs~\citep{shao2024deepseekmath} (an accuracy-only baseline, four further seeds, one shaped-reward variant, $100$--$300$ steps, so that band mixes seed, step and reward variance; L7). Every run starts from $M_0$; the seeded LoRA runs are re-seeded before model construction, so they differ in adapter initialization and not only in data order. A second backbone (\texttt{Llama-3.1-8B}~\citep{grattafiori2024llama3}, four LoRA-SFT runs) and the public Tulu-3 staircase complete the roster (App.~\ref{app:setup}).

\paragraph{Co-activation graph.} Let $X$ be a fixed probe set ($n{=}500$ Pile prompts~\citep{gao2020pile}, last-position residual states) and $f^{\mathrm{src}},f^{\mathrm{dst}}$ SAE encoders trained on $M_0$ only~\citep{bricken2023monosemanticity,templeton2024scaling,gao2024scaling} (\texttt{qwen-scope}~\citep{deng2026qwenscope}, TopK with $k{=}50$, $d{=}32{,}768$, layers $14\to22$). With codes $z^{\mathrm{src}}(x;M)=f^{\mathrm{src}}(h_{14}(x;M))\in\mathbb{R}^{d}_{\ge0}$ and likewise $z^{\mathrm{dst}}$,
\begin{equation}
w_{ij}(M)=\frac{1}{|X|}\sum_{x\in X} z^{\mathrm{src}}_i(x;M)\,z^{\mathrm{dst}}_j(x;M)\,\mathbf{1}[(i,j)\in T_x],\quad E(M)=\mathrm{top}_N\,w(M).
\label{eq:graph}
\end{equation}
Here $E(M)$ keeps the $N=10{,}000$ largest weights, and $T_x$ holds the $2{,}000$ largest products at state $x$ (of at most $2{,}500$ non-zero ones), as in the builder in the supplement. Because one basis serves every checkpoint, $E(M)$ and $E(M')$ are subsets of the same index set. Base-trained SAEs usually transfer to chat-tuned checkpoints~\citep{kissane2024saes}; here the base SAE reconstructs base, SFT and GRPO states equally well (App.~\ref{app:extended}). The graph is a co-activation proxy, not a gradient or causal attribution; App.~\ref{app:localization} shows an edge can move while the causal coupling it summarizes does not.

\paragraph{Levels and the matched-base control.} A single-graph statistic $T$ is either the Gini coefficient of inbound weights $W_j=\sum_i w_{ij}$ over active destination features, or the share of $\sum_j W_j$ held by the top $1\%$ of them. Single-checkpoint studies read $T(M^A_s)$; we attribute to post-training only $\Delta T=T(M^A_s)-T(M_0)$, computed with the same SAE and probes. A degree-preserving permutation of the weights answers a different question (concentrated versus chance).

\paragraph{Displacement and the own-run band.} Edge Jaccard is $J(M,M')=|E(M)\cap E(M')|/|E(M)\cup E(M')|$ and displacement is $D=1-J$. For $k$ runs of recipe $A$, the contrasts are $c_s=J(M_0,M_s)$ and the seed pairs are $d_{st}=J(M_s,M_t)$ for $1\le s<t\le k$, $m=k(k-1)/2$ of them. We summarize the band by
\begin{equation}
\mu_{\mathrm{seed}}=\frac1m\sum_{s<t}d_{st},\qquad
\sigma^2_{\mathrm{seed}}=\frac{1}{m-1}\sum_{s<t}\big(d_{st}-\mu_{\mathrm{seed}}\big)^2,\qquad
z_s=\frac{c_s-\mu_{\mathrm{seed}}}{\sigma_{\mathrm{seed}}},
\label{eq:band}
\end{equation}
and call $c_s$ \emph{outside} the band when $c_s<\min_{s<t}d_{st}$, reporting $K_{\mathrm{out}}=\#\{s:c_s<\min d\}$ out of $k$. The $m$ pairs share runs, so $z_s$ is a distance scale, not a tail probability. \emph{Precision rule:} a statistic is scored only if $\sigma_{\mathrm{seed}}$ exceeds its reporting step $\delta=0.0001$. We therefore commit to \textbf{edge Jaccard} as the scored statistic ($\sigma_{\mathrm{seed}}>10\delta$ on every band in Table~\ref{tab:main}); the weighted edge cosine fails the rule on trained dictionaries and is reported only descriptively (App.~\ref{app:extended}). \emph{Build rule:} our graphs rebuild bit-for-bit on the GPU architecture that built them but not across architectures, where a rebuild moves edge Jaccard by about one seed-pair distance (App.~\ref{app:build}). Every band and its contrasts are therefore scored with graphs built on one architecture, and each rebuild job records its GPU and library versions.

\paragraph{Dose.} For DPO every quantity in (\ref{eq:band}) is indexed by the step count $\tau\in\{12,29,60\}$. The mean displacement $\bar D(\tau)=1-\frac14\sum_s c_s(\tau)$ is flat across the three doses, the seed-pair band moves little, and at every dose the contrasts straddle the band's maximum (Table~\ref{tab:main}, Fig.~\ref{fig:bands}; values in App.~\ref{app:extended}). Under one architecture $K_{\mathrm{out}}(\tau)=3,1,3$ of $4$ although the contrasts $c_s(\tau)$ are flat: the counts show no consistent dose ordering. The archived graphs, whose base was built on a different GPU architecture from most run graphs, gave $0,2,4$, a curve that lay within the build difference (App.~\ref{app:build}), and we withdraw the dose-dependent reading we drew from it.

\paragraph{What a band can certify.} Model the null ``the stage changes the graph no more than re-running the recipe does'' as exchangeability: under it the base behaves as one more run.

\begin{proposition}[Band exceedance under run exchangeability]\label{prop:band}
Let $R_0,\dots,R_k$ $(k\ge2)$ be random checkpoints whose joint law is invariant under permutations of the indices, let $S$ be any symmetric similarity, and let $\mathcal P=\{\{s,t\}:1\le s<t\le k\}$. Whatever the dependence among the $m=k(k-1)/2$ pairs,
\textup{(a)} $\Pr\big[S(R_0,R_1)<\min_{\{s,t\}\in\mathcal P}S(R_s,R_t)\big]\le 1/k$; and
\textup{(b)} $\Pr\big[\max_{1\le s\le k}S(R_0,R_s)<\min_{\{s,t\}\in\mathcal P}S(R_s,R_t)\big]\le 1/(k+1)$.
Both bounds are attained by some exchangeable law. \textup{(c)} If $n_1$ runs of one recipe and $n_2$ of another are jointly exchangeable, their displacements from a fixed base occupy disjoint ranges with probability at most $2\,n_1!\,n_2!/(n_1+n_2)!$.
\end{proposition}
The intuition: the band is built from only $k$ runs, and any of the $k+1$ runs could have been the odd one out. Part~(b) is the rank floor of permutation tests and conformal $p$-values~\citep{phipson2010permutation,angelopoulos2023conformal}; what the proposition adds is that the $m$ seed pairs, though more numerous than the runs, do not lower it, and that both bounds are attained. So the strongest guarantee that band exceedance gives under exchangeability alone is $p\le1/k$ for one contrast and $p\le1/(k+1)$ when all $k$ are outside, however large $|z|$ is (proof in App.~\ref{app:proof}). Three design choices follow. With $k=4$, SFT's four seeded runs, all outside, give $p\le1/5$ under this null and no smaller (the audited checkpoint, seeded differently, is not one of the $k$), so we quote $z$ descriptively and rest the SFT claim on its size, its replication on a second backbone and SAE family, and the band-free comparison of part~(c). A single public run pair ($k=2$) gives at best $p\le1/3$, which is why the Tulu-3 staircase is a consistency check (Sec.~\ref{sec:results-tulu}). And a rank quantile $1/(m+1)$, which treats the contrast as exchangeable with the $m$ seed pairs, does not follow from run exchangeability; we report such ranks as ranks. The null is a modelling choice: every run starts from $M_0$, so the base is the runs' common origin rather than an exchangeable draw, and we do not claim that the bound is conservative for edge Jaccard on these checkpoints. A contrast at band level also cannot rule out a change all runs share, so we read within-band contrasts only as non-detections.

\section{Results}
\label{sec:results}

\begin{table}[t]
\centering
\caption{\textbf{SFT moves the graph further than its own reruns on two backbones; DPO's band counts follow no dose ordering; light GRPO is not resolved from a wider band.} Each arm has $k$ independently seeded runs and $k(k-1)/2$ seed pairs; entries are displacements $D=1-J$ (Eq.~\ref{eq:band}). ``Outside'' counts contrasts more displaced than every seed pair; $z$ is descriptive (Proposition~\ref{prop:band}) and computed at full precision. Base and run graphs of every row were built on one GPU architecture (H200); the archived readings of the submitted version, which crossed architectures, are in Table~\ref{tab:archived}.}
\label{tab:main}
\footnotesize
\setlength{\tabcolsep}{4pt}
\resizebox{\textwidth}{!}{%
\begin{tabular}{llcccccc}
\toprule
Backbone & Arm & $k$ & seed-pair $D$ & $\sigma_{\mathrm{seed}}$ & contrast $D$ & outside & $z$ \\
\midrule
Qwen3-1.7B & LoRA-SFT, 29 steps & 4 & $0.0344$--$0.0381$ & $0.0014$ & $0.0571$--$0.0615$ & $5/5^{a}$ & $-19.0$ to $-15.8$ \\
Llama-3.1-8B & LoRA-SFT & 4 & $0.0222$--$0.0257$ & $0.0013$ & $0.0442$--$0.0452$ & $4/4$ & ---$^{b}$ \\
\midrule
Qwen3-1.7B & LoRA-DPO, 12 steps & 4 & $0.0298$--$0.0325$ & $0.0011$ & $0.0317$--$0.0336$ & $3/4$ & $-1.86$ to $-0.06$ \\
 & LoRA-DPO, 29 steps & 4 & $0.0303$--$0.0334$ & $0.0012$ & $0.0317$--$0.0356$ & $1/4$ & $-2.76$ to $+0.52$ \\
 & LoRA-DPO, 60 steps & 4 & $0.0294$--$0.0329$ & $0.0012$ & $0.0317$--$0.0367$ & $3/4$ & $-4.77$ to $-0.71$ \\
\midrule
Qwen3-1.7B & GRPO family$^{c}$ & 5 & $0.0354$--$0.0394$ & $0.0011$ & $0.0338$--$0.0386$ & $0/5$ & $-1.0$ to $+3.2$ \\
\midrule
\multicolumn{8}{p{\dimexpr\textwidth-2\tabcolsep\relax}}{Matched dose, no band: SFT $0.0571$--$0.0615$ vs.\ DPO $0.0317$--$0.0356$ at $29$ steps; gap $0.0216=5.0\times$ and $5.6\times$ the arms' spreads ($0.0043$, $0.0039$).} \\
\bottomrule
\multicolumn{8}{p{\dimexpr\textwidth-2\tabcolsep\relax}}{\footnotesize $^{a}$four seed runs plus the audited checkpoint. $^{b}$the registered rule is band exceedance; no $z$ is quoted. $^{c}$four seeds and one shaped reward, $100$--$300$ steps (L7); the baseline checkpoint was deleted and cannot be rebuilt.}
\end{tabular}}
\end{table}

\subsection{Single-checkpoint levels are set by the base model}
\label{sec:results-h1}
\begin{figure}[t]
\centering
\includegraphics[width=\textwidth]{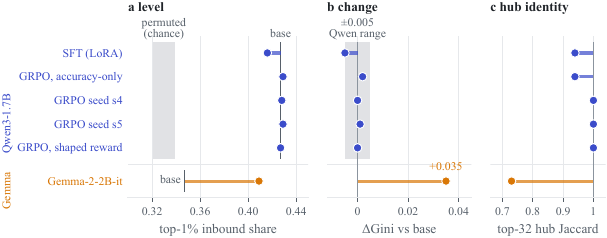}
\caption{\textbf{The concentration a single-checkpoint study would credit to RL is already present in the zero-RL base.} One row per checkpoint in the base-diff summary. (a)~Top-$1\%$ inbound share: every Qwen checkpoint sits on or next to its base (vertical rule), far from the degree-preserving permutation null of the base graph ($95\%$ range shaded). (b)~Change in Gini against the base; the shading is $\pm$ the largest $|\Delta\mathrm{Gini}|$ among the five Qwen rows ($0.005$), not a statistical null. (c)~Top-$32$ hub Jaccard against the base. Gemma-2-2B-it is a different family and a heavier recipe; the comparison is descriptive.}
\label{fig:base-control}
\end{figure}

The Qwen base graph is highly concentrated: the top $1\%$ of active destination features receive $42.7\%$ of inbound weight, far above a degree-preserving permutation null. A single-checkpoint reading of the GRPO graph would call that an RL signature. The matched base removes it (Fig.~\ref{fig:base-control}): the zero-RL base has the same concentration and nearly all of the same top hubs, and every light Qwen recipe stays within a narrow Gini band of its base with near-identical hubs (App.~\ref{app:h1extra}). \texttt{Gemma-2-2B}~\citep{gemmateam2024gemma2} with \texttt{gemma-scope}~\citep{lieberum2024gemmascope} shows the same base-set level; its heavily instruction-tuned \texttt{-it} checkpoint moves further (Fig.~\ref{fig:base-control}b), on a two-point contrast possibly confounded by the chat template (App.~\ref{app:limits}).

\paragraph{The activation-side narrowing battery is base-dominated too.}\label{sec:results-battery} The same control applies to the metrics read as narrowing: firing-frequency Gini, mean-activation Gini and usage entropy (Table~\ref{tab:battery}). Mean $L_0$ is uninformative by construction on \texttt{qwen-scope}: a TopK dictionary reads its $k$ on every checkpoint, so we drop it where sparsity is fixed and report it only for JumpReLU dictionaries~\citep{rajamanoharan2024jumprelu}. On the metrics that can move, light GRPO is base-dominated (relative changes under $1\%$, on Pile and on GSM8K~\citep{cobbe2021gsm8k} probes), and heavy instruction tuning on Gemma moves its two largest-moving metrics against narrowing ($L_0$ rises and firing-frequency Gini falls, while the other two barely move; Table~\ref{tab:battery}). A dictionary-width denominator had manufactured a second artifact, an apparent scaling law in concentration, that width-independent statistics remove (App.~\ref{app:h1extra}).

\subsection{The seed-noise-null audit}
\label{sec:results-rewiring}
\paragraph{SFT is detected against its own runs.} Scored against its six seed pairs, base$\to$SFT sits at $z=-17.8$; all five SFT contrasts fall outside the band (Table~\ref{tab:main}, Fig.~\ref{fig:bands}). Scoring SFT against the GRPO band would have been the borrowed-null substitution this audit exists to catch; on one architecture the two bands have similar widths (SD $0.0014$ vs.\ $0.0011$), so the verdict does not hinge on the substitution, but only the SFT band answers the question asked (App.~\ref{app:extended}). Three checks keep this from being an artifact of the instrument. A random-direction encoder at matched width and sparsity reproduces it against the GRPO-family band (Table~\ref{tab:randsae}; one random draw), so the detection does not need the trained dictionary; only the feature naming does. The band construction does not depend on the statistic, so crosscoder latents or activation-difference readouts can be scored against it the same way (we have not run them); we score the co-activation graph because the concentration and scaling readings that our base and width controls overturn (Sec.~\ref{sec:results-h1}) were made on it. Subsets of $2{,}000$ and $5{,}000$ edges keep SFT detected. The pre-registered Llama-3.1-8B replication, with base and run graphs now built on one architecture, puts all four contrasts beyond its band, the nearest at $1.7\times$ the band maximum. LoRA touches a small fraction of parameters while the GRPO runs are full-parameter, so parameter count does not explain the ordering, and on the Tulu-3 staircase the full-parameter SFT stage writes the largest increment (Table~\ref{tab:tulu}, no band); objective, data and duration still differ between the stages.

\paragraph{Where the displacement lands.} Of the edges that the base and all six GRPO runs keep but SFT drops, most, carrying $63\%$ of the dropped weight, feed a single destination: the base graph's largest hub, a generic numeric, whitespace and markup feature that fires on most tokens. The sources losing the most weight are boilerplate and function-token features, while the edges SFT adds are diffuse, spread more evenly over their destinations than the drops (App.~\ref{app:localization}). A plausible reading is instruction tuning learning to down-weight boilerplate. A causal probe keeps that reading correlational: ablating the source features moves the hub up to several times more than count-matched random features (an upper bound; the control is not magnitude-matched), but the effect is suppressive and no weaker in SFT than in base, so the edge drop is a change in co-activation statistics rather than a pruned causal coupling (App.~\ref{app:localization}).

\paragraph{The SFT/DPO ordering needs no band.} At SFT's $29$ steps the gap between the four SFT and four DPO runs is several times either arm's spread (Table~\ref{tab:main}), and SFT's mean displacement is well above DPO's (App.~\ref{app:extended}). Disjoint ranges are the most extreme of the $70$ equal splits of eight runs, so under exchangeable runs the two-sided permutation probability is $1/35$ (Proposition~\ref{prop:band}c; computed after the fact, not registered). Four matched-construction DPO runs with per-run data shuffling land inside the fixed-order $29$-step contrast range, with none outside their own band, and stay disjoint from SFT (App.~\ref{app:extended}). The arms still differ in learning rate and dropout, so this compares two standard recipes at matched adapter geometry and step count, not the objectives in isolation (L8).

\paragraph{DPO's band counts follow no dose ordering.} At the registered $12$ steps DPO passes its positive control (its displacement is several times the SFT band's $95\%$ width; App.~\ref{app:extended}) and does not separate under the registered rule. At every dose its contrasts straddle the maximum of their own band (the smallest is $0.95$--$0.98\times$ it), so whether a contrast lands just inside or just outside varies: $3/4$, $1/4$, $3/4$ at $12$, $29$, $60$ steps, with displacement flat (Fig.~\ref{fig:bands}). The archived graphs gave $0/4$, $2/4$, $4/4$, which we had read as dose-dependent; that curve lay within the build difference (App.~\ref{app:build}), and we withdraw the reading. No dose puts all four contrasts outside, and the DPO band lacks the data-order variance the SFT band contains (the shuffled runs above, which include it, stay inside their band; the $60$-step arm was not rerun this way), so we follow the branch registered for doses that disagree: report the counts and promote no single $z$. Training is not bitwise reproducible at fixed seed, so the registered continuity check failed and the doses are separate arms; each band is a genuine run-to-run null (App.~\ref{app:extended}).

\paragraph{Light GRPO is not resolved from its run noise.} Against the $10$ pairs of the five GRPO-family runs that can still be rebuilt on one architecture (the baseline checkpoint was deleted), all five RL contrasts sit inside the band ($0/5$; Table~\ref{tab:main}); within one run, displacement does not grow from step $100$ to $300$ (App.~\ref{app:extended}). The archived six-run band put one contrast (s1) just outside; that exceedance lay within the build difference (App.~\ref{app:build}). This band mixes seeds, steps and a reward variant, which is conservative for detection and anti-conservative for certifying absence (L7); with no equivalence margin defined, these are non-detections; at $|z|\ge2$ this band flags an edge-Jaccard shift of $0.0023$, about one-tenth of SFT's shift against it (App.~\ref{app:build}). GRPO did learn its task: BFCL accuracy rises over base, with a 95\% interval that excludes zero ($n{=}100$; App.~\ref{app:extended}). Accuracy is measured on task states and displacement on general text, and the matched on-task instrument fails its own positive control, so the pair supports no behavioral/internal comparison (L5, App.~\ref{app:ontask}).

\paragraph{Displacement is not mechanism.} Structurally distinct circuits can compute the same function~\citep{bayatmakou2026manycircuits}, so we registered their causal test and ran it at three edge budgets on $4{,}816$ held-out prompts. Corrupting the sources unique to the SFT graph moves SFT's output far more than frequency-matched random sources ($12\times$ in mean, $3.5\times$ in median at $10{,}000$ edges), but relative to the base-specific sources they matter \emph{less} in SFT than in base (the interaction $J$ is negative at $5{,}000$ and $10{,}000$ edges, with CIs excluding zero; Table~\ref{tab:functional}), and at $2{,}000$ the base-specific sources fail the gate (registered verdict: MIXED). We retire the word \emph{rewiring} (App.~\ref{app:functional}). Two auxiliary probes read each checkpoint's \emph{generated text} through a fixed external reader (Pythia-70m~\citep{biderman2023pythia} with a public SAE~\citep{marks2024sparse_circuits}) rather than the audited activations; activation deltas stay polarity-preserving and the per-feature entropy change stays below the collapse gate we set in advance (App.~\ref{app:h2h3}).
\label{sec:results-h2}
\label{sec:results-h3}

\subsection{Production scale: the Tulu-3 staircase}
\label{sec:results-tulu}
A public pipeline releases one checkpoint per stage and algorithm, so it shows what a diff can say without a seed band. On the public \texttt{Llama-3.1-8B} $\to$ Tulu-3 SFT $\to$ DPO $\to$ RLVR pipeline~\citep{lambert2024tulu3,rafailov2023dpo,schulman2017ppo}, read through base-trained \texttt{Llama-Scope} SAEs~\citep{he2024llamascope}, levels again stay close to the base (Table~\ref{tab:tulu}), and here $L_0$ is a real measurement (JumpReLU) that barely moves (App.~\ref{app:limits}). Cumulative displacement is large, with top-$32$ hub overlap falling along the staircase. SFT and DPO write most of it. The RLVR increments sit at or above the similarity of two public RLVR runs branched from the same DPO parent, a reference that also differs in algorithm and recipe. It is a single run pair with no dispersion, and by Proposition~\ref{prop:band} it could certify at best $p\le1/3$, so the staircase is a production-scale consistency check, not a calibrated null.

\begin{table}[t]
\centering
\caption{\textbf{At production scale SFT and DPO write most of the displacement, but one checkpoint per stage gives no seed band to score it against} (Tulu-3 staircase, Llama-3.1-8B, Llama-Scope layers $16{\to}25$, $n{=}500$, unnormalized v1 operating point; at the normalized one $\Delta$Gini${}\le0.019$, App.~\ref{app:tulu_robust}). Left: levels vs.\ base (base Gini $0.780$). Right: stage increments; the reference is one PPO/GRPO run pair ($0.903$), not a seed distribution.}
\label{tab:tulu}
\footnotesize
\begin{tabular}{lrr|lrr}
\toprule
Stage (vs.\ base) & $\Delta$Gini & hub Jacc & Stage increment & edge Jacc & hub Jacc \\
\midrule
SFT       & $+0.009$ & $0.73$ & base$\to$SFT   & $0.669$ & $0.73$ \\
DPO       & $+0.004$ & $0.60$ & SFT$\to$DPO    & $0.796$ & $0.73$ \\
RLVR-PPO  & $+0.005$ & $0.60$ & DPO$\to$PPO    & $0.958$ & $0.94$ \\
RLVR-GRPO & $+0.005$ & $0.56$ & DPO$\to$GRPO   & $0.906$ & $0.88$ \\
\bottomrule
\end{tabular}
\end{table}

\section{Discussion}
\label{sec:discussion}
The opening diff can mislead in a small number of recurring ways, and each has a control. A statistic can be large, significant and scale-stable and still be a property of the base model (Sec.~\ref{sec:results-h1}). A denominator tied to dictionary width can produce a scaling law from nothing. A null whose SD sits below the reporting step turns rounding into significance, which is why the precision rule exists. And agreement between groups on a direction~\citep{shi2026rl_generalize,rojasnunez2026forgetting} supplies no width; only same-recipe runs do. Proposition~\ref{prop:band} adds a quantitative version of the last point: what a band can certify grows with the number of runs, not with the effect size, so for a fixed budget more seeds buy more than a larger $z$. In numbers: under run exchangeability, one recipe's band certifies $p\le0.05$ only with $k\ge19$ reruns, all outside (part~b); a $20$-run LoRA-SFT band, run after submission under a registration, puts all $20$ contrasts outside, which reaches $p\le1/21$ (App.~\ref{app:build}). Two recipes with four runs each reach $p\le1/35$ when their displacement ranges are disjoint (part~c), a cheaper certificate for a different question (recipe versus recipe, not base versus post). Where contrasts sit at the band's edge, as DPO's do, the count is fragile (it moved with the GPU that built the graphs), so contrast and band should be reported per dose, from graphs built on one architecture, rather than as one verdict. Our nulls do not resolve a footprint of light GRPO at $1.7$B from its run-to-run variation, and this non-detection neither measures that footprint nor confirms the small-footprint weight-space and state-distribution accounts above~\citep{relex_rank1_rlvr2026,posttraining_states_not_tokens2026}: GRPO's contrasts sit at seed-pair distance (Table~\ref{tab:main}), and a within-band contrast cannot exclude a change that all runs share (Sec.~\ref{sec:method}).

What the audit does not show matters as much. Displacement beyond the band says the co-activation structure changed; it does not say the computation changed, and our registered causal sweep found the stage-specificity interaction reversed at two of three budgets. The base-set hubs admit a pretraining-crystallized reading, consistent with persona directions that form early in pretraining and still steer the post-trained model~\citep{moskvoretskii2026persona}, but we have tested it only through the matched base. And the GRPO non-detection covers light, accuracy-only GRPO at $1.7$B read on general text; long-horizon math RLVR at scale is outside this envelope.

\section{Limitations}
\label{sec:limitations}
Nine scope conditions bound the claims (full statements in App.~\ref{app:limits}). \textbf{(L1)} the full detection protocol covers one backbone ($1.7$B), one SFT recipe and general-text probe states; the one cross-family post checkpoint for the light-recipe level control (\texttt{gemma-2-2b-it}) is heavy instruct-tuning. \textbf{(L2)} the top-$32$ hub set is chosen by the inbound weight whose concentration it summarizes, not at random. \textbf{(L3)} the two generated-text probes re-tokenize each checkpoint's output for the Pythia reader, and that step has its own noise floor. \textbf{(L4)} the top-$1\%$ statistic is not width-comparable. \textbf{(L5)} the on-task (BFCL) arm is \emph{uninformative}: its $2\sigma$ flag threshold is ${\sim}3\times$ SFT's pile effect (${\sim}4\times$ in relative terms) and the positive control does not separate; hence no dissociation claim. \textbf{(L6)} the OLMo-2~\citep{olmo2025olmo2} fourth staircase is blocked: no public base-trained SAE weights. \textbf{(L7)} the GRPO band conflates seed, step and one reward recipe --- wider than pure seed noise: conservative for detection, \emph{anti}-conservative for the RL null; its baseline checkpoint is deleted, so on one architecture it has five runs. \textbf{(L8)} the DPO evidence is one LoRA recipe at $1.7$B, with lr $10\times$ below SFT's and dropout $0.0$ vs.\ SFT's $0.05$; its contrasts sit at the band's edge and its band counts follow no dose ordering, so neither a single $z$ nor a detection or non-detection at any dose is promoted, and the public Tulu-3 staircase has only one DPO checkpoint, read against the RLVR run-pair reference (App.~\ref{app:extended}). \textbf{(L9)} the behavioral check is $n{=}100$; the SFT gain is not resolved from zero on it.

\section{Conclusion}
Against same-recipe reruns, LoRA-SFT's change is large and reproducible: it displaces the co-activation graph by $17.8$ pairwise-distance SDs (a descriptive $z$) beyond its own four runs' mean, exceeds its band again on a second backbone and SAE family, and at the identical step count its four-run range is \emph{disjoint} from DPO's without a band. LoRA-DPO's displacement stays flat over three doses at the edge of its band, so its band counts carry no dose ordering, and light GRPO is not resolved from its run noise. A pre-registered causal sweep finds the displaced sources load-bearing but their stage-specificity interaction reversed at two of three budgets, so calibration buys structural change, not mechanism. Model diffs should take single-checkpoint statistics as changes from the base, sit beside their recipe's rerun band (per dose, from graphs built on one GPU architecture), treat within-band diffs as non-detections, and spend budget on reruns: under run exchangeability, $k$ runs certify no better than $p\le1/(k+1)$.

\bibliography{references}

\section*{Ethics Statement}
This work uses public language models and public corpora, involves no human
subjects and no personal data, and makes no deployment claim. Post-training
diffs are routinely read as evidence about what an alignment procedure did; the
ethical risk we address is that such readings are made without a noise floor, so
that ordinary run-to-run variation is reported as a property of the method. We
therefore calibrate every claim against repeated runs of the same algorithm and
state which effects exceed that floor and which do not. Any use of this
signature to audit a model's training history inherits the scope limits stated
in the limitations section.

\section*{Reproducibility Statement}
Protocols, the run roster and hyperparameters are in App.~\ref{app:setup}. The
supplementary material (released at \coderepo) contains every co-activation graph and receipt
JSON named in the paper; the readers and checkers that recompute the headline
SFT, DPO, matched-construction and Llama statistics without a GPU (its
\texttt{README.md} lists the commands); the graph builder; the figure
generators; the SFT and DPO trainers with their small BFCL-derived inputs, job
specifications and the DPO-loss unit tests; and all pre-registrations. It does
not contain model checkpoints, the GRPO training pipeline or the software
environments, so it supports recomputation of the reported statistics, not
bit-for-bit end-to-end retraining (App.~\ref{app:setup}). For this revision it
adds the graphs rebuilt on one GPU architecture, with each job's record of its
GPU and library versions, and \texttt{code/scripts/samearch.py}, which
recomputes Tables~\ref{tab:main} and~\ref{tab:archived} from them without a GPU.
Two sets of checkpoints can no longer be re-graphed: the GRPO baseline
checkpoint is no longer available, and the full-parameter SFT checkpoints of
the discussion-period runs were deleted after their graphs were built (their
sha256 is kept) (App.~\ref{app:build}).

\section*{Use of AI Assistants}
We used generative AI tools (Anthropic Claude and OpenAI Codex agents) to implement experiment code
and check it for correctness, to collect public datasets, to check references, and to polish the
writing. The authors take full responsibility for all content.

\appendix
\section{Proof of Proposition~\ref{prop:band}}
\label{app:proof}
Write $S(a,b)$ for $S(R_a,R_b)$ and $[k]_0=\{0,1,\dots,k\}$. Say a pair $\{u,v\}$ \emph{avoids} $a$ if $a\notin\{u,v\}$. No assumption is made about the dependence among the similarities: pairs sharing a run may be arbitrarily correlated.

\paragraph{Part (a).} For an ordered pair $(a,b)$ of distinct indices in $[k]_0$ define the event
\[
E_{ab}=\big\{S(a,b)<S(u,v)\ \text{for every pair}\ \{u,v\}\subseteq[k]_0\ \text{that avoids}\ a\big\}.
\]
The pairs avoiding $0$ are exactly the seed pairs $\mathcal P$, so $E_{01}$ is the event in (a). For any permutation $\pi$ of $[k]_0$ with $\pi(0)=a$, $\pi(1)=b$, $E_{ab}$ is $E_{01}$ evaluated on $(R_{\pi(0)},\dots,R_{\pi(k)})$, which has the law of $(R_0,\dots,R_k)$; hence $\Pr(E_{ab})=\Pr(E_{01})$ for all $k(k+1)$ ordered pairs. Let $N=\sum_{(a,b)}\mathbf 1[E_{ab}]$. Then $k(k+1)\Pr(E_{01})=\mathbb E N$, and (a) follows from the deterministic claim $N\le k+1$.

\emph{Step 1: if $E_{ab}$ and $E_{cd}$ both hold with $a\ne c$, then $d=a$ or $b=c$.} Otherwise $\{c,d\}$ avoids $a$, so $E_{ab}$ gives $S(a,b)<S(c,d)$, and $\{a,b\}$ avoids $c$, so $E_{cd}$ gives $S(c,d)<S(a,b)$, a contradiction.

\emph{Step 2: true events have at most two distinct first indices.} Suppose $E_{ab},E_{cd},E_{gh}$ hold with $a,c,g$ distinct. If $b=c$, Step~1 on $(a,g)$ forces $h=a$ (since $b=c\ne g$), and then on $(c,g)$ forces $d=g$; the events are $E_{ac},E_{cg},E_{ga}$, and since each of the pairs $\{c,g\},\{g,a\},\{a,c\}$ avoids the first index of the preceding event, $S(a,c)<S(c,g)<S(g,a)<S(a,c)$, a contradiction. If $b\ne c$, Step~1 on $(a,c)$ forces $d=a$, then on $(c,g)$ forces $h=c$, then on $(a,g)$ forces $b=g$; the events are $E_{ag},E_{ca},E_{gc}$ and the same argument gives the cycle $S(a,g)<S(g,c)<S(c,a)<S(a,g)$.

\emph{Step 3: counting.} With one first index $a$ there are at most $k$ true events (one per $b\ne a$). With two, $a$ and $c$, let $B$ and $D$ be the sets of second indices. By Step~1, every $b\in B$ and $d\in D$ satisfy $d=a$ or $b=c$. If some $b\in B$ differs from $c$, then $D\subseteq\{a\}$ and $N\le|B|+1\le k+1$; otherwise $B\subseteq\{c\}$ and $N\le1+|D|\le k+1$.

\paragraph{Part (b).} Define $F_a=\{S(a,x)<S(u,v)$ for every $x\ne a$ and every pair $\{u,v\}$ avoiding $a\}$; $F_0$ is the event in (b), and by exchangeability all $F_a$ have the same probability. For $a\ne c$, pick $g\notin\{a,c\}$ (possible since $k+1\ge3$): $F_a$ gives $S(a,g)<S(c,g)$ and $F_c$ gives $S(c,g)<S(a,g)$. So the $k+1$ events are disjoint and $\Pr(F_0)\le1/(k+1)$.

\paragraph{Attainment.} Let $U$ be uniform on $[k]_0$ and $\xi_0,\dots,\xi_k$ i.i.d.\ uniform on $(0,1)$, independent of $U$, and set $R_i=(\mathbf 1[i=U],\xi_i)$, which is exchangeable. Let $S(R_i,R_j)=\xi_j$ if $i=U$, $\xi_i$ if $j=U$, and $1+\xi_i\xi_j$ otherwise. Then $F_0$ holds exactly when $U=0$, so (b) holds with equality. For (a), $E_{01}$ holds when $U=0$, and when $U=1$ it holds exactly when $\xi_0<\xi_j$ for all $2\le j\le k$ (probability $1/k$); for $U\notin\{0,1\}$ it fails because the pairs $\{U,j\}$ avoid $0$ and have similarity below $1$. So $\Pr(E_{01})=\frac1{k+1}+\frac1{k+1}\cdot\frac1k=\frac1k$.

\paragraph{Part (c).} With the base fixed, the displacements $D_i=1-S(M_0,R_i)$ of $n=n_1+n_2$ jointly exchangeable runs are exchangeable. For each of the $n!/(n_1!\,n_2!)$ subsets $L$ of size $n_1$, the event ``every $D_i$, $i\in L$, strictly exceeds every $D_j$, $j\notin L$'' has the same probability, and these events are disjoint; the same holds with the inequality reversed. The observed labelling is one subset, so disjoint ranges in either order have probability at most $2\,n_1!\,n_2!/n!$, with equality for continuous i.i.d.\ displacements. For $n_1=n_2=4$ this is $2/70=1/35$.

\paragraph{Scope.} Exchangeability of the base with the runs is the formal content of ``no change beyond run-to-run variation''; it is a modelling choice, not something the base checkpoint satisfies literally. The proposition bounds what a band can certify under that null. It says nothing about power, and it does not turn a descriptive $z$ into a tail probability.

\section{Additional related-work comparisons}

\subsection{Independent convergence on the SFT-over-RL ordering}\label{app:related-convergence}
\paragraph{Independent convergence on the direction.} \citet{rojasnunez2026forgetting} introduce \emph{differential circuit vulnerability}, a head-level measure of how much a circuit degrades under fine-tuning, and report on \texttt{Qwen2.5-3B-Instruct} that SFT adapts faster to the target task but produces substantially greater circuit disruption, while RL (Dr.GRPO, applied on top of their SFT model) preserves a larger fraction of the base circuit. That is the ordering our dose-matched arm finds, reached independently through a different backbone, a different mechanistic unit (attention heads, not SAE features), and a different comparator (no DPO). We read the \emph{direction} --- SFT writes more than the compared alternative (Dr.GRPO there, DPO here; neither group tested the other's comparator) --- as converging across groups. What is not settled is how much of the ordering survives calibration against run-to-run variance --- the question this protocol answers, and why we report the SFT--DPO separation as disjoint four-run ranges (App.~\ref{app:related}).

\subsection{The null-calibration gap}\label{app:related-null-gap}
\paragraph{The null-calibration gap in the RL-internals literature.} The gap has teeth: \citet{nasermoghadasi2026reading} report that the strongest of $146$ SAE features clearing a Holm-corrected threshold ($d{=}+2.93$) is the top-$|d|$ feature in only $1$ of $5$ seeded runs; \citet{paulo2025different} measure ${\sim}30\%$ cross-seed feature overlap on identical data. \citet{sinha2026mechanistic} span six preference-optimization methods and three model families, using probes, SAEs and crosscoders, yet do not compare between-algorithm differences with the spread across repeated post-training runs; \citet{castillo2026direction} calibrate a fingerprint by permutation but with one model per regime, confounding regime with family. A 2025--26 wave of mechanistic work argues RL/RLVR changes model internals, each with real but partial controls --- and, to our knowledge, \emph{none} calibrates its claimed change against the run-to-run variability of the training algorithm itself. \citet{shi2026rl_generalize} is the closest prior art (RL induces \emph{restrained} feature changes vs.\ SFT's many new features) but does not ask whether the residual changes exceed seed noise: every detection in this wave is a \emph{difference read at a single run} (App.~\ref{app:related}). \citet{rl_circuit_intensity2026} report the sharpest apparent conflict, and also report that DPO deviates from their trends; we do not read that as bearing on our DPO reading, whose band counts follow no dose ordering (App.~\ref{app:related}).

\section{Behavioral sanity-check visualization}
\begin{figure}[t]
\centering
\includegraphics[width=0.42\textwidth]{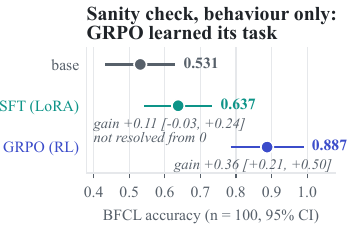}
\caption{\textbf{Sanity check: GRPO did learn the behavior it was trained for.} On the held-out BFCL validation split ($100$ items, $40$/$20$/$20$/$20$ over four categories, from the same release files as the training rows; accuracy is the unweighted category mean), GRPO improves accuracy $+0.36$ over base (95\% CI $[+0.21,+0.50]$); SFT's $+0.11$ is not resolved from zero by this harness ($[-0.03,+0.24]$). Intervals are stratified normal approximations and the gains are unpaired. \emph{Only behavior is plotted --- there is deliberately no internal axis here}, because placing the two on one pair of axes is the withdrawn dissociation claim drawn: accuracy is measured on BFCL task states, displacement on general text, and the matched on-task internal arm fails its own positive control (App.~\ref{app:ontask}, L5). The corresponding internal readings are reported where they are calibrated (Sec.~\ref{sec:results-rewiring}), not here.}
\label{fig:dissociation}
\end{figure}

Figure~\ref{fig:dissociation} plots the behavioral sanity check; provenance and intervals are in App.~\ref{app:extended}.
\section{Setup and reproducibility}
\label{app:setup}

\paragraph{Compute.}
Single-node L40S/A100/H100 (48--80\,GB), total budget approximately $35$ GPU-hours across all v1 results (a notebook estimate from job wall-clocks, with no committed accounting receipt and no itemization, so whether it covers the original GRPO training runs is not recorded; each co-activation graph or metric-battery pass takes $<10$ minutes on an L40S).

\paragraph{Backbones, checkpoints, and SAEs (v1 protocol).}
Base: \texttt{Qwen3-1.7B}. Post-training checkpoints trained by us (the SFT, DPO and step-$300$ s1 runs for this study; the other GRPO-family runs earlier, on the same task; task: BFCL function calling from the gorilla repository's release files \texttt{BFCL\_v4\_*}, single-turn categories \texttt{simple\_python}, \texttt{multiple}, \texttt{parallel} and \texttt{parallel\_multiple}, split per category (seed $0$) into the training rows and a held-out validation split; accuracy outcome-correctness reward; the SFT/DPO training scripts are in the released supplement, \texttt{code/training/} and \texttt{code/scripts/slurm/}): SFT (LoRA $r{=}16,\ \alpha{=}32$ on \texttt{q,k,v,o\_proj} --- $6{,}422{,}528$ trainable parameters, $0.37\%$ of the model; $900$ BFCL SFT rows, $1$ epoch $=29$ optimizer steps, lr 5e-5, batch $8\times$accum $4$, bf16; merged at analysis; the recipe is the one in \texttt{code/training/sft\_warmup.py} and \texttt{code/scripts/slurm/run\_sig\_sft\_seed\_null.sbatch} --- no \texttt{adapter\_config.json} is committed), plus \emph{four further seeded LoRA-SFT runs} (seeds $0$--$3$, identical recipe and step count, built for the SFT seed null of Sec.~\ref{sec:results-rewiring}; \texttt{code/scripts/slurm/run\_sig\_sft\_seed\_null.sbatch}, graphs \texttt{notes/basediff\_sft\_seed\{0..3\}\_l14to22.json}. Seeding note: the audited checkpoint set only the trainer seed, which fixes data order but not the LoRA initialization, so the four null runs are re-seeded before model construction via \texttt{code/scripts/sft\_seed.py}; seed $0$ matches the audited run's loss curve to three decimals, and bit-exact reproducibility of the SFT chain was not tested, see the determinism probe in App.~\ref{app:extended}), plus \emph{four seeded LoRA-DPO runs} built for the pre-registered DPO seed null of Sec.~\ref{sec:results-rewiring} (seeds $0$--$3$ through the same \texttt{code/scripts/sft\_seed.py} wrapper; LoRA $r{=}16,\ \alpha{=}32$ on \texttt{q,k,v,o\_proj} --- the identical $6{,}422{,}528$ trainable parameters as the SFT arm; $398$ BFCL-derived preference pairs, $1$ epoch $=12$ optimizer steps, lr 5e-6, $\beta{=}0.1$, batch $4\times$accum $8$, bf16, max length $1024$; trainer \texttt{code/training/dpo\_warmup.py} with the reference policy realized as the same weights under \texttt{disable\_adapter()}, unit-tested in \texttt{code/tests/test\_dpo\_loss.py}; pair builder \texttt{code/training/build\_dpo\_pairs.py} run \emph{once} with greedy decoding so all four seeds see an identical set, its output vendored at \texttt{data/bfcl\_dpo\_pairs.parquet}; job script \texttt{code/scripts/slurm/run\_sig\_dpo\_seed\_null.sbatch}, graphs \texttt{notes/basediff\_dpo\_seed\{0..3\}\_l14to22.json}), GRPO baseline (@ step $100$), GRPO seeds s1/s3/s4/s5 (@ steps $200$ and $300$ / $100$ / $300$ / $300$; s1 was additionally re-run to step $300$ to build the matched-step control of Sec.~\ref{sec:results-rewiring}; s3 did not reach step $300$, App.~\ref{app:limits}), shaped-$\lambda0.1$ GRPO (@ step $150$; accuracy reward plus an SAE-feature shaping penalty on the v0 panel, weight $\lambda{=}0.1$) --- all GRPO runs full-parameter verl~\citep{verl2024}, and all initialized \emph{from the base model}: the training script prefers an SFT-warmed init only when a merged \texttt{config.json} exists, which the LoRA-adapter directory lacks. Headline SAEs: \texttt{qwen-scope-3-1.7b-base-w32k-l50} residual ($d_{\text{SAE}}{=}32{,}768$; src layer $14\to$ dst $22$); scale arms: \texttt{qwen-scope-3-8b-base-w64k-l50} and the circuit-tracer Qwen3 transcoders~\citep{hanna2025circuittracer} ($0.6$--$14$B); cross-family arm: \texttt{Gemma-2-2B}/\texttt{-it} with \texttt{gemma-scope-2b-pt-res} (JumpReLU, $16$k, l13$\to$20). H2/H3 auxiliary probes use the frozen Pythia-70m reader with the public \texttt{pythia-70m-deduped-res-sm} SAE~\citep{marks2024sparse_circuits} and its v0 panel as labelled in the text (they read generated text, not the audited model's activations).

\paragraph{H1 (concentration + base control) protocol.}
Co-activation graph from $n{=}500$ pile-10k prompts (seq len $64$, last-position states), per-state top-$2{,}000$ products, then the global top-$10{,}000$ edges; top-$1\%$-of-active inbound-weight share, Gini, top-$32$ hub Jaccard, on every checkpoint with the same base-trained SAE. Degree-preserving permutation null with $3{,}000$ permutations (destinations and the weight multiset kept, weight-to-edge assignment shuffled).

\paragraph{Activation battery and displacement protocols.}
Per-checkpoint SAE encodings over all token positions of the same $n{=}500$ probe set (plus a GSM8K variant) yield mean $L_0$, dead fraction, firing-frequency and mean-activation Ginis, usage entropy, and activation-intensity levels; displacement is scored as edge-set Jaccard / weighted edge cosine between $10{,}000$-edge graphs, with the seed-noise null taken over all $15$ pairs of the six GRPO-family runs.

\paragraph{Random-dictionary control protocol.}
The control of Table~\ref{tab:randsae} replaces the trained \texttt{qwen-scope} encoder with a random-direction encoder: $d_{\text{SAE}}=32{,}768$ random unit encoder directions (Gaussian, row-normalized; seed $0$), with sparsity matched by top-$k$ on the pre-activation, $k$ calibrated to the trained SAE's mean $L_0$ on the same probe states ($\approx50$). The identical graph-building and displacement pipeline is then re-run for base, SFT, and the six GRPO-family runs on the pile protocol, and the $15$-pair seed null is recomputed under the random dictionary. Receipt: \texttt{notes/circuit\_rewiring\_randsae.json}.

\begin{table}[h]
\centering
\caption{\textbf{Random-dictionary control (pile arm): a single random dictionary preserves the SFT separation and the GRPO non-detection against the GRPO-family band.} The trained-SAE column reads the archived graphs (App.~\ref{app:build}); the random-direction graphs were not rebuild-tested. The displacement audit re-run with the trained SAE replaced by a random-direction encoder (matched $d_{\text{SAE}}=32{,}768$, top-$k$ calibrated to the trained SAE's mean $L_0\approx50$); the seed null is recomputed from the same $15$ GRPO-family pairs under each dictionary. One random draw (seed $0$): the control shows the detection does not need the trained dictionary, not that it is invariant across random ones. Random projections churn more in absolute terms (lower null mean), but the standardized separation is preserved.}
\label{tab:randsae}
\small
\begin{tabular}{lrr}
\toprule
 & trained SAE & random-dir \\
\midrule
seed null, edge Jaccard ($15$ pairs, mean $\pm$ SD) & $0.9626\pm0.0016$ & $0.8802\pm0.0025$ \\
base$\to$SFT edge Jaccard ($z$) & $0.937$ ($-15.9$) & $0.840$ ($\mathbf{-16.4}$) \\
base$\to$GRPO edge Jaccard ($z$) & $0.962$ ($-0.5$) & $0.880$ ($-0.0$) \\
base$\to$SFT weighted edge cosine & $0.9976$ (n/a$^{\ddagger}$) & $0.9645$ ($\mathbf{-16.9}$) \\
\bottomrule
\multicolumn{3}{l}{\footnotesize $^{\ddagger}$ no $z$: the trained-SAE cosine null takes $3$ distinct values over $15$ pairs} \\
\multicolumn{3}{l}{\footnotesize (SD $0.000046$, below the $0.0001$ reporting step) --- not precision-adequate.} \\
\end{tabular}
\end{table}

\paragraph{H2 (generation-side activation delta) protocol.}
Each audited checkpoint (SFT pre, GRPO post) generates a continuation for each of $16$ probe prompts using its \emph{own} tokenizer; the generated text is re-encoded with the Pythia-70m tokenizer, run through the frozen Pythia-70m reader, and its layer-4 residuals encoded with the public residual SAE \texttt{pythia-70m-deduped-res-sm}~\citep{marks2024sparse_circuits}. We report per-feature mean activation over the $32$-feature panel preselected by a function-word-outcome screen --- features ranked by association with the next token falling in a closed set of function words, over $2{,}048$ rollouts at sequence length $64$ (positive rate $0.092$); the selection receipt, including all $32$ indices and the screen's definition, is committed at \texttt{notes/panel\_v0\_pythia70m.json} and subsequently evaluated on the v0 pilot graph in this reader space (receipt: \texttt{h2\_diff\_v0\_cross\_tokenizer.json}). \emph{Cross-tokenizer alignment} thus happens at the text interface: each checkpoint generates with its own tokenizer, and the fixed reader guarantees a shared, checkpoint-independent feature space; the audited model's activations are never encoded by this SAE.

\paragraph{H3 (entropy) protocol.}
Same reader pipeline as H2: each checkpoint samples $4$ continuations ($40$ new tokens, temperature $0.7$) for each of $24$ probe prompts (the first $24$ nonempty Pile-10k rows among the first $2{,}000$ of the \texttt{train} split); each prompt-plus-continuation sequence is decoded, truncated to $128$ reader tokens, and encoded by the fixed Pythia-70m reader and SAE, giving a per-sequence mean feature activation. For each of the same $32$ panel features, we normalize its per-sequence mean activation across the $96$ sequences and compute the entropy of that distribution --- low entropy means the feature's usage concentrates onto few sequences. Compare pre (SFT) and post (GRPO) per-feature entropies (script \texttt{sig4\_h3.py}).

\paragraph{ICLR reproducibility checklist.}
\begin{itemize}[leftmargin=*]
\item All result artifacts are in the released supplement (\texttt{notes/}): every co-activation graph (Qwen base $+$ SFT $+$ six GRPO-family runs $+$ step-dose checkpoints; Gemma pair; Tulu staircase v1 and v2-normalized), the per-checkpoint activation-battery and FVU JSONs (pile, GSM8K, and on-task BFCL; App.~\ref{app:ontask}), the seed-null / displacement / localization / feature-interpretation summaries, the random-dictionary control (\texttt{circuit\_rewiring\_randsae.json} and the \texttt{randsae\_*} graphs), the four DPO-seed graphs with their null receipt (\texttt{circuit\_rewiring\_dpo\_seednull.json}) and the preference set they were trained on (\texttt{data/bfcl\_dpo\_pairs.parquet}), and the behavioral-calibration JSON --- each headline number is traceable to a named file (the transcoder scale rows of Table~\ref{tab:h1_scale} come from the sweep job scripts in the supplement, without a committed per-run JSON).
\item Analysis scripts in the released supplement (\texttt{code/} and \texttt{code/scripts/}): graph builder, battery, seed-null differencing, displacement, localization, interpretation, and log miners.
\item \emph{What is self-contained, and what is not.} The headline SFT/DPO statistics are reproducible from the supplement's graph JSONs with no GPU via \texttt{code/scripts/check\_rewiring.py}. The supplement also includes the seeded-arm trainers, small input parquets, seeding wrapper, job specifications and preregistrations; rerunning training additionally requires the named public model/SAE weights, a compatible CUDA/Slurm environment and large checkpoints that are not vendored. Specifically, the training script (\texttt{code/training/sft\_warmup.py}), its $900$-row input (\texttt{data/bfcl\_sft\_train.parquet}), the seeding wrapper (\texttt{code/scripts/sft\_seed.py}, which seeds before model construction so the LoRA initialization is fixed and not just the data order) and the job script (\texttt{code/scripts/slurm/run\_sig\_sft\_seed\_null.sbatch}) are all committed here, and the sbatch references no other repository. The graph builder now records the checkpoint that produced each graph (\texttt{checkpoint}, \texttt{hf\_model\_path}), and we have backfilled that field for the $23$ graphs feeding the headline analyses --- the Qwen base/SFT/GRPO trio, the four SFT seeds, the Tulu-3 staircase, the Gemma pair, the 8B arm, and all eight random-dictionary graphs (which additionally record the random encoder, its seed and its sparsity calibration, without which a random-dictionary graph is indistinguishable from a trained-SAE graph of the same checkpoint) --- using the literal \texttt{--hf-model-path} argument recorded in each committed job script (\texttt{code/scripts/stamp\_graph\_provenance.py}). \emph{The remaining committed graphs, largely the 8B layer sweep, predate the field and are not stamped}; their provenance is recoverable from the job scripts but is not carried in the JSON, and we do not backfill it by inference. The four SFT-seed graphs additionally carry \texttt{training\_seed} and a \texttt{training\_recipe} block (LoRA configuration, optimizer settings, Slurm job id, SHA-256 of the training script, seeding wrapper and parquet). \emph{One asymmetry we disclose rather than backfill:} the four DPO-seed graphs carry the \texttt{checkpoint}/\texttt{hf\_model\_path} stamp but \emph{not} that richer \texttt{training\_recipe} block, because the DPO trainer was written after the stamping helper; their recipe is recoverable from the committed job script and from the training log of JID $15102095$, but it is not carried inside the JSON as the SFT-seed graphs' is. One honest limit on lineage: the \emph{audited} SFT checkpoint was trained before we seeded the LoRA initialization, so while the four seed-null recipes and inputs are in the supplement, that original checkpoint is not re-derivable bit-for-bit --- the descriptive $z$ ($-17.8$ on one architecture, $-19.9$ on the archived graphs) and range separation are reproducible from the committed graph arrays, but we do not claim bit-for-bit end-to-end training reproduction. The bundle verifier checks its manifest, required files, core statistics and the DPO dose counts stated in this paper (same-architecture and archived) in a fresh directory. What a clean checkout does \emph{not} contain: the two conda environments (\texttt{trl}~$0.11.4$/\texttt{peft} for training, \texttt{sae-lens}/\texttt{transformer-lens} for analysis), the Hugging Face caches for \texttt{Qwen3-1.7B} and the SAE weights, and the GRPO/Tulu checkpoints, which are third-party artifacts or large training outputs referenced by name rather than vendored; the training scripts that produce them are released here.
\item Cross-tokenizer alignment: documented in Appendix~\ref{app:setup}; the graph-differencing code is separately checked by a self-diff identity test (\texttt{h2\_diff\_sanity.json}: one graph against itself, Jaccard/cosine $=1.0$), which is not itself a tokenizer test.
\item Statistical reporting (H3): on the committed cross-tokenizer run (24 probes $\times$ 4 continuations, \texttt{h3\_entropy\_cross\_tokenizer\_v0.json}), mean per-feature entropy delta is $-0.004$ nats with Cohen's $d=-0.005$ over 32 features (the $|d|=0.5$ entropy-collapse gate is not passed), and the hub-widening of $+0.206$ first measured on a smaller probe set does not survive --- that sub-claim is withdrawn, with the discrepancy detailed in App.~\ref{app:h2h3} (Table~\ref{tab:h3_delta}).
\end{itemize}

\section{Extended results: statistics, controls and calibration}
\label{app:extended}

This appendix holds the full text of results whose summaries appear in Sec.~\ref{sec:results}. \emph{Build note.} Paragraphs that quote the receipts \texttt{notes/circuit\_rewiring\_*.json} read the archived graphs, several of which pair a base and run graphs built on different GPU architectures; they are marked \emph{[archived build]}. Where the same-architecture reading differs, it is given in the same paragraph and in App.~\ref{app:build}.

\paragraph{The calibrated estimator replicates on a second backbone and SAE family.} We pre-registered the same construction on Llama-3.1-8B with Llama-Scope SAEs before any Llama quantity existed: layer pair fixed by fractional-depth formula (layers $16\to25$, as in Table~\ref{tab:tulu}, read at the normalization-folded operating point of App.~\ref{app:tulu_robust}), identical recipe flags, four seeds, a committed reader with four branches (\texttt{PREREGISTRATION\_llama\_second\_backbone\_v2.md}; its predecessor closed attempted-and-uninformative by its own escalation clause, having produced no number). The replication branch fired: all four base$\to$SFT contrasts displace $0.0434$--$0.0444$, every one beyond the $6$-pair seed band ($0.0222$--$0.0257$); the smallest contrast is $1.7\times$ the band's maximum. That reading paired a base graph built on A100 a month before the run graphs, which were built on H200. Rebuilt on one H200 under a registered check (\texttt{notes/PREREG\_graph\_build\_audit\_20261004.md} \S1, branch LL1), the four run graphs reproduce bit-for-bit, the contrasts are $0.0442$--$0.0452$, all four stay outside, and the smallest is $1.72\times$ the band maximum ($1.69\times$ on the archived graphs at full precision). The calibrated SFT signature is not a property of one backbone or SAE family. (Receipt: \texttt{circuit\_rewiring\_llama\_sft\_seednull.json}, read only by \texttt{llama\_repl\_read.py}; its key \texttt{rl\_signature\_beyond\_seed\_noise} keeps the committed reader's pre-existing name and its large-effect rule (band minimum $-0.02$), under which it reads \texttt{false}; the registered R1 rule is band exceedance, read by \texttt{llama\_repl\_read.py}, and the arm is SFT.)

\begin{figure}[h]
\centering
\includegraphics[width=1.0000\textwidth]{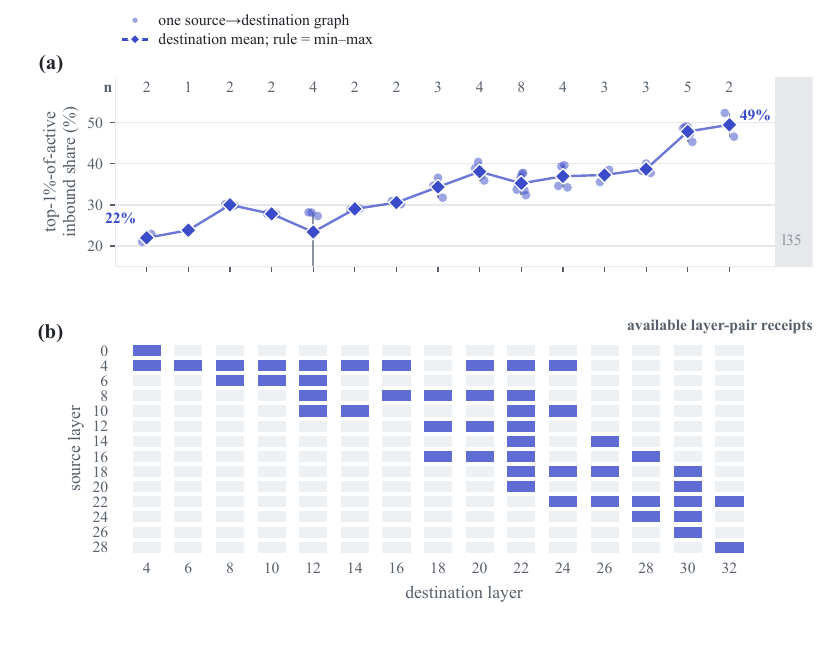}
\caption{\textbf{Available layer-pair coverage and concentration.} Every point is one source-to-destination co-activation graph; diamonds are destination means and rules the observed min--max range. Numbers above the plot count available source layers. The lower panel shows the unbalanced source/destination coverage. The rise from about 22\% to 49\% is descriptive of this available scan, not an isolated destination-depth effect. Terminal layer 35 is excluded as unembedding-dominated.}
\label{fig:h1-concentration}
\end{figure}

\paragraph{H1 graph statistics.} Across $10{,}000$ edges and $2{,}226$ destination features with non-zero in-degree (the headline \texttt{qwen-scope} protocol of Sec.~\ref{sec:method}), the top 1\% of active features capture $42.7\%$ of inbound edge weight and the top 10\% capture $65.0\%$ (Table~\ref{tab:h1_weights} and Fig.~\ref{fig:h1-concentration}, App.~\ref{app:h1extra}); the GRPO checkpoint is indistinguishable ($42.9\%$ over $2{,}225$ active features). Both the destination-degree distribution (max in-degree 2061) and the edge-weight magnitude (max $3597$, $209\times$ the median) are extreme-tailed: a small subset functions as a hub --- and a single-checkpoint reading of either graph invites an ``RL installs a reweighting signature'' conclusion; the rest of this section shows why that would be wrong.

\paragraph{The audit's detection threshold.} \emph{[archived build; the same-architecture threshold, $0.0023$, is in App.~\ref{app:build}]} \emph{The audit's detection threshold is explicit:} with null SD $0.0016$, any contrast shifting edge Jaccard by $\ge0.0032$ --- about $17$ more of the $10{,}000$ retained edges swapped out than the run-noise mean ($33$ more in the symmetric difference, since \mbox{$J=(N-x)/(N+x)$} for $x$ swapped edges) --- is flagged at $|z|\ge2$ (a flag threshold over $15$ dependent pairs, not a power calculation at a stated $\alpha$). SFT's shift is $7.9\times$ this threshold; no RL contrast exceeds $1.0\times$ it.

\paragraph{SFT against the GRPO-family band, and GRPO across steps.} \emph{[archived build; on one architecture none of the five rebuildable GRPO contrasts is outside its band, App.~\ref{app:build}]} First, the \emph{SFT} step displaces the graph at $z=-15.9$; because $n{=}15$ pairs cannot license a Gaussian tail we also give a descriptive range statement: base$\to$SFT falls four times the entire null range below its minimum (rank $1$ of $16$, a rank rather than a tail probability; Proposition~\ref{prop:band}), while only one base$\to$RL contrast (s1) falls below the band, by $0.21\times$ the null range ($19\times$ less). By rank alone s1 ties SFT (both rank $1$ of $16$); the separation is carried by \emph{effect size}, not band membership. Notably SFT is the \emph{LoRA} stage (rank-$16$, $\alpha{=}32$, attention projections only --- $0.37\%$ of the model) while the GRPO runs are full-parameter, so the detected displacement is not a trivial consequence of updating more parameters. That rules out a parameter-count triviality, not the algorithm-level reading: the stages also differ in objective, data and duration (Tulu-3's full-parameter SFT, which writes the staircase's largest increment but has no seed band, only partially closes the parameterization axis). Second, the seed-noise null converts ``the base$\to$GRPO change is small'' into ``not resolved above RL run-to-run noise'' (no equivalence test; s1 falls $0.0006$ below the three-pair matched band) --- a non-detection even where a difference measure had the best chance of exposing one. Nor does one accumulate with training: within a single GRPO run (seed s4), base$\to$checkpoint displacement is flat across training steps ($z=+0.3$/$+0.1$/$+0.8$ at steps $100$/$200$/$300$; hub Jaccard $0.94$--$1.00$ throughout).

\paragraph{The concentration is far above a degree-preserving null --- but that only rules out chance, not the base model.} A degree-preserving permutation null ($3{,}000$ permutations; App.~\ref{app:setup}) puts the null top-1\% share at $33.0\%\pm0.5\%$; the observed $42.7\%$ is $z=20.0$ above it ($p\le1/3{,}001\approx0.000333$, the empirical floor of a $3{,}000$-permutation test; $z=10.5$ at 8B). \emph{A trap worth naming:} the null answers ``above chance?,'' not ``did RL cause it?'' --- a single-checkpoint study stopping here reports a $z=20$ effect and lets the reader infer an RL mechanism.

\paragraph{Which statistic we score on, and why the other one cannot carry a $z$.} A standardized score is only meaningful if the null's dispersion exceeds the precision at which the underlying quantity is computed and reported. We commit to \textbf{edge Jaccard} as the audit's scored statistic, and to reporting the weighted edge cosine \emph{descriptively}, because on the trained dictionary the cosine fails that test. Our similarities are recorded to four decimals; the $15$-pair trained-SAE cosine null takes only \emph{three distinct values} ($0.9989$, $0.9990$, $0.9991$) with SD $0.000046$ --- \emph{below} one unit in the last recorded place. Dividing by that SD manufactures large numbers from rounding: the $z=-30.7$ a previous draft quoted for base$\to$SFT weighted cosine is an artifact of the reporting grid, not a measured effect size, and we withdraw it. Edge Jaccard has no such problem: its null spans $14$ distinct values across the $15$ pairs with SD $0.0016$, sixteen times the reporting step, so $z=-15.9$ (SFT) and $z=-0.5$ (GRPO) are computed against a genuinely resolved distribution. The same check passes for \emph{both} statistics under the random dictionary (null SDs $0.0025$ and $0.00057$, i.e.\ $25$ and $6$ reporting steps), which is why the random-dictionary cosine $z=-16.9$ in Table~\ref{tab:randsae} is quoted while its trained-SAE counterpart is not. The substantive conclusions are unchanged --- they were always carried by edge Jaccard --- but the discipline is the paper's own: a null-calibrated statistic must first beat its own numerical precision. \emph{[archived build]} The other base$\to$RL contrasts fall at $z\in[-2.0,+0.8]$ (the one marginal case, s1 at $z=-2.0$, at matched step $300$ still sits $0.0006$ below the three-pair same-step band while inside the $15$-pair one ($z=-0.98$), so its marginal exceedance is not a step artifact; it is $19\times$ smaller than SFT): GRPO displaces the graph no more than one GRPO seed does relative to another --- not resolved beyond run noise, which neither measures nor excludes a change shared by all runs.

\paragraph{Weighted edge cosine at full precision.} The four-decimal limit above belongs to the stored similarity receipts, not to the graphs. Recomputing $1-{}$weighted edge cosine (cosine of $|w|$ over the union of retained edges) at full floating-point precision from the committed graphs, for the same runs and bands, gives, on the archived graphs, the same outside counts as edge Jaccard for LoRA-SFT ($4/4$; band maximum $0.00106$, smallest contrast $0.00228$, $2.2\times$ the band maximum), the GRPO family ($1/6$) and DPO at $29$ and $60$ steps ($2/4$, $4/4$); at $12$ steps and for the matched-construction DPO runs it puts one contrast outside where edge Jaccard puts none ($1/4$ vs.\ $0/4$ each). We report counts only, no $z$. The check was pre-registered as descriptive, with no branch (\texttt{notes/PREREG\_sftband\_20260926.md}, analysis R0, which names SFT, GRPO and DPO at $12$ and $29$ steps; DPO at $60$ steps and the matched-construction runs were read in the same pass). Receipt: \texttt{results/sftband\_20260926/r0\_stored/r0\_stored\_graphs.json}; reader \texttt{code/scripts/sftband\_readouts.py}, which reproduces the committed archived \mbox{Qwen3-1.7B} edge-Jaccard bands and counts (Table~\ref{tab:archived}) before adding the cosine column. On the same-architecture rebuilds the cosine puts LoRA-SFT $20/20$ outside ($k=20$), GRPO $0/5$ and DPO at $29$ steps $1/4$ (App.~\ref{app:build}).

\paragraph{A per-algorithm null for the SFT stage: the positive half, calibrated by its own standard.} The audit's thesis is that a post-training-internals claim must be scored against \emph{that algorithm's} run-to-run variability. One SFT checkpoint scored against the GRPO-family band would not meet that bar: it is a borrowed null, exactly the substitution the introduction criticizes. We therefore trained three further LoRA-SFT runs plus a re-seeded replicate of the audited recipe (four seeded runs, identical data, hyperparameters and step count; App.~\ref{app:setup}) and built the SFT stage's own $6$-pair run null. These six distances share four run endpoints, so $z$ is a descriptive distance scale, not a calibrated tail statistic; the four-run range separation and cross-backbone replication carry the inference. Its band is $[0.9619,0.9656]$ (six distinct values, SD $0.00136$, $14\times$ the reporting step). With base and runs built on one architecture, base$\to$SFT sits at $\mathbf{z=-17.8}$, $5.9\times$ the null range beyond the most extreme null pair, and all five base$\to$SFT contrasts land at $z\in[-19.0,-15.8]$ ($J=0.939$--$0.943$), every one of them below the SFT seed band, while the six seed pairs stay inside it. Against the five-run GRPO band rebuilt on the same architecture (SD $0.00113$, mean $0.9625$) the audited contrast would read $z=-19.8$: the two bands have similar widths, so the right null changes the scale little but answers the right question. \emph{[archived build]} On the archived graphs the SFT band was the tighter one (SD $0.00137$ vs.\ $0.00160$; mean $0.9644$ vs.\ $0.9626$), and base$\to$SFT moved from $z=-15.9$ to $z=-19.9$ ($6.7\times$ the null range, against $4.0\times$), with all five contrasts at $z\in[-20.0,-17.0]$ ($0.937$--$0.941$). The stage that writes a footprint does so reproducibly, and the stage that does not (GRPO) is compared against a like-for-like construction. Receipt: \texttt{notes/circuit\_rewiring\_sft\_seednull.json}. (Consistent with Sec.~\ref{sec:method}'s precision rule, the weighted cosine again fails the precision-adequacy check on this null --- SD $0.000041$, below the reporting step --- and we quote no $z$ for it.)

\paragraph{A per-algorithm null for DPO: the same construction, returning a negative at the registered 12-step dose.} The asymmetry of calibrating SFT and GRPO against their own algorithms but DPO against neither is closed here by measurement rather than by scoping. \emph{The design was fixed first.} \texttt{PREREGISTRATION\_dpo\_seed\_null.md} was committed before any DPO seed was trained: recipe, seeds, null construction, contrast, a $<\!200$-pair data floor, and four exhaustive interpretation branches (D1 separates, D2 does not, D3 degenerate band, D4 under-powered data). Git history is the timestamp, and \texttt{notes/circuit\_rewiring\_dpo\_seednull.json} did not exist when it was written.

\emph{The null [archived build].} Four LoRA-DPO runs (seeds $0$--$3$) through the same pre-model-construction seeding wrapper the SFT null used, on $398$ BFCL-derived preference pairs, give the same $6$-pair null over $4$ runs: mean $0.96705$, SD $0.002202$, band $[0.9643,0.9706]$, six distinct values, SD $22\times$ the reporting step --- so the statistic is precision-adequate here as it is for SFT and GRPO.

\emph{The 12-step verdict is D2, and the D3 positive control is what makes it readable.} A null is worth nothing if the arm could not have detected anything, so the registration required the DPO runs to displace the graph by more than the SFT seed band's width or be declared unable to resolve. They clear it comfortably: base$\to$DPO displacement ($1-J$) is $0.0352$/$0.0354$/$0.0352$/$0.0327$ across the four seeds, i.e.\ $6.1$--$6.6\times$ the width of a $95\%$ interval on the SFT seed null ($3.92\times0.001367=0.00536$) and $8.8$--$9.6\times$ its literal min--max width ($0.9656-0.9619=0.0037$). We quote the $95\%$ width because it is the more conservative reading of ``the SFT seed band's width'' as registered; on the min--max reading the control clears by more, not less. The recipe trained and the instrument can resolve --- and at this dose \emph{every} base$\to$DPO contrast falls \emph{inside} the DPO band ($z=-1.02$, $-1.11$, $-1.02$, $+0.11$; the scored run is seed $0$ at $z=-1.02$). Put the two stages side by side on the same axis: SFT's base-to-post change is $1.8\times$ its own seed-to-seed change ($0.0628$ vs.\ $0.0356$), DPO's is $1.07\times$ ($0.0352$ vs.\ $0.03295$). DPO displaces $3.5\%$ of a $10{,}000$-edge graph and that is, on this recipe, what two runs of it do to each other. At this registered dose, this is the paper's thesis meeting its sharpest case: a difference can be genuinely large and still not be resolved from run-to-run noise, and only the null tells the two apart. The weighted edge cosine agrees in sign and magnitude ($z=-1.11$) but its null SD is $0.000075$, below the reporting step, so by the precision rule of Sec.~\ref{sec:method} it is reported descriptively and carries no part of the claim. \emph{Same architecture.} The four $12$-step run graphs were built on A100 and the base on L40S. Rebuilt with the base on one H200 (App.~\ref{app:build}), the band is $[0.9675,0.9702]$ and three of the four contrasts lie just outside it ($z=-1.86$ to $-0.06$; the smallest contrast is $0.98\times$ the band maximum in $D$). D1 required every contrast outside, so the registered verdict stays D2, and the positive control still passes ($5.9$--$6.3\times$).

\emph{What this does not license.} Not ``preference learning leaves no trace.'' The claim is scoped to LoRA-DPO ($r{=}16$, $\alpha{=}32$, attention projections) on \texttt{Qwen3-1.7B}, on $398$ BFCL-derived preference pairs for $12$ optimizer steps, read on general-text probe states through one co-activation statistic. Three bounds in particular. \emph{(i) Dose.} The registered ladder below gives $1/4$ contrasts outside at $29$ steps and $3/4$ at $60$ on one architecture ($2/4$ and $4/4$ on the archived build), while absolute displacement remains flat. \emph{(ii) Scale and parameterization.} $1.7$B, LoRA, one preference set. \emph{(iii) It does not transfer to Tulu-3.} The public staircase's DPO stage is full-parameter DPO on a $271$k-pair mixture; our $1.7$B LoRA run is a different procedure, and reading its seed noise onto that stage would be exactly the borrowed-null substitution this paper exists to criticize. The Tulu-3 DPO increment therefore keeps the RLVR run-pair reference it has (Sec.~\ref{sec:results-tulu}, L8). Receipt: \texttt{notes/circuit\_rewiring\_dpo\_seednull.json}; graphs \texttt{notes/basediff\_dpo\_seed\{0..3\}\_l14to22.json}; job script \texttt{code/scripts/slurm/run\_sig\_dpo\_seed\_null.sbatch}.

\paragraph{The dose-matched ladder: separating dose from objective (L8a), and what it costs.} The 12-step arm above confounds objective with dose --- it is $12$ optimizer steps against the SFT arm's $29$. We registered the fix before running it (\texttt{PREREGISTRATION\_dpo\_dose\_matched.md}, committed at 01:05:07 with the job starting 01:05:10; a three-second margin, which we state rather than round up) and retrained all four seeds on a single $60$-step trajectory with adapters at the two registered rungs: $29$ steps (SFT's exact count, the primary) and $60$ ($\approx2\times$). We chose step-matching over displacement-matching in advance because matching displacement requires building a graph, reading the number, and deciding whether to keep training --- an adaptive stopping rule on the same statistic that carries the verdict. Job 15103738; receipts \texttt{notes/circuit\_rewiring\_dpo\_dose\{29,60\}.json}.

\emph{The result that needs no null.} At the identical $29$ optimizer steps, with base and runs built on one architecture, the four SFT runs displace the graph by $0.0571$--$0.0615$ and the four DPO runs by $0.0317$--$0.0356$. The ranges are \emph{disjoint}, separated by a gap ($0.0216$) $5.0\times$ and $5.6\times$ the two ranges' own widths ($0.0043$ and $0.0039$), so SFT's mean displacement is $1.80\times$ DPO's at equal dose (archived build: $0.0588$--$0.0630$ vs.\ $0.0323$--$0.0365$, gap $0.0223$, $5.3\times$ either range's width at four decimals, ratio $1.78$). We quote the ratio of means, not of extremes, which rest on two single runs. This comparison invokes no seed band, no $z$, and no assumption about reproducibility; it is the strongest thing the ladder buys. And dose does not rescue the gap: DPO's displacement is flat across the whole $5\times$ ladder (4-run means $0.0327$/$0.0329$/$0.0342$ at $12$/$29$/$60$, with $\mu_{\mathrm{seed}}$ at $0.9684$/$0.9677$/$0.9692$ and $\sigma_{\mathrm{seed}}$ at $0.0011$/$0.0012$/$0.0012$; all twelve contrasts in $0.0317$--$0.0367$), never approaching SFT's mean $0.0594$. (The archived $12$-step rung read a contrast mean of $0.0346$, band mean $0.9671$ and SD $0.0022$.) We say ``flat over this range'' and not ``saturates'': three rungs in a narrow band support the absence of growth, not convergence to a limit.

\emph{The registered checks (numbered as registered).} (1) The D3-style positive control (a manipulation check that DPO moved the graph at all, distinct from the instrument-level positive control of Sec.~\ref{sec:intro}) passes at every rung ($5.9$--$6.9\times$ the SFT band's $95\%$ width on one architecture; $6.0$--$6.8\times$ archived). (4) Edge Jaccard is precision-adequate at every rung (null SD $11$/$12$/$12$ reporting steps; $22$/$12$/$12$ archived); the weighted cosine fails everywhere and carries nothing. (3) Monotonicity \emph{fails} --- displacement does not grow with dose --- which the registration required us to state outright rather than read a null off a run that had stopped moving; that non-growth is the finding of the previous paragraph. (2) The trajectory-continuity check \emph{fails}, and the registered response (branch M5) is to report the rungs as three independent arms at three doses rather than as one dose-response curve, which is what we do.

\emph{Why continuity failed, diagnosed rather than assumed.} Steps $1$--$12$ of the ladder runs do not reproduce the registered $12$-step runs (max $|\Delta\text{loss}|=0.064$, diverging at step $2$; these loss traces are recorded in the lab notebook, not in a committed JSON). We ran a probe (job 15109030) that trains seed $0$ twice on one node under the old and new code paths. The two agree exactly for five optimizer steps and then diverge, on provably identical batches --- as does a fresh run of the \emph{unmodified} path against the registered one on a different node of the same GPU model. So the divergence is neither our \texttt{--max-steps} edit nor the A100$\to$H100 move: \textbf{this training is not bitwise reproducible at fixed seed, code and hardware}, presumably through nondeterministic bf16 reductions, and the registered check assumed a determinism the system does not have. That is a defect in our registration, not a finding about DPO. It has one benign consequence --- each null is then a genuine \emph{run-to-run} null, not merely an initialization null, which is what ``two runs of the same algorithm'' should mean --- and one warning: no claim in this paper should be read as asserting bit-exact reproducibility of the DPO chain. (The SFT chain is untested by this probe; its reported claim is a three-decimal loss-curve match, which this does not contradict.)

\emph{Two ways of reading the band, which disagree, so we report both.} The \texttt{rl\_signature\_beyond\_seed\_noise} field in every receipt fires on \texttt{base}$\to$\texttt{focus} $<$ band minimum $-\,0.02$ --- a \emph{large-effect} rule, not band membership. Under it no DPO rung separates (on one architecture SFT's audited checkpoint clears it at a margin of $0.0217$, $0.0247$ archived; no DPO rung comes within $0.016$). Under the rule our own SFT paragraph states in prose --- every contrast outside the band while every seed pair stays inside --- no rung separates either, but the partial counts differ: on one architecture $3/4$, $1/4$ and $3/4$ contrasts lie outside at doses $12$, $29$, $60$ (per-contrast $z$ ranges $[-1.86,-0.06]$, $[-2.76,+0.52]$, $[-4.77,-0.71]$). \emph{[archived build]} The archived graphs gave $0/4$, $2/4$, $4/4$ (full-precision $z$ ranges $[-1.08,+0.13]$, $[-3.58,+0.03]$, $[-4.14,-2.58]$), which the submitted version read as dose-dependent. The registered wording rule for the same-architecture recount (\texttt{PREREG\_graph\_build\_audit\_20261004.md} \S3, branch V2) withdraws that reading: the counts show no consistent dose ordering, and the $0/4$, $2/4$, $4/4$ curve was within the build difference. Registered branch M3 (the rungs disagree) prescribes reporting the counts and promoting no single $z$, which is what we do. At every dose the contrasts straddle the band's maximum (the smallest is $0.95$--$0.98\times$ it), so a count read this close to the band's edge is not a detection.

\emph{One asymmetry between our two nulls, disclosed because it cuts the wrong way.} \texttt{dpo\_warmup.py} uses a fixed, unshuffled index order identical across seeds (registered as a known property), so the DPO null varies by initialization and execution nondeterminism but \emph{not} by data order. The SFT arm runs through \texttt{trl}'s \texttt{SFTTrainer} with \texttt{seed} set in \texttt{SFTConfig}, so its null \emph{does} include data-order variance. On one architecture the DPO band is narrower than SFT's at every dose (SD $0.0011$--$0.0012$ vs.\ $0.0014$; on the archived build it was the wider one at dose $12$, $0.0022$), partly because a variance component is structurally absent from it. That is \emph{anti}-conservative for detecting DPO, and it is the second reason we promote no dose's count. A matched-construction DPO null --- per-run seeded reshuffling --- is the specific fix; we pre-registered and ran it (\texttt{PREREGISTRATION\_dpo\_matched\_construction.md}): four fresh $29$-step runs land at $0.0330$--$0.0340$ on one architecture, inside the fixed-order $29$-step contrast range ($0.0317$--$0.0356$) and inside their own reshuffled band ($0/4$ outside; SD $0.0019$ vs.\ $0.0012$ fixed-order), and still disjoint from SFT's $0.0571$--$0.0615$; the gap to SFT is ${\sim}24\times$ the MC contrast range ($0.0231$ vs.\ $0.0010$). (Archived build: $0.0330$--$0.0356$ inside $0.0323$--$0.0365$, disjoint from $0.0588$--$0.0630$.) The data-order concern therefore retires for the disjointness comparison, deliberately not for the band reading: with data-order variance included the $29$-step reading is $0/4$ rather than $1/4$ ($2/4$ archived), and dose $60$ was not rerun this way. Receipt: \texttt{circuit\_rewiring\_dpo\_mc29.json}, read only by the committed \texttt{mc\_null\_read.py}.

\paragraph{Two disclosures the DPO arm requires.} \emph{(1) The preference set was rebuilt after the registration, and the reason is recorded in it (AMENDMENT 1).} The first build satisfied the registered condition --- ``rejected'' strings string-unequal to the gold call --- at $900/900$ rows and was nevertheless invalid for the registered purpose: \emph{every} rejected string was a Qwen3 \texttt{<think>} monologue (mean $400$ characters against the gold call's $76$; $0$ of $900$ parsed as a function call), so the preference contrast was \textbf{format} (concise call vs.\ reasoning block), not \textbf{correctness}. A DPO null trained on it would have measured what suppressing the think-block does to activation geometry, under a valid pre-registration timestamp. We fixed it at the source rather than by post-filtering: generations now use \texttt{apply\_chat\_template(\ldots, enable\_thinking=False)}, any generation that still does not parse as \texttt{name(} is dropped and counted separately as a format contrast, and \texttt{</think>} residue is stripped before comparison. The rebuilt set is $398$ usable pairs of $900$ rows --- $273$ dropped because the base model already emits the gold call (no pair exists) and $229$ dropped as non-calls --- clearing the registered $200$-pair floor. The amendment was written before any seed was trained, no DPO co-activation graph existed, and the invalid parquet was deleted rather than kept as a fallback. The general lesson is worth stating: a registration can be satisfied to the letter by data that answers a different question, and the arithmetic that exposed it here (``$0$ of $900$ rejected strings parse as a call'') was not in the registration because it did not occur to us to require it.

\emph{(2) The DPO trainer is ours, not \texttt{trl}'s.} \texttt{trl}~$0.11.4$'s \texttt{DPOTrainer} is incompatible with the \texttt{transformers}~$4.57.3$ in this environment (\texttt{get\_batch\_samples() takes 3 positional arguments but 4 were given}; JID $15099744$), so rather than chase versions we implemented the objective directly in \texttt{code/training/dpo\_warmup.py}: $\mathcal{L}=-\log\sigma(\beta[(\log\pi_\theta(y_c)-\log\pi_{\text{ref}}(y_c))-(\log\pi_\theta(y_r)-\log\pi_{\text{ref}}(y_r))])$ with $\beta{=}0.1$, completion-only masking, and the reference model realized as the \emph{same weights with the LoRA adapter disabled} --- so the reference is provably the base policy and no second model is loaded. Because a hand-written DPO can be wrong in two ways that still train and still yield a seed null of the wrong quantity, both are unit-tested against hand-computed values in \texttt{code/tests/test\_dpo\_loss.py} ($5$ checks): prompt tokens leaking into the sequence log-probability (masking, padding-invariance) and an inverted preference sign (prefer-chosen $<$ neutral $<$ prefer-rejected, with neutral $=\log 2=0.6931$ exactly). The measured loss is interpretable against that constant: seed $0$ runs $0.6641\to0.6356$ and seeds $1$--$3$ start at $0.6931$ and end $0.6198$/$0.6527$/$0.6295$, i.e.\ every run moves off neutral in the direction of preferring the chosen completion. Writing the loss out also had the side benefit of matching \texttt{sft\_warmup.py}'s data path and LoRA construction exactly, which the \texttt{trl} route did not.

\paragraph{Random-dictionary control, in full.} Trained-SAE evaluations often fail to beat random baselines --- random feature directions match trained dictionaries on interpretability, sparse probing, and causal editing~\citep{korznikov2026sanity} --- so a detection claim made \emph{through} a trained dictionary must run that gate on itself. We re-ran the entire pile-arm displacement audit with the trained SAE replaced by a \emph{random-direction} encoder --- random unit encoder directions at matched width ($d_{\text{SAE}}=32{,}768$), sparsity matched by top-$k$ calibrated to the trained SAE's mean $L_0$ ($\approx50$) --- rebuilding all eight graphs and recomputing the $15$-pair seed null under the random dictionary (protocol in App.~\ref{app:setup}). The result (Table~\ref{tab:randsae}): the SFT detection \emph{fully survives} ($z=-16.4$ random vs.\ $-15.9$ trained on edge Jaccard; $z=-16.9$ on the random dictionary's weighted cosine, whose trained-SAE counterpart is too quantized to score) and the RL null holds under both ($z=-0.0$ random, $-0.5$ trained). Two consequences, one deflationary and one useful. Deflationary: the detection is \emph{not} evidence about trained SAE features --- the single matched random projection tested here reads it out as well, which is why we do not brand it an ``SAE signature.'' Useful: the detector does not need the trained dictionary, so the seed-noise-null audit is cheap. \emph{We state this control's scope exactly, because it is one run:} it draws a \emph{single} random dictionary (seed $0$; App.~\ref{app:setup}), so it licenses ``the detection does not require the \emph{trained} dictionary,'' not ``the detection is invariant to the choice of projection'' --- that would need a null over projection seeds, which we did not run. Only the \emph{interpretive} layer (naming the displaced hub features, App.~\ref{app:localization}) requires the trained SAE, and this control cannot certify that layer.

\paragraph{Behavioral sanity check, in full (provenance and intervals).} We evaluate the checkpoints with the verl \texttt{val\_only} harness on the held-out BFCL validation split (SFT LoRA merged before evaluation), drawn from the same release files and categories as the training rows (App.~\ref{app:setup}). Accuracy is the unweighted mean of the four category accuracies on $100$ held-out items ($40$/$20$/$20$/$20$), and each 95\% CI is a stratified normal approximation, $\mathrm{Var}=\frac1{16}\sum_c p_c(1-p_c)/n_c$. The base model scores $0.531$ ($[0.43,0.63]$); SFT reaches $0.6375$ $[0.54,0.73]$; the audited GRPO checkpoint reaches $0.8875$--$0.900$ $[0.79,1.00]$ (lower: independent re-evaluation of the audited weights, upper: in-training validation; our receipts keep only the category mean for GRPO (the per-category values were not extracted from the verl logs), so its interval uses the bound $p_c(1-p_c)\le1/4$; $[0.79,1.00]$ spans the intervals of the two GRPO readings, its upper end truncated at $1$). The two gains are not on equal statistical footing: GRPO's is unambiguous (conservative unpaired $\Delta=+0.356$, SE $0.072$, 95\% CI $[+0.21,+0.50]$) while SFT's is \emph{not} resolved by this harness ($\Delta=+0.106$, SE $0.070$, 95\% CI $[-0.03,+0.24]$, which includes zero). \emph{What this measurement is for, and what it is not.} It exists to defeat one objection --- that the checkpoint reading flat internally simply never learned anything --- and GRPO's interval settles that. We draw \emph{no} behavioral/internal dissociation from it. A withdrawn earlier draft reported the contrast as a headline dissociation and quoted SFT's gain as $3\times$ smaller; both claims are superseded because the ratio divides a solid estimate by one indistinguishable from zero, and the two arms are measured on different state distributions (accuracy on BFCL task states, displacement on general text) with no calibrated on-task internal instrument to pair them, since the matched on-task audit fails its own positive control (SFT $z=-1.06$ on-task vs.\ $-15.9$ on pile, both against the GRPO-family band). \emph{Provenance:} the harness is the RL task's own held-out validation split of $100$ items ($40$/$20$/$20$/$20$ over the four training categories); the CIs are unpaired, which is conservative if an item's correctness is non-negatively correlated across checkpoints, since all checkpoints are scored on the same $100$ items but we hold no per-item records to pair on. The split sizes follow from the split script and are confirmed by the per-category accuracies (e.g.\ base \texttt{simple\_python} $0.575=23/40$). The three scores are \emph{not} from a single job: base and SFT come from JID $12408266$ and the GRPO merge-eval from JID $8439216$, the same verl \texttt{val\_only} configuration run at different times, so an unmeasured harness-version difference is not excluded. A single re-scoring job producing per-item outputs and paired intervals is the clean fix and is listed as open work.

\paragraph{On-task arm, in full (see also App.~\ref{app:ontask}).} Re-running the full audit on BFCL task states ($n{=}500$ chat-templated training prompts, last-token decision points, same protocol; App.~\ref{app:ontask}) was intended to close the cross-state gap. It cannot: the on-task instrument is structurally uninformative at the relevant effect size. Its $2\sigma$ minimum detectable effect on edge Jaccard is $0.0806$ --- $10.7\%$ of the statistic, against SFT's actual pile-arm effect of $0.0254$ ($2.6\%$) --- so the arm can only detect effects ${\sim}3\times$ larger in absolute Jaccard (${\sim}4\times$ relative to each arm's statistic) than the one the audit's positive control produces, and the SFT positive control indeed does not separate on it ($z=-1.06$ vs.\ pile $-15.9$). We therefore report the arm's outcomes descriptively (base$\to$GRPO $z=-0.23$; all seven post checkpoints $|\Delta\text{Gini}|\le0.026$; battery within seed noise on all eight informative metrics, the ninth being the TopK-pinned mean $L_0$; validation-split trio flat) together with its MDE, and draw \emph{no conclusion from it in either direction}: the RL-null claim lives on the pile arm only (App.~\ref{app:ontask} quantifies this, including the arm's contaminated null band).

\paragraph{The null is not basis-blindness: the delta itself is measured.} A remaining escape route is that the RL footprint hides where the SAE cannot see --- e.g.\ in the ${\sim}6\%$ reconstruction-error term the coarse FVE bound cannot exclude. We close this by measuring, per token, the true activation change $\Delta h$ and its SAE-represented counterpart $\Delta\hat h = \mathrm{SAE}(h_{\text{post}})-\mathrm{SAE}(h_{\text{base}})$ ($n{=}200$ prompts, $12.8$k tokens). GRPO \emph{does} move activations ($\overline{\|\Delta h\|}=25.5$; SFT $1.6\times$ more, $41.9$ --- the same ordering as every other measure), and the code over-responds rather than misses ($\|\Delta\hat h\|^2/\|\Delta h\|^2 = 3.0$ for GRPO, $1.9$ for SFT; cosine $0.26$/$0.32$; even the $10$th-percentile token retains $\ge21\%$ of its delta energy in-code) --- the signature of sparse thresholded codes. These checks rule out \emph{complete} SAE insensitivity to the RL delta --- the code responds, and over-responds, rather than missing it --- but the moderate delta-cosine ($0.26$/$0.32$) means the SAE captures the difference \emph{direction} only partially, so difference-specific blind spots cannot be fully excluded. The SAE churns under the RL delta at roughly the rate two RL seeds churn against each other, which is why every seed-nulled statistic reads null. With the FVE bound, the insensitivity objection is bounded from both ends.

\paragraph{Localizing the SFT signature (App.~\ref{app:localization}).} Having \emph{detected} SFT displacement, the same machinery names \emph{where} it acts: $63\%$ of the edge weight SFT drops (but base and all six GRPO-family runs retain) is inbound to a single destination --- $\#30329$, the rank-$1$ concentration hub --- a localized pruning of boilerplate/function-word wiring that GRPO leaves intact. A causal ablation probe keeps the claim honest: the interaction is real and feature-specific, but the edge-drop is a change in \emph{co-activation statistics}, not (by this probe) a pruned causal coupling --- strictly correlational. This interpretive layer is also the only part of the audit that requires the trained SAE: the random-dictionary control (Table~\ref{tab:randsae}) certifies the detection but has no feature identities to offer, so the hub-level naming rests on the trained dictionary and cannot be certified by that control.

\paragraph{The SAE blind-spot bound, in full.} The FVU blind-spot bound lives on the \texttt{qwen-scope} family, where the decode path is verified --- and it is tight: the base-trained SAE reconstructs base, SFT, and GRPO equally well (FVU $0.0592$/$0.0588$/$0.0593$, FVE $\approx94.1\%$). Had post-training moved activations into directions the base SAE cannot represent, post-checkpoint FVU would rise; it does not: $94\%$ of variance stays on the base basis at every stage --- the nulls are not an artifact of a \emph{wholly} blind basis (this marginal bound does not by itself exclude difference-specific blind spots; those are addressed by the delta projection above).

\paragraph{Operating-point robustness, in full (see also App.~\ref{app:tulu_robust}).} The staircase was run at both \texttt{Llama-Scope} normalization operating points and every conclusion is robust (the post-SFT stage increments agree to within $0.001$; the base$\to$SFT increment, the largest-displacement stage, shifts $0.669\to0.697$ but stays by far the dominant stage; DPO$\to$PPO straddles a rounding boundary, v1 $0.958$ / v2 $0.959$); absolute $L_0$ is never compared across SAE families, and \texttt{Llama-Scope} FVU is excluded (verified decode-path loader inconsistency).
\section{Graph-build environment, archived readings and discussion-period analyses}
\label{app:build}

\paragraph{What we found, and the registered diagnosis.} After submission we found that the archived Qwen base graph (\texttt{notes/attribution\_graph\_v1\_qwen3-1p7b\_qwenscope\_l14to22.json}) does not rebuild in the environment that built the later run graphs: the two builds are $D=0.0367$ apart, about one seed-pair distance. We registered a diagnosis before running it (\texttt{notes/PREREG\_graph\_build\_audit\_20261004.md}). Job records show that every archived graph that failed to rebuild had been built on L40S or A100 and every one that rebuilt bit-for-bit on Hopper (H100/H200). Rebuilt today on its original architecture, each of four tested archived graphs is reproduced bit-for-bit (the base and GRPO s4@$300$ on L40S; the audited SFT checkpoint and a $12$-step DPO run on A100), two builds in one job are identical on both architectures, and the same graphs rebuilt on H200 lie $D=0.03$--$0.037$ from them. Code, libraries, probe data, SAE and model weights were unchanged. The DPO graph was built on A100 a month after the others and still does not rebuild on Hopper, so the cause is the GPU architecture, not the date (registered verdict: R1 confirmed; nondeterminism within one environment excluded). Presumably the bf16 forward pass takes architecture-dependent numerical paths that reorder edges near the $10{,}000$-edge cutoff.

\paragraph{Consequence for the readings.} Several archived contrasts paired a base built on one architecture with run graphs built on another (Table~\ref{tab:archived}). Every band in Table~\ref{tab:main} is recomputed from graphs built on H200: the SFT, $29$-step DPO, GRPO and $20$-run graphs in the discussion-period jobs registered in \texttt{notes/PREREG\_sftband\_20260926.md} (JIDs $18577379$--$82$), and the Llama, $12$-step, $60$-step and matched-construction DPO graphs in the registered audit (JIDs $19223775$--$76$). Archived run graphs that had been built on Hopper rebuild bit-for-bit, so only the bases and the non-Hopper runs changed. The same-architecture recount was first made without a registration (for the $29$-step DPO, SFT and GRPO families); the audit then registered the $12$- and $60$-step rebuilds, the Llama rebuild and the wording rule before running them. What changes: (i) DPO's counts at $12$/$29$/$60$ steps are $3/4$, $1/4$, $3/4$ instead of $0/4$, $2/4$, $4/4$; by the registered rule (branch V2) the counts show no consistent dose ordering, the $0/4$, $2/4$, $4/4$ curve was within the build difference, and the dose-dependent reading is withdrawn. (ii) The GRPO family reads $0/5$ instead of $1/6$; its baseline checkpoint is deleted, so the six-run band cannot be rebuilt, and s1's marginal exceedance lay within the build difference. The step-$300$ triplet (App.~\ref{app:limits}) also crosses builds and was not rebuilt. (iii) The audited SFT checkpoint's $z$ is $-17.8$ instead of $-20.0$ (printed $-19.9$ in the submitted version, from four-decimal similarities), and the four seeded contrasts span $0.0571$--$0.0615$ instead of $0.0588$--$0.0630$. (iv) The Llama replication holds (branch LL1): the smallest contrast is $1.72\times$ the band maximum instead of $1.69\times$. Unchanged: all five SFT contrasts lie outside the SFT band, and SFT and DPO stay disjoint at $29$ steps (gap $5.0\times$ and $5.6\times$ the arms' spreads; mean ratio $1.80$, archived $1.78$). Both tables give $z$ at full precision; the submitted Table~1 used four-decimal similarities, which moved the DPO $29$-step upper end from $+0.03$ to $-0.01$ without changing a count. On one architecture the five-run GRPO band (SD $0.0011$) flags $|z|\ge2$ shifts of $0.0023$; SFT's shift against it is $9.9\times$ that, and the most displaced RL contrast reaches $0.5\times$ it.

\begin{table}[h]
\centering
\caption{\textbf{Archived build: the readings of the submitted Table~1, recomputed at full precision from the archived graphs.} Base and run graphs of most rows were built on different GPU architectures (column ``built on'': base / runs), so these readings are superseded by Table~\ref{tab:main}. The submitted version printed $z$ from four-decimal similarities (DPO $29$ steps $-3.61$ to $-0.01$, audited SFT $-19.9$); counts are unchanged by that rounding.}
\label{tab:archived}
\footnotesize
\setlength{\tabcolsep}{3pt}
\resizebox{\textwidth}{!}{%
\begin{tabular}{llcccccc}
\toprule
Arm & built on & $k$ & seed-pair $D$ & $\sigma_{\mathrm{seed}}$ & contrast $D$ & outside & $z$ \\
\midrule
Qwen LoRA-SFT, 29 steps & L40S / H200, A100$^{a}$ & 4 & $0.0344$--$0.0381$ & $0.0014$ & $0.0588$--$0.0630$ & $5/5$ & $-20.1$ to $-17.1$ \\
Llama LoRA-SFT & A100 / H200 & 4 & $0.0222$--$0.0257$ & $0.0013$ & $0.0434$--$0.0444$ & $4/4$ & --- \\
Qwen LoRA-DPO, 12 steps & L40S / A100 & 4 & $0.0294$--$0.0357$ & $0.0022$ & $0.0327$--$0.0354$ & $0/4$ & $-1.08$ to $+0.13$ \\
Qwen LoRA-DPO, 29 steps & L40S / H100 & 4 & $0.0303$--$0.0334$ & $0.0012$ & $0.0323$--$0.0365$ & $2/4$ & $-3.58$ to $+0.03$ \\
Qwen LoRA-DPO, 60 steps & L40S / H100 & 4 & $0.0294$--$0.0329$ & $0.0012$ & $0.0340$--$0.0359$ & $4/4$ & $-4.14$ to $-2.58$ \\
Qwen DPO, matched construction & L40S / H200 & 4 & $0.0315$--$0.0357$ & $0.0019$ & $0.0330$--$0.0356$ & $0/4$ & $-0.97$ to $+0.36$ \\
Qwen GRPO family & L40S / A100, L40S, H200 & 6 & $0.0336$--$0.0394$ & $0.0016$ & $0.0361$--$0.0406$ & $1/6$ & $-1.95$ to $+0.82$ \\
\midrule
\multicolumn{8}{p{\dimexpr\textwidth-2\tabcolsep\relax}}{Matched dose, no band: SFT $0.0588$--$0.0630$ vs.\ DPO $0.0323$--$0.0365$ at $29$ steps; gap $0.0223=5.4\times$ and $5.2\times$ the arms' spreads ($0.0041$, $0.0042$); the submitted version printed $5.3\times$ from four-decimal values.} \\
\bottomrule
\multicolumn{8}{p{\dimexpr\textwidth-2\tabcolsep\relax}}{\footnotesize $^{a}$four seeded runs on H200, the audited checkpoint on A100.}
\end{tabular}}
\end{table}

\paragraph{What was not rebuilt.} The six-run GRPO family (Table~\ref{tab:rewiring}; Fig.~\ref{fig:seednull}a shows the five rebuildable runs on one architecture), the step-$300$ triplet, the trained-SAE column of Table~\ref{tab:randsae} and the activation battery read archived graphs or metric files that mix architectures; they are marked \emph{[archived build]}, and the battery's registered build check (branch B3) makes it descriptive (App.~\ref{app:h1extra}). The on-task arm (four jobs, all L40S) and the normalized Tulu-3 staircase (one job, A100) were each built on one architecture. The unnormalized staircase of Table~\ref{tab:tulu} was built in three jobs on H100 and A100 and was not rebuild-tested; its post-SFT increments agree within $0.001$ with the single-job normalized staircase (App.~\ref{app:tulu_robust}).

\paragraph{Records.} Each rebuild job writes the GPU name, driver, CUDA and package versions to a record (\texttt{results/graph\_build\_audit\_20261005/run/meta/}; the discussion-period manifest records the GPU of every graph). \texttt{code/scripts/samearch.py} recomputes Tables~\ref{tab:main} and~\ref{tab:archived} from the graph files, and its self-test checks that the five same-architecture base rebuilds are identical and that the archived recompute reproduces the committed receipts.

\paragraph{Analyses registered and run after submission.} Registered in \texttt{notes/PREREG\_sftband\_20260926.md} before any of its jobs ran (JIDs $18577379$--$82$, H200), with branch sentences fixed in advance and every contrast scored against a base built in the same job (Table~\ref{tab:readouts}). \emph{L (more reruns), branch L1.} Raising SFT's band from $4$ to $20$ seeded reruns ($190$ seed pairs), all $20$ base-to-SFT contrasts ($0.0571$--$0.0635$) still lie outside it ($0.0298$--$0.0390$), the Proposition~\ref{prop:band}(b) event, which under run exchangeability certifies $p\le1/21$. \emph{F (full-parameter SFT), branch F1.} Full-parameter SFT with the same data, steps and learning rate also exceeds its own four-run band ($4/4$ contrasts outside; band $0.095$--$0.126$, contrasts $0.308$--$0.319$), so the SFT verdict is not specific to the LoRA adapter. Its checkpoints were written to node-local disk and deleted after their graphs were built; their sha256 is kept (\texttt{results/sftband\_20260926/run/meta/full\_s*\_weights.sha256}). \emph{R (other readouts), branch RS1.} Under a full-precision weighted-edge cosine, a per-feature activation difference (relative $L_2$ distance of mean SAE codes) and two dictionary-free readouts of the same layer-$22$ states (paired relative $L_2$ distance; $1-$linear CKA), every LoRA-SFT and full-parameter SFT contrast stays outside its band. \emph{RG (GRPO readouts): not run.} The registered GRPO family is the six-run band, and its baseline checkpoint is deleted; on the five remaining runs, reported descriptively, edge Jaccard and the cosine put $0/5$ contrasts outside and the other three readouts $1/5$ each.

\begin{table}[h]
\centering
\caption{\textbf{Discussion-period band counts by readout} (contrasts outside their own band; one GPU architecture; same-job bases). LoRA-SFT and full-parameter SFT carry registered branches (L1, F1, RS1); DPO at $29$ steps and GRPO are descriptive (GRPO: five runs, the baseline checkpoint being deleted).}
\label{tab:readouts}
\small
\begin{tabular}{lcccc}
\toprule
Readout & LoRA-SFT ($k{=}20$) & full SFT ($k{=}4$) & DPO, 29 steps ($k{=}4$) & GRPO ($k{=}5$) \\
\midrule
edge Jaccard & $20/20$ & $4/4$ & $1/4$ & $0/5$ \\
weighted edge cosine & $20/20$ & $4/4$ & $1/4$ & $0/5$ \\
SAE-code difference & $20/20$ & $4/4$ & $3/4$ & $1/5$ \\
dense residual distance & $20/20$ & $4/4$ & $2/4$ & $1/5$ \\
$1-$linear CKA & $20/20$ & $4/4$ & $0/4$ & $1/5$ \\
\bottomrule
\end{tabular}
\end{table}

\section{Full limitations and unabridged body paragraphs}
\label{app:limits}
\paragraph{Limitations (full text).} (L1) The decisive \emph{light}-RL base/SFT/GRPO control is at $1.7$B, replicated across five Qwen runs and a second family; the one cross-family \emph{post} checkpoint (\texttt{gemma-2-2b-it}) is heavy production instruct-tuning, not light GRPO, so it bounds ``heavy post-training,'' not ``light GRPO''; no same-backbone 8B light-GRPO trio exists in \texttt{qwen-scope}, so 8B-scale generalization is carried by the Tulu-3 staircase (Sec.~\ref{sec:results-tulu}), whose RL stage is production RLVR. Long-horizon math-RLVR at larger scale sits outside this envelope entirely: nothing here bounds RL's internal footprint in that regime, and the RL non-detection should not be read as a claim about it. (L2) The top-32 panel is H1-selected, not random. (L3) Cross-tokenizer alignment has its own noise floor. (L4) The dictionary-relative ``top-1\%-of-$d_{\text{SAE}}$'' statistic ($84.0\%$ at 8B) is not comparable across SAEs of different width (cf.\ the width dependence of SAE statistics in \citealp{geometric_wall2026}); the 8B coincidence ($\#35239$) is a shared \emph{base-model} hub pending a same-backbone 8B base/GRPO check. (L5) The on-task (BFCL) arm (Sec.~\ref{sec:results-rewiring}; App.~\ref{app:ontask}) is reported as \emph{uninformative}, not as a null: its $2\sigma$ MDE ($0.0806$ edge Jaccard, $10.7\%$ of the statistic) is ${\sim}3\times$ SFT's actual pile-arm effect ($0.0254$, $2.6\%$; ${\sim}4\times$ in relative terms), and the SFT positive control does not separate on it ($z=-1.06$ vs.\ pile $-15.9$); we state its MDE and draw no conclusion from it, so the RL-null claim rests on the pile arm alone. (L6) A fourth staircase on OLMo-2 (public base$\to$SFT$\to$DPO$\to$RLVR chain, $60$-step RLVR branches) is \emph{blocked, not omitted}: no base-trained OLMo-2 SAE weights are public (the one candidate repository holds interpretation exports, not weights), so that audit first requires training an SAE --- future work. (L7) The seed-noise band conflates seed, checkpoint-step ($100$--$300$), and reward-recipe (one shaped-$\lambda0.1$ run) variance: of the six runs entering the $15$-pair null, the only same-recipe, same-step pair is s4$\leftrightarrow$s5@$300$, so the band is wider than pure seed noise --- conservative for detection, \emph{anti}-conservative for certifying the RL null. The contamination is directly visible on-task, where base$\to$s4 lands at $z=+1.83$, \emph{above} the band's upper edge --- a base$\to$RL contrast more base-like than the seed pairs are to each other. The pile-arm matched-step control (s1/s4/s5@$300$, Sec.~\ref{sec:results-rewiring}) is the partial remedy; a $\ge3$-seed same-recipe, same-step null on every arm is the full one. (L8) \emph{Null coverage is uniform across the three algorithms we train, and the residual limits are about dose and transfer.} SFT has its own $4$-run seed null, GRPO a $6$-run one, and DPO a pre-registered $4$-run one. At the registered $12$ steps DPO does not separate (D2; on the archived build no contrast left its band, $z=-1.02$ for the scored run against mean $0.96705$, SD $0.002202$); on one architecture $3/4$, $1/4$ and $3/4$ contrasts lie outside at $12$, $29$ and $60$ steps while displacement stays flat, so the counts follow no dose ordering and the title claims only SFT. Four bounds on that result. \emph{(a) The step-count difference between the arms is removed; the recipes still differ.} The registered arm is $12$ optimizer steps on $398$ pairs against the SFT arm's $29$ on $900$ rows, so ``DPO writes less footprint than SFT'' was not separable from ``this DPO run was shorter.'' A pre-registered dose ladder addresses the step count: at the \emph{identical} $29$ steps the four SFT and four DPO runs occupy disjoint displacement ranges ($0.0571$--$0.0615$ vs.\ $0.0317$--$0.0356$ on one architecture; SFT's mean $1.80\times$ DPO's), and DPO's displacement is flat across a $5\times$ dose range, so ``still too short'' is not available as an objection. What remains is a difference in training data ($398$ pairs vs.\ $900$ rows), in \emph{learning rate} (5e-6 vs.\ 5e-5; the $10\times$ ratio was fixed in the registration, not tuned) and in LoRA dropout (DPO $0.0$, SFT $0.05$) and the fact that the band reading moves with the dose and with the GPU that built the graphs while the displacement comparison does not. \emph{(b) The claim is one recipe at one scale:} LoRA-DPO on \texttt{Qwen3-1.7B} with a BFCL-derived preference set; it licenses nothing about preference learning in general. \emph{(c) It does not transfer to the public staircase:} the Tulu-3 DPO stage is full-parameter DPO on a $271$k-pair mixture and still supplies a \emph{single DPO checkpoint}, so its increment ($0.796$) is still read against the two RLVR runs' run-pair null ($0.903$) --- a single descriptive reference that gives no guaranteed bound on run noise and is not DPO's own. Substituting our $1.7$B LoRA null for it would be the borrowed-null error this paper exists to criticize. \emph{(d) The preference data was rebuilt mid-arm} after the first build proved to encode a format rather than a correctness contrast (App.~\ref{app:extended}, AMENDMENT 1); the registered floor and the amendment are both on the record, but the arm is one build away from its registration. (L9) The behavioral calibration is $n{=}100$ ($40$/$20$/$20$/$20$ over four categories, unweighted category mean) and comes from two separate evaluation jobs; the SFT behavioral gain is not resolved from zero on it (Fig.~\ref{fig:dissociation}, App.~\ref{app:extended}).

\paragraph{Domain robustness: the null holds on a second probe domain.} Re-running the battery on the same trio with GSM8K question$+$solution chains ($n{=}500$) --- far more concentrated than pile (base firing-freq Gini $0.840$ vs.\ $0.683$), a genuinely different region of the basis --- leaves the base$\to$GRPO deltas nil (every base$\to$GRPO metric $|\text{rel }\Delta|<1\%$; App.~\ref{app:h1extra}). \emph{Scope:} GSM8K is still \emph{off}-training-distribution; the on-distribution (BFCL) audit is in App.~\ref{app:ontask}.

\paragraph{Do our top-$k$ conventions select the answer?} Three statistics in this paper select a top-$k$: the $10{,}000$ retained edges, the $32$ destination hubs, and the random-encoder top-$k$. Two exposures follow. \emph{(i) Tie-breaking.} \texttt{argsort} and \texttt{topk} order ties by index, so over sparse SAE values a slot of exactly zero would be ranked by feature id. We check the committed graphs: no exact-zero weight survives into the retained edge set, and the $32$nd and $33$rd hubs never tie, so no hub Jaccard in this paper turns on an index-ordered tie-break. One residual is disclosed: the smallest retained weight has multiplicity up to $5$, and the committed graphs do not contain the $10{,}001$st edge, so ties exactly at the cutoff cannot be bounded from them. \emph{(ii) Does the convention choose the conclusion?} \emph{[archived build; on one architecture the $N=10{,}000$ value is $-17.8$, Table~\ref{tab:main}.]} Recomputing the SFT seed null over top-$2{,}000$ and top-$5{,}000$ subsets of the same graphs leaves SFT detected at every $N$ ($z=-14.6$, $-14.9$, $-20.0$ at $N=2{,}000/5{,}000/10{,}000$; computed from unrounded similarities, whereas the four-decimal receipt gives the $-19.9$ quoted elsewhere at $N=10{,}000$). The cutoff moves the magnitude but not the conclusion, so we quote $N$ alongside the statistic. Receipt: \texttt{code/scripts/audit\_conventions.py}.

\paragraph{Why the \emph{public} DPO stage still has no per-algorithm null: the search and the cost.}
\emph{Superseded in part:} an earlier draft used this paragraph to argue that no DPO seed null should be run at all. That argument is withdrawn as to our own arm --- the pre-registered $1.7$B LoRA-DPO null above was run, and it answers a well-posed question --- but its core reasoning was right and still stands as to the \emph{public} stage, so we keep it here rather than delete it. The public Tulu-3 staircase provides exactly one DPO policy checkpoint, so the DPO stage increment ($0.796$) is scored against the two RLVR runs' pair null ($0.903$) --- a single descriptive reference that also varies algorithm and recipe, gives no guaranteed bound on same-recipe run noise, and is not DPO's own. We checked whether a second DPO policy exists publicly: it does not. The only candidate, \texttt{Llama-3.1-Tulu-3-8B-DPO-RM-RB2}, is a \emph{reward model} (a sequence classifier released with RewardBench~2) built on top of the DPO checkpoint, not a second DPO policy branched from the same SFT parent. Replicating the stage ourselves remains out of budget: Tulu-3's DPO is \emph{full-parameter} DPO on a $271$k-pair preference mixture, so $2$--$3$ seeds of it is tens of GPU-hours on multi-GPU nodes, more than the ${\sim}35$ GPU-hours estimated for all results here. And the cheap substitute is not a substitute: our LoRA-DPO run at $1.7$B measures the run noise of a \emph{different} procedure, so it is reported as its own result about its own recipe and is \emph{not} transferred onto the Tulu-3 stage. Doing that would be precisely the borrowed-null error this paper exists to criticize. Consequently the title claims only SFT; the Tulu-3 DPO increment is reported descriptively against one non-matched RLVR pair; and a seeded null on the public recipe, at production scale, is still the experiment that would calibrate it.

\paragraph{Seed-null band accounting, Tulu staircase and discussion.}
\emph{[archived build; the six GRPO-family graphs and the step-$300$ triplet mix L40S, A100 and Hopper builds, and the triplet was not rebuilt.]} The GRPO-family comparison is in Table~\ref{tab:rewiring} (Fig.~\ref{fig:seednull}a repeats it on one architecture for the five rebuildable runs). Alongside the band-exceedance count $K_{\mathrm{out}}$ of Sec.~\ref{sec:method}, we give a descriptive standardized score against the $15$ GRPO-family pairwise similarities (mean $0.9626$, SD $0.0016$) and how far a contrast falls below the most extreme null pair, in units of the null range. (The $15$ pairs share six underlying runs and are not independent; we therefore lean on the descriptive rank/range statements and quote $z$ as an effect size, not a calibrated tail probability.) \emph{An honest accounting of the band itself:} the null mixes seeds, checkpoint steps, and one shaped-reward recipe, which widens it --- conservative for detection but \emph{anti}-conservative for certifying the RL null (a wider band is easier to sit inside). The matched same-recipe, same-step (@$300$) seed null now spans \emph{three} pairs --- s1, s4, s5 all at step $300$ (s1 re-run to the matched step for this control) --- with seed-to-seed Jaccard $\{$s1$\leftrightarrow$s4 $0.9616$, s1$\leftrightarrow$s5 $0.9621$, s4$\leftrightarrow$s5 $0.9644\}$, a band $[0.9616, 0.9644]$ (mean $0.9627$); the three matched base$\to$post@$300$ contrasts are base$\to$s1 $=0.9610$, base$\to$s4 $=0.9639$, base$\to$s5 $=0.9631$ (mean $0.9627$): two lie inside the band and one (s1) lies $0.0006$ below its lower edge --- a gap smaller than the band's own width ($0.0028$) --- and the two group means are equal to four decimals ($0.9627$ vs.\ $0.9627$). At matched recipe and step, then, the two means agree to four decimals --- descriptively consistent with the base-to-RL change being the same scale as seed-to-seed run noise. We deliberately claim no more: the three seed pairs come from only three runs and the three base contrasts all share one base checkpoint, so these are dependent observations, not six independent replicates, and no equivalence margin is applied. This is a descriptive consistency, not a test of equivalence, and not a certification of the RL null. The fourth seed (s3) did not reach step $300$, so under branch~C of the pre-registered equivalence analysis (\texttt{preregistrations/PREREG\_matched\_step300\_equivalence.md} in the supplement, whose margin, statistic and branches were committed before any four-run matched graph existed) we report this three-pair result as it stands, make no equivalence claim, and record the shortfall. The SFT detection, whose $z$ only shrinks under a widened band, is unaffected. The base$\to$GRPO edge Jaccard is $0.962$: $z=-0.5$, rank quantile $0.38$ --- dead center of the seed null; the weighted edge cosine ($0.999$) is identical to the seed-pair cosine to three decimals. \emph{We name this checkpoint's step, because it is the shortest run in the audit:} the ``base$\to$GRPO'' headline is the accuracy-only GRPO baseline at \texttt{global\_step\_100}, whereas the GRPO runs reach $300$ steps. The null does not depend on reading the shortest run: the two seeds carried to step $300$ under the identical recipe give base$\to$s4 $z=+0.84$ and base$\to$s5 $z=+0.34$ --- if anything \emph{more} base-like than the seed pairs are to each other --- and s1@$300$ gives $z=-0.98$, inside the band. All three endpoint contrasts sit inside the seed band, so the step-100 choice is not what produces the null --- but we should not overstate the uniformity: s1@$300$ ($z=-0.98$) is in fact farther from zero than the step-100 headline ($z=-0.5$), so ``every endpoint is at least as null'' would be false as written. The weaker, sufficient claim is what we assert: no endpoint contrast escapes the band.

Because the Qwen arm is limited to \emph{light} GRPO, we run the full audit on a public production post-training \emph{staircase}~\citep{lambert2024tulu3}: \texttt{Llama-3.1-8B} base~\citep{grattafiori2024llama3} $\to$ \texttt{Tulu-3-SFT} $\to$ \texttt{Tulu-3-DPO}~\citep{rafailov2023dpo} $\to$ \texttt{Tulu-3-8B} (RLVR, PPO~\citep{schulman2017ppo}), plus \texttt{Tulu-3.1-8B}, a \emph{second} RLVR run (GRPO, improved recipe) branched from the same DPO parent --- a public ``run-pair null.'' Because the two runs also differ in algorithm and recipe, this run-pair null is expected to exceed pure run noise, though a single pair guarantees no such bound --- conservative for \emph{detecting} SFT/DPO-scale effects, \emph{anti}-conservative for \emph{certifying} the RL null (DPO$\to$GRPO $0.906$ vs.\ the $0.903$ pair value is razor-thin); the RL-null claim therefore rests on the Qwen seed null, with Tulu as the production-scale consistency check, not an independent certification. SAEs: base-trained \texttt{Llama-Scope}~\citep{he2024llamascope} residual SAEs (l16$\to$25, $131$k, matched relative depth). Three findings (Table~\ref{tab:tulu}; App.~Fig.~\ref{fig:staircase-data}).

\textbf{(1) Levels stay base-set even at production scale.} The base model's Gini is $0.780$, and every post-training stage sits within $+0.009$ of it (v1; $\le0.019$ at the normalized point, App.~\ref{app:tulu_robust}); the activation battery agrees (mean $L_0$ $-0.6\%$, firing-frequency Gini $-0.9\%$, usage entropy unchanged) --- the level nulls replicate in a third family and SAE basis, at production intensity. \texttt{Llama-Scope} is JumpReLU, so unlike the Qwen arm its $L_0$ is data-dependent and free to move ($1409.4\to1401.3$ across the staircase): this is one of the two arms where the sparsification metric is an actual measurement. \textbf{(2) Differences \emph{do} detect production post-training.} Cumulative displacement against base is large --- edge Jaccard $0.60$--$0.67$, with top-$32$ hub turnover $0.73\to0.56$ --- and far beyond the run-pair null of $0.90$. \textbf{(3) Stage decomposition: no RLVR increment is resolved above the single available reference.} SFT displaces most, DPO adds more, and the RLVR increments sit at or above the $0.903$ similarity of the two RLVR runs to each other while preserving the parent's hubs (Table~\ref{tab:tulu}, right): the pipeline's footprint is dominated by the SFT and DPO stages, while no RLVR increment is resolved above the single run-pair reference --- a stage-resolved dose curve consistent, at production scale, with the Qwen seed-null finding~\citep{yue2025,venhoff2026base_reason}, read as a consistency check rather than an independent replication of the null.

A signature claim can be large, statistically significant, and scale-stable --- and still be an artifact of the base model: our concentration measurement had all three properties ($42.7\%$ top-$1\%$ share, $z=20$, Gini flat across a $23\times$ parameter range) and none survived the matched base checkpoint, the one control most studies omit; a top-$1\%$-of-dictionary denominator likewise fabricated a ``scaling law'' that width-independent metrics erase --- both traps report a quantity at one operating point silently read as a contrast. A third trap is dictionary branding: our own detection headline survives random encoders (Table~\ref{tab:randsae}), so what the audit measures is activation geometry through a matched sparse projection, not trained-feature structure --- we ran the random-baseline gate of \citet{korznikov2026sanity} on ourselves and rebranded accordingly.

These cautions instantiate broader concerns: \citet{dead_salmons2026} note that attribution, probing and SAEs can produce plausible explanations even for \emph{randomly initialized} networks; ours is the training-axis instance of their warning, made constructive (matched real base, named metric, permutation/width controls, cross-family replication). Against the entropy-collapse picture~\citep{cui2025entropy_mechanism} and its behavioral counterpart~\citep{yue2025,wang2025}, our nulls do not resolve a footprint of accuracy-only GRPO at $1.7$B from its run-to-run variation --- compatible with weight-space~\citep{relex_rank1_rlvr2026} and state-distribution~\citep{posttraining_states_not_tokens2026} accounts and with \citet{venhoff2026base_reason}, though a within-band contrast does not measure the footprint's size (Sec.~\ref{sec:method}).

Given the base control, the index slots of Fig.~\ref{fig:backbone-recurrence} (App.~\ref{app:h2h3}; $13$ of a $15$-slot panel appear in at least one graph's top-$30$ across a $31$-layer scan; ids, not identities) admit a pretraining-crystallized reading that we have not tested beyond the matched base (Sec.~\ref{sec:discussion}).

\paragraph{Method, related work, battery and auxiliary probes.}
\paragraph{Backbones and SAEs.} We use the \texttt{Qwen3-1.7B}~\citep{yang2025qwen3} pre-trained \emph{base} model and its matched SFT and GRPO~\citep{shao2024deepseekmath} checkpoints we trained for this study on function calling (BFCL~\citep{patil2024bfcl}, accuracy reward --- not mathematical reasoning) supplying every audited Qwen checkpoint: SFT, an accuracy-only GRPO baseline, four further GRPO seeds, and one \emph{shaped-reward} variant (\emph{shaped-}$\lambda0.1$: accuracy reward plus an auxiliary SAE-feature shaping penalty, weight $\lambda{=}0.1$); the base checkpoint is the load-bearing control. \emph{Training provenance (disclosed in full, App.~\ref{app:setup}):} the SFT checkpoint is a LoRA~\citep{hu2021lora} adapter merged into the base at analysis time; all GRPO runs are \emph{full-parameter} verl updates initialized \emph{from the base model} --- SFT and GRPO are \emph{parallel branches from base}, so base$\to$GRPO is a clean base-vs-RL comparison (the serial-pipeline case is covered by the Tulu-3 staircase, Sec.~\ref{sec:results-tulu}). The six GRPO-family runs are checkpointed at heterogeneous step counts ($100$--$300$) and include one shaped-reward recipe, so the null band conflates seed, step, and recipe variance --- wider than pure seed noise: conservative for \emph{detecting} an effect, \emph{anti}-conservative for \emph{certifying} the RL null (Limitation L7); App.~\ref{app:limits} quantifies this with the matched same-recipe, same-step comparison. The H1 co-activation graph uses the \texttt{qwen-scope-3-1.7b-base-w32k-l50} residual-stream SAEs (src layer $14\to$ dst $22$, $d_{\text{SAE}}=32{,}768$); the 8B arm and the scale sweep use \texttt{qwen-scope-3-8b-base-w64k-l50} and the circuit-tracer Qwen3 transcoders~\citep{hanna2025circuittracer} (App.~\ref{app:setup}). Crucially the SAE~\citep{bricken2023monosemanticity,templeton2024scaling,gao2024scaling} is trained on the base model only, so the \emph{same} feature basis probes all checkpoints --- exactly why the base$\to$GRPO comparison is well-posed.

\paragraph{The probes (and the control they require).} \emph{\Hone\ (concentration + base control):} we extract a $10{,}000$-edge co-activation graph from a pinned probe set, with edge weight $w_{ij}=\frac{1}{|X|}\sum_{x\in X} z^{\mathrm{src}}_i(x)\, z^{\mathrm{dst}}_j(x)$ (mean co-activation over probe states $X$) over base-SAE codes of last-position residual streams (per-prompt top-$2{,}000$, global top-$10{,}000$ by $|w|$; App.~\ref{app:setup}). This is a co-activation (linear-probe) attribution proxy, not a gradient or causal attribution; the identical estimator is applied to every checkpoint, so all contrasts are measured alike (which does not by itself guarantee that checkpoint-dependent estimation bias cancels). We measure the top-$1\%$-of-\emph{active} inbound-weight share (and Gini) on \emph{all three} checkpoints, so any RL-specific component is the base$\to$GRPO \emph{difference}, not the level; a degree-preserving permutation null separates ``concentrated vs.\ chance'' from ``RL-caused vs.\ base''. \emph{\Htwo\ and \Hthree\ (auxiliary generation-side probes):} unlike \Hone, these never encode the audited model's activations: each checkpoint \emph{generates} continuations of a probe set, a frozen external reader (Pythia-70m with the public \texttt{pythia-70m-deduped-res-sm} residual SAE, the v0 pilot pair) encodes the generated text, and a $32$-feature panel preselected by a function-word-outcome screen --- features ranked by association with the next token falling in a closed set of function words, over $2{,}048$ rollouts at sequence length $64$ (positive rate $0.092$); the selection receipt, including all $32$ indices and the screen's definition, is committed at \texttt{notes/panel\_v0\_pythia70m.json} and subsequently evaluated on the v0 pilot graph in the \emph{reader's} SAE space is measured (App.~\ref{app:setup}, \ref{app:h2h3}). \Htwo: per-feature mean activation ($16$ probe prompts); \Hthree: per-feature entropy of per-sequence mean activation over $24$ prompts $\times4$ continuations, testing the $|d|\geq0.5$ collapse gate (a $|d|=0.5$ threshold we pre-registered ourselves, following Cohen's conventional medium effect; \citet{yue2025,wang2025} report behavioral results and state no feature-level gate, so this bar is ours to defend, not theirs) --- both probe post-training's effect on the \emph{generated prompt-plus-continuation text} through a fixed SAE lens.

\paragraph{The control generalizes across recipes/seeds and families.} The null is not specific to \texttt{Qwen3-1.7B} and the \texttt{qwen-scope} SAE; we replicate it along two axes (Table~\ref{tab:h1_robust}, Fig.~\ref{fig:base-control}; App.~\ref{app:h1extra}). \emph{Within Qwen}, we measure the same base$\to$post deltas for five independent post-training runs sharing the base-trained SAE: SFT, accuracy-only GRPO, two further GRPO seeds, and the shaped-reward recipe (shaped-$\lambda0.1$, App.~\ref{app:setup}); all five give $|\Delta\text{Gini}|\le 0.005$ and top-$32$ hub Jaccard $0.94$--$1.00$ (three at exactly $1.00$) --- the razor-thin null is not a single-seed accident. (The displacement null of Sec.~\ref{sec:results-rewiring} adds seeds s1/s3; full run roster with steps in App.~\ref{app:setup}.) \emph{Across families}, we run the identical base-vs-post control on \texttt{Gemma-2-2B}~\citep{gemmateam2024gemma2} with the \texttt{gemma-scope-2b-pt-res} SAE~\citep{lieberum2024gemmascope} --- a different family, SAE architecture (JumpReLU), and dictionary width ($16$k). The zero-RL \texttt{Gemma-2-2B} base already carries essentially the same concentration as the Qwen base (Gini $0.708$ vs.\ $0.728$, top-$1\%$-active $0.347$ vs.\ $0.427$): the base-set concentration is not specific to one backbone or SAE.

\emph{One honest nuance.} The Gemma post-checkpoint we can access is the production \texttt{gemma-2-2b-it} instruct model (heavy SFT$+$RLHF$+$distillation), \emph{not} a matched light GRPO run; its base$\to$post shift is genuinely larger ($\Delta\text{Gini}=+0.035$, hub Jaccard $0.73$). So a post-trained checkpoint can move the level (the heavy instruct \texttt{-it} checkpoint does, light GRPO does not), by an order of magnitude less than the single-checkpoint level; with one cross-family pair we do not attribute the difference to post-training intensity; the \emph{level} is base-set in both families, so a single-checkpoint level is never safe to read as an RL effect. A further caveat: \texttt{gemma-2-2b-it} is a chat model probed on raw (untemplated) pile text, so part of its larger shift may be chat-template distribution mismatch rather than post-training per se; a chat-templated re-probe is the clean follow-up.

\paragraph{Concurrent base-vs-RL studies (the claims our control tests).} Attribution graphs~\citep{ameisen2025circuit,lindsey2025biology,heimersheim2024patching,vig2020causal} and sparse feature circuits~\citep{marks2024sparse_circuits} extract the feature panels we audit; following \citet{five_stage_causal_sae2026}, we frame \Hone\ as a reweighting finding about attribution mass, not a direct causal claim, and treat \Htwo\ activation deltas as a polarity probe (App.~\ref{app:related}). Two closely related works diff a model against a post-trained descendant and report internal change --- \citet{rl_circuit_intensity2026} (before vs.\ after RL fine-tuning: RL increases activation \emph{intensity}/\emph{diversity}, online RL $>$ DPO) and \citet{transcoder_diffing2026} (base vs.\ reasoning-distilled: adapter features for hesitation tokens are necessary and sufficient) --- but neither reports a matched-base concentration-\emph{level} control; our matched-base control locates the concentration \emph{level} in the base, and the only larger \emph{residual} we observe is on a confounded cross-family \texttt{gemma-2-2b-it} checkpoint (heavy SFT+RLHF+distillation), which we do not attribute to intensity or to an online-RL$>$DPO ordering (Table~\ref{tab:h1_robust} carries no matched DPO comparator).

\paragraph{Entropy-collapse and the narrowing-feature picture.} \citet{yue2025} argue, from pass@$k$ curves, that RLVR does not expand the base model's reasoning capacity beyond what sampling already reaches; \citet{wang2025} show a single training example can recover most of the RLVR gain; \citet{lin2024} show base models can be aligned in-context without weight updates. \emph{These are behavioral results, and none of them measures SAE features} --- but together they make a feature-level prediction that the internals literature has since asserted directly: if RL only reweights what the base already computes, the features it uses should narrow. \citet{cui2025entropy_mechanism} supply the optimization-side version, policy-entropy collapse. That extrapolated prediction is what our probes test. Our cross-tokenizer entropy probe finds no collapse (committed cross-tokenizer run $\Delta H=-0.004$ nats, $d=-0.005$; App.~\ref{app:h2h3}); since the entropy drop is on-policy-specific~\citep{enaction2026}, our null may be regime-dependent --- a matched on-policy/off-policy H3 control is pre-registered; the entropy--gradient readout of \citet{entropy_gradient_inversion2026} is future work.

\paragraph{H2/H3 auxiliary nulls (App.~\ref{app:h2h3}).}
Two panel-internal probes read the real \texttt{Qwen3-1.7B} checkpoints' \emph{generated text} through a fixed external SAE reader (Pythia-70m with the \texttt{pythia-70m-deduped-res-sm} SAE; the $32$-feature panel and its indices live in the reader's space) and agree with the headline: activation deltas are polarity-preserving (weighted cosine $0.9998$), and the $|d|{=}0.5$ collapse gate (a $|d|=0.5$ threshold we pre-registered ourselves, following Cohen's conventional medium effect; \citet{yue2025,wang2025} report behavioral results and state no feature-level gate, so this bar is ours to defend, not theirs) is never approached: across the committed accuracy-only GRPO runs the per-feature entropy delta is small and sign-unstable (cross-tokenizer run $d=-0.005$; probe-budget reruns of the same baseline checkpoint span $d=-0.19$ to $+0.32$), and a reward-shaped regime reaches only $d=-0.29$ --- all far below the gate. Being output-side probes on a pilot panel, they are corroboration, not internal-activation nulls; full tables, provenance, and two \emph{withdrawn} earlier-draft claims in App.~\ref{app:h2h3}.

\subsection{Base control for the activation-side ``narrowing'' metrics}
\begin{figure}[h]
\centering
\includegraphics[width=1.0000\textwidth]{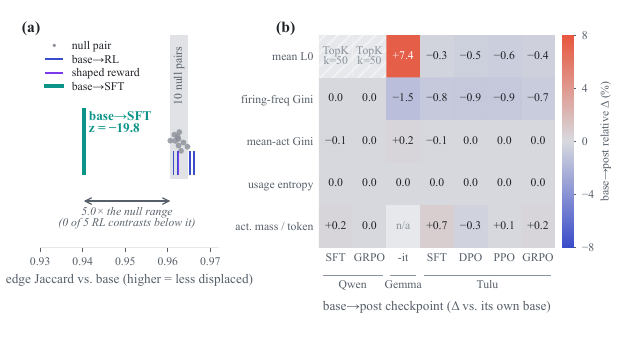}
\caption{\textbf{A run-variation reference and the diagnostic battery.} (a)~Base and run graphs built on one architecture (H200), the $10$ seed pairs of the five rebuildable GRPO-family runs, 0 of the five base-to-RL contrasts below the band, and base-to-SFT at $z=-19.8$, $5.0\times$ the null range beyond it; the shaped-reward arm is marked separately, and the archived six-run version is Table~\ref{tab:rewiring}. The SFT shift exceeds this empirical range; it does not establish the absence of a smaller RL effect. (b)~Level deltas against each checkpoint's own base from the archived metric files (descriptive, App.~\ref{app:build}); TopK-fixed and unavailable values are marked.}
\label{fig:seednull}
\end{figure}

\begin{figure}[t]
\centering
\includegraphics[width=1.0000\textwidth]{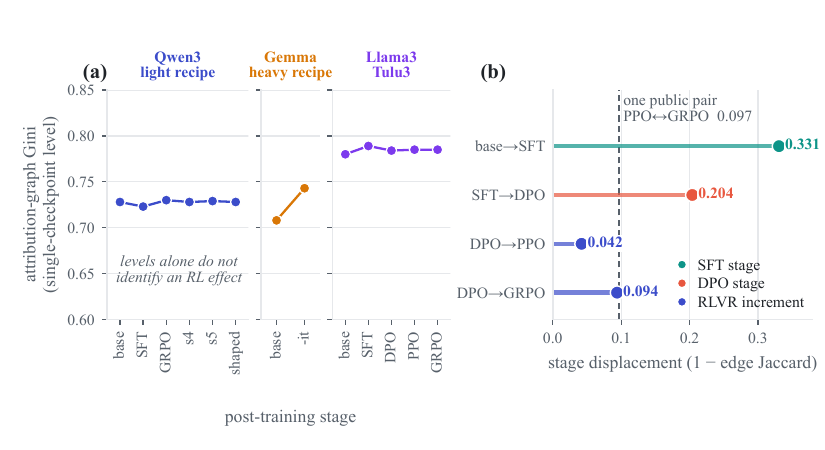}
\caption{\textbf{Checkpoint levels and stage increments answer different questions.} (a) Co-activation-graph Gini for Qwen3-1.7B light recipes, Gemma-2-2B heavy instruction tuning, and the Llama-3.1-8B Tulu-3 sequence. Levels alone do not identify an RL effect. (b) Stage displacement is 1 minus edge Jaccard. The dashed line is one public PPO/GRPO pair (0.097), not a same-recipe seed-null distribution. The RLVR increments are unresolved against that single reference; they are not certified nulls.}
\label{fig:staircase-data}
\end{figure}

The base control above concerns co-activation-graph \emph{concentration}, but the RL-internals literature reads RL as ``narrowing'' through an activation-side family of single-checkpoint metrics (features fire more selectively, usage sparsifies, entropy drops), our feature-level extrapolation of behavioral and policy-entropy results~\citep{yue2025,wang2025,cui2025entropy_mechanism}. We run the same base-vs-post control on a battery computed from SAE encodings ($n{=}500$, ${\sim}32$k tokens, dst-layer SAE): firing-frequency Gini, mean-activation Gini, and usage entropy (Table~\ref{tab:battery} and Fig.~\ref{fig:seednull}b, App.~\ref{app:h1extra}). \emph{A metric we must exclude, and why.} Mean $L_0$ --- the most commonly quoted ``sparsification'' statistic --- is \emph{uninformative by construction} on our headline Qwen dictionary: \texttt{qwen-scope-3-1.7b-base-w32k-l50} is a fixed-$k$ TopK SAE~\citep{gao2024scaling} with $k{=}50$, so its $L_0$ is a property of the dictionary, not of the model being probed. It reads exactly $50.0$ on all $20$ per-checkpoint Qwen receipts we computed --- base, SFT and five GRPO-family runs across three probe distributions (pile, GSM8K, BFCL train and validation) --- and its seed band is literally $[0,0]$: it \emph{cannot} move, so a null on it is not evidence of anything. We therefore drop mean $L_0$ from the Qwen battery and report it only where the SAE's sparsity is data-dependent and it can genuinely vary --- Gemma's JumpReLU dictionary and \texttt{Llama-Scope} (Sec.~\ref{sec:results-tulu}), where it does vary. On the metrics that can move, the result is, in both halves, \emph{negative for narrowing}: (i) on the light-RL trio every metric is base-dominated (firing-freq Gini $0.683$ throughout, $|\text{rel }\Delta|<1\%$); (ii) on the heavy-instruct Gemma pair the two metrics that move most shift \emph{opposite} to the prediction (JumpReLU mean $L_0$ \emph{rises} $184.8\to198.6$, firing-freq Gini \emph{falls} $0.494\to0.486$; mean-activation Gini and usage entropy move $\le0.002$). ``RL narrows the SAE basis'' is not observed in either regime.

\section{Extended related work}
\label{app:related}

\paragraph{The sharpest apparent conflict with our RL null.} \citet{rl_circuit_intensity2026} is the sharpest apparent conflict: in the same method family they find online RL raises activation intensity and diversity, while \emph{also} reporting DPO ``deviate[s] from these trends''; we do not read that as bearing on our DPO reading, whose band counts follow no dose ordering. Their readout is activation summaries, not a graph contrast against the algorithm's own band, and their RL is production-scale to our light GRPO; we report their positive as live for their regime and scope ours to what we null-calibrate (L1, L7).

\paragraph{Matched configuration across the SFT and DPO arms.} Both arms are LoRA runs, so the adapter is a candidate carrier: \citet{shuttleworth2024lorailluson} find LoRA adds ``intruder'' directions absent from full fine-tuning, and \citet{prasanth2026lorageometry} find LoRA dictionaries align weakly with the pretrained SAE basis our graphs use. Matched across the arms: base model, LoRA rank $16$/$\alpha\,32$, target modules (\texttt{q,k,v,o\_proj}), step count ($29$), effective batch ($32$), graph protocol, SAE and scoring code. Not matched, disclosed rather than buried: lr ($10\times$, already L8) and \texttt{lora\_dropout} (SFT $0.05$ vs.\ DPO $0.0$, inherited from the upstream scripts); the lr difference plausibly pushes \emph{toward} the observed gap (a smaller lr mechanically reduces DPO's
displacement), so the step-matched contrast is between the methods' standard recipes,
not an lr-matched isolation of the objective; neither changes adapter rank or placement, and \citet{prasanth2026lorageometry} find the divergence from the pretrained SAE basis relatively stable across LoRA ranks $4$--$32$, so rank is an unlikely carrier of the gap. The claim is about two objectives at matched adapter geometry and dose, and licenses nothing about full fine-tuning.

\paragraph{Attribution-graph interpretability.}
Attribution graphs --- canonically the circuit-tracing line of \citet{ameisen2025circuit,lindsey2025biology}, with activation-patching and causal-mediation precursors~\citep{heimersheim2024patching,vig2020causal} and the SAE-feature-level circuits of \citet{marks2024sparse_circuits} --- are widely used to extract feature panels; we use $10{,}000$ edges and the top-32 destination features. \citet{five_stage_causal_sae2026} show on GPT-2 IOI that the most \emph{selective} SAE features are not the most \emph{causal}, so we frame \Hone\ (our concentration measure; Sec.~\ref{sec:method}) as a reweighting / concentration finding about attribution mass, not a direct causal claim, and treat \Htwo\ activation deltas as a polarity probe.

\paragraph{The null-calibration gap in the RL-internals literature (full positioning).} \citet{shi2026rl_generalize} is the closest prior art: training SFT and RL from the same base on identical data (a strong design), they find RL induces \emph{restrained} feature changes that preserve base structure while SFT introduces many new features --- directionally consonant with our SFT-positive/RL-null result, but without asking whether the residual RL changes exceed seed noise. \citet{rojasnunez2026forgetting} reach the same ordering from a different direction entirely: their \emph{differential circuit vulnerability} is a head-level degradation measure, applied to \texttt{Qwen2.5-3B-Instruct} on scientific QA, and it finds SFT adapting faster while disrupting circuits substantially more than Dr.GRPO, which retains more of the base circuit. Three scope differences are worth naming, none of which we read as a conflict: their RL arm is \emph{serial} (Dr.GRPO refining an SFT policy) where our Qwen SFT and GRPO are parallel branches from base; their comparison contains no preference-optimization arm, so it says nothing about the SFT/DPO ordering that is our dose-matched result; and the unit is attention heads rather than sparse features. That two groups converge on the direction from designs sharing neither backbone, unit, nor comparator is the strongest kind of agreement available --- and it is agreement about a direction, which is a different thing from a width. Supplying the width is what the seed-null protocol here does, and it is why the SFT/DPO statement we lead with is a disjointness over four seeded runs per arm rather than a ratio between two runs. \citet{crosscoder_tooluse2026} localize RL-induced tool use to a compact crosscoder feature set on a matched base/RL pair, yet themselves flag the missing maximally-elicited-base control and decline the strong installation claim. \citet{spurious_rewards_paradox2026} give architecture controls (Llama, OLMo-2 negatives for a Qwen-specific shortcut mechanism); \citet{linear_dynamics_rlvr2026} show weights and output log-probabilities drift linearly under RLVR across configurations (consistency across runs, not a null for structure); \citet{rlvr_direction2026} find sparse directional token-level updates; \citet{sinha2026mechanistic} find six alignment algorithms leave different probe/SAE-anchor geometric traces across three families. \citet{neutral_mask2026}'s shallow-alignment finding (alignment training deactivates policy features while leaving latent partisan structure intact) is rhetorically adjacent, and our seed-null converts that style of claim into a calibrated one. On the elicitation side, \citet{venhoff2026base_reason,activation_control_cot2025,thinklogit2025} show reasoning behaviors can be turned on \emph{without weight updates}, and \citet{sal2025} show pretrained SAE-rule structure \emph{predicts} post-RLVR performance --- behavioral and predictive evidence to which our audit adds the missing measurement-side complement: a null-calibrated post-training diff in sparse feature space, on one full public staircase. Finally, the crosscoder-diffing line~\citep{minder2025crosscoders} shows naive L1 crosscoders \emph{fabricate} chat-specific latents (an artifact class our base-trained-SAE design avoids by construction), reinforcing that model-diffing claims need explicit nulls.

\paragraph{Tokenizer alignment, weight-space, and state-distribution accounts.}
Pre/post checkpoints can tokenize identical strings differently (e.g.\ added chat-template tokens), activating \emph{different} SAE features; we adopt explicit cross-tokenizer alignment. Our non-detection is compatible with weight- and state-space accounts: \citet{relex_rank1_rlvr2026} find a rank-1 approximation of the RLVR weight delta captures most of its gain; \citet{posttraining_states_not_tokens2026} reframe post-training as state-distribution reshaping; \citet{not_all_tokens_alike2026} find that a random $20\%$ of tokens preserves much of the full-token RL update and that low-attention-entropy ``anchor'' tokens carry its stable backbone --- the token-space counterpart of our feature-space concentration. \citet{select_improve2026} decompose RL into \emph{strategy selection} (from an existing repertoire) vs.\ \emph{improvement} (new patterns from hard data); our base-present concentration and absent entropy collapse are consistent with a \emph{selection}-dominant regime, on which reading novel-circuit signal should grow as RL data is made harder. At the behavior level \citet{kirk2024diversity} established the RLHF-vs-SFT contrast on generalization and output diversity, \citet{li2025thinking} show explicit chain-of-thought reasoning can \emph{degrade} instruction-following accuracy, and \citet{valueaxis2026} show DPO raises the model's internal value estimate, along a pre-existing expected-success axis, for rewarded behaviors --- a behavior-space instance of the ``post-training reweights a pre-existing direction'' picture our H1 measures; the recurring theme is that post-training moves behavior substantially even where representational change is small (our own data do not test this: the GRPO non-detection does not measure the footprint, and the behavioral and internal arms are on different state distributions; L5). The reward is itself a design axis separable from the RL update rule; our single shaped-$\lambda0.1$ run moves H3 slightly more ($d=-0.29$) than the accuracy-only baseline, within the spread of our probe-budget reruns (App.~\ref{app:h2h3}). The H1 hubs may also be \emph{parent} nodes in a sparse feature hierarchy; hierarchy-aware accounting is left for future work.

\section{Additional H1 material}
\label{app:h1extra}

\begin{table}[h]
\centering
\caption{\textbf{Headline control: concentration is base-set across training stages} (\texttt{Qwen3-1.7B}, \texttt{qwen-scope} src $14\to$ dst $22$, $n{=}500$, base-trained SAE). The zero-RL base already shows the full concentration; RL-attributable deltas are negligible and $31$--$32$ of the $32$ top hubs are shared (Jaccard $0.94$--$1.00$). An uncontrolled single-checkpoint analysis would have read the GRPO column alone as an RL signature.}
\label{tab:h1_basecontrol}
\small
\begin{tabular}{lrrr}
\toprule
Checkpoint & top-1\% (active) & Gini & \# active dst features \\
\midrule
Base (zero RL) & $0.427$ & $0.728$ & $2226$ \\
SFT            & $0.416$ & $0.723$ & $2229$ \\
GRPO           & $0.429$ & $0.730$ & $2225$ \\
\midrule
$\Delta$(GRPO$-$base) & $+0.002$ & $+0.002$ & --- \\
\multicolumn{4}{l}{Top-32 hub Jaccard: base$\cap$GRPO $0.94$, \; SFT$\cap$GRPO $1.00$, \; base$\cap$SFT $0.94$} \\
\bottomrule
\end{tabular}
\end{table}

\begin{table}[h]
\centering
\caption{\textbf{The base control replicates across $5$ Qwen recipes/seeds and a second family.} Width-robust deltas (base$\to$post) of the same co-activation-graph concentration. \emph{Within Qwen} (\texttt{qwen-scope}, $32$k, $n{=}500$): all five post-training runs give $|\Delta\text{Gini}|\le0.005$ and hub Jaccard $0.94$--$1.00$. \emph{Cross-family} (\texttt{Gemma-2-2B} $+$ \texttt{gemma-scope}, JumpReLU, $16$k, $n{=}500$): the zero-RL base already carries the concentration (Gini $0.708$); the larger \texttt{-it} shift reflects heavy production instruct-tuning, not light GRPO; with one heavy checkpoint and a possible chat-template confound (App.~\ref{app:limits}) we do not attribute it to intensity.}
\label{tab:h1_robust}
\small
\resizebox{\textwidth}{!}{%
\begin{tabular}{lllrrr}
\toprule
Family / SAE & Checkpoint & Post-training & $\Delta$Gini & $\Delta$top-1\% & hub Jaccard \\
\midrule
Qwen3-1.7B / qwen-scope & base   & --- (Gini $0.728$) & --- & --- & --- \\
 & SFT        & SFT (LoRA)              & $-0.005$ & $-0.011$ & $0.94$ \\
 & GRPO       & accuracy-only GRPO      & $+0.002$ & $+0.002$ & $0.94$ \\
 & GRPO-s4    & GRPO (seed)             & $\phantom{+}0.000$ & $+0.001$ & $1.00$ \\
 & GRPO-s5    & GRPO (seed)             & $+0.001$ & $+0.002$ & $1.00$ \\
 & shaped-$\lambda0.1$ & shaped-reward GRPO & $\phantom{+}0.000$ & $\phantom{+}0.000$ & $1.00$ \\
\midrule
Gemma-2-2B / gemma-scope & base & --- (Gini $0.708$) & --- & --- & --- \\
 & \texttt{-it} & heavy SFT+RLHF+distill & $+0.035$ & $+0.062$ & $0.73$ \\
\bottomrule
\end{tabular}%
}
\end{table}

\begin{table}[h]
\centering
\caption{\textbf{H1 edge-weight percentile distribution (left) and the six destination features with the largest total inbound weight (right)} (\texttt{Qwen3-1.7B} + \texttt{qwen-scope-3-1.7b}, src $14\to$ dst $22$). The two halves are independent summaries of the same $10{,}000$-edge graph; rows do not correspond across the divider.}
\label{tab:h1_weights}
\small
\begin{tabular}{lr|lrr}
\toprule
Percentile & Edge weight & Feature ID & Total inbound $w$ & In-degree \\
\midrule
$p_{25}$ & $14.3$ & 30329 & $71560$ & 2061 \\
median & $17.2$ & 26766 & $9453$ & 241 \\
$p_{75}$ & $23.9$ & 10420 & $3212$ & 88 \\
$p_{90}$ & $36.4$ & 25789 & $2537$ & 80 \\
$p_{99}$ & $112.2$ & 11799 & $1679$ & 48 \\
max & $3597$ & 23554 & $1109$ & 36 \\
\bottomrule
\end{tabular}
\end{table}

\paragraph{The dictionary-width artifact in full.} Independently of the base control, an earlier draft reported a ``$51.1\%\to84.0\%$'' jump from 1.7B to 8B as evidence that concentration intensifies with scale. Re-deriving both numbers from the committed graphs shows \emph{three} confounds stacked, not one. (i) \emph{Provenance}: the $51.1\%$ endpoint is not a 1.7B number at all --- it is the top-$1\%$-of-active share of the v0 \texttt{Pythia-70m} pilot graph ($0.5111$, \texttt{notes/attribution\_graph\_v0.json}), a different backbone and SAE family. (ii) \emph{Denominator}: the $84.0\%$ endpoint is top-$1\%$-of-\emph{dictionary} ($0.8395$ on the 8B graph), so the two endpoints did not share a denominator either. (iii) \emph{Width}: with the denominator held fixed at top-$1\%$-of-$d_{\text{SAE}}$, the honest 1.7B$\to$8B pair is $0.7110\to0.8395$, and the remaining rise is mechanical --- $1\%$ of the $2\times$-wider 8B dictionary is $655$ features against $328$. Under any width-independent statistic the effect reverses or vanishes. We record the corrected numbers rather than quietly deleting the claim: the failure mode being illustrated is exactly a number whose provenance was never re-checked. We measure concentration at matched relative depth ($\text{src}\approx0.5L$, $\text{dst}\approx0.79L$, $n{=}500$, $10{,}000$ edges) across two SAE families and $0.6$--$14$B (Table~\ref{tab:h1_scale}): in the \texttt{qwen-scope} residual family the Gini of inbound weight is $0.728$ at 1.7B and $0.719$ at 8B; in the circuit-tracer transcoder family it is $0.688$/$0.654$/$0.645$ at $0.6$/$1.7$/$14$B. Both families are flat-to-slightly-decreasing across a $23\times$ parameter range---this sparse sweep shows no monotonic increase of concentration with scale.

\begin{table}[h]
\centering
\caption{\textbf{H1 concentration is scale-stable.} Top-1\% (of active features) inbound-weight share and Gini at matched relative depth ($n{=}500$, $10{,}000$ edges), across two SAE families and $0.6$--$14$B. Both families are flat-to-slightly-decreasing with scale; the dictionary-relative ``top-1\%-of-$d_{\text{SAE}}$'' (last column) rises only because wider dictionaries place more features in a fixed $1\%$. The \texttt{qwen-scope}~\citep{deng2026qwenscope} rows come from a committed receipt; the transcoder rows (circuit-tracer Qwen3 transcoders~\citep{hanna2025circuittracer}) come from the sweep job scripts (no committed per-run summary).}
\label{tab:h1_scale}
\small
\begin{tabular}{llrrr}
\toprule
SAE family & Model & top-1\% (active) & Gini & top-1\% (dict) \\
\midrule
qwen-scope (resid)   & Qwen3-1.7B & $42.7\%$ & $0.728$ & $71.1\%$ \\
qwen-scope (resid)   & Qwen3-8B   & $35.2\%$ & $0.719$ & $84.0\%$ \\
transcoder (circuit-tracer) & Qwen3-0.6B & $35.9\%$ & $0.688$ & --- \\
transcoder (circuit-tracer) & Qwen3-1.7B & $31.1\%$ & $0.654$ & --- \\
transcoder (circuit-tracer) & Qwen3-14B  & $32.6\%$ & $0.645$ & --- \\
\bottomrule
\end{tabular}
\end{table}

\begin{table}[h]
\centering
\caption{\textbf{Activation-side ``narrowing'' metrics are base-dominated (light RL) or move \emph{against} narrowing (heavy instruct).} Base$\to$post deltas of four common single-checkpoint SAE metrics ($n{=}500$, dst-layer SAE). \emph{Qwen3-1.7B} (light GRPO, \texttt{qwen-scope} l22): all metrics base-dominated, $|\text{rel }\Delta|<1\%$. \emph{Gemma-2-2B} (heavy SFT+RLHF+distill, \texttt{gemma-scope} l20): mean $L_0$ \emph{rises} (less sparse) and firing-freq Gini \emph{falls} (less concentrated) --- the opposite of the predicted RL ``narrowing''. Narrowing arrow shows the direction RL would move each metric under the entropy-collapse/narrowing hypothesis. \emph{The mean-$L_0$ row is informative only in the Gemma columns}: \texttt{qwen-scope} is a fixed-$k{=}50$ TopK dictionary, so its $L_0$ is pinned by construction and its Qwen entries (greyed) are a property of the SAE, not a measurement of the model; Gemma's JumpReLU $L_0$ is data-dependent and does move.}
\label{tab:battery}
\small
\resizebox{\textwidth}{!}{%
\begin{tabular}{lcrrrrr}
\toprule
 & narrowing & \multicolumn{3}{c}{Qwen3-1.7B / qwen-scope l22} & \multicolumn{2}{c}{Gemma-2-2B / gemma-scope l20} \\
\cmidrule(lr){3-5}\cmidrule(lr){6-7}
Metric & predicts & base & $\Delta$SFT & $\Delta$GRPO & base & $\Delta$\texttt{-it} \\
\midrule
mean $L_0$            & $\downarrow$ & \textcolor{gray}{$50.0^{\dagger}$} & \textcolor{gray}{$+0.0^{\dagger}$} & \textcolor{gray}{$+0.0^{\dagger}$} & $184.8$ & $+13.8$ \\
firing-freq Gini      & $\uparrow$   & $0.683$ & $-0.000$ & $+0.000$ & $0.494$ & $-0.007$ \\
mean-act Gini         & $\uparrow$   & $0.733$ & $-0.001$ & $+0.000$ & $0.643$ & $+0.002$ \\
usage entropy (norm)  & $\downarrow$ & $0.895$ & $+0.000$ & $-0.000$ & $0.950$ & $-0.000$ \\
\bottomrule
\multicolumn{7}{l}{\footnotesize $^{\dagger}$ structurally pinned: TopK dictionary with $k{=}50$; excluded from all Qwen metric counts.}
\end{tabular}%
}
\end{table}

\paragraph{The full seven-metric seed-null battery, reported descriptively.} \emph{[archived build]} Applying the differencing framework to the informative part of the activation-side battery (five GRPO-family runs re-measured per checkpoint, pile probe set): for all \emph{seven} metrics that can vary --- dead fraction, firing-frequency Gini and top-$1\%$ share, mean-activation Gini, normalized usage entropy, and two activation-\emph{intensity} levels (total activation mass per token; mean activation per firing event) --- the base$\to$GRPO delta lies within the GRPO-seed pairwise band (e.g.\ activation mass per token: $|\Delta_{\text{base}\to\text{GRPO}}|=0.31$ vs.\ seed band $[0,3.09]$; mean act per firing: $0.006$ vs.\ $[0,0.062]$). The receipt \texttt{notes/featstats\_seednull.json} carries an eighth row, mean $L_0$, at $50.0$ with a $[0,0]$ band; we exclude it from the count because the TopK dictionary pins it (Sec.~\ref{sec:results-battery}) --- it is not a null result, it is a constant. Three of the seven retained metrics are themselves close to the precision floor (Sec.~\ref{sec:method}, ``a statistic must beat its own reporting precision''): firing-freq top-$1\%$ and mean-activation Gini have $|\Delta_{\text{base}\to\text{GRPO}}|$ equal to the seed band maximum at $0.0002$, one unit of the reported precision, and normalized usage entropy moves $0.0001$ against $0.0002$; for these the honest statement is ``no movement resolvable at our precision.'' \emph{Build check (registered, branch B3).} The archived metric files mix architectures (base and GRPO on A100, s1--s5 on L40S), and none of the five that can be recomputed reproduces at stored precision on H200 (\texttt{notes/PREREG\_graph\_build\_audit\_20261004.md} \S4). We therefore report the battery descriptively and do not count band membership as evidence. Recomputed on one architecture (four runs s1, s3, s4, s5; the GRPO baseline checkpoint is deleted, so base$\to$s$_i$ deltas replace base$\to$GRPO), the base$\to$s$_i$ delta lies within the four-run max-pairwise band for $4/4$ runs on dead fraction, firing-frequency Gini, mean-activation Gini and normalized usage entropy, $3/4$ on firing-frequency top-$1\%$, and $2/4$ on activation mass per token and mean activation when active (\texttt{results/graph\_build\_audit\_20261005/run/battery\_same\_env\_descriptive.json}); these are counts against the largest of six pairwise differences, not tests. We do not claim \citet{rl_circuit_intensity2026}'s intensity result is wrong at their heavier operating point --- our Gemma arm shows heavy post-training does move activation statistics --- but the contrast shows their headline does not generalize downward to light GRPO.

\begin{table}[h]
\centering
\caption{\textbf{[Archived build.] Co-activation displacement is within run noise for five of six RL contrasts (s1 falls $0.2\times$ the null range below the band, vs SFT's $4.0\times$); SFT clears the null at $z=-15.9$.} The base and several runs here were built on different GPU architectures; on one architecture none of the five rebuildable GRPO contrasts is outside its band (Table~\ref{tab:main}). Pairwise graph similarity (\texttt{Qwen3-1.7B}, l14$\to$22, $10{,}000$ edges) vs.\ the $15$-pair GRPO-family null. Negative $z$ = more displaced than seed noise; only SFT separates by a large margin, while s1 falls marginally below the band at step $200$ ($z=-2.0$); at matched step $300$ it is inside the $15$-pair band ($z=-0.98$) and $0.0006$ below the three-pair same-step band.}
\label{tab:rewiring}
\small
\begin{tabular}{lrrr}
\toprule
Pair & edge Jaccard & $z$ vs.\ seed null & rank$^{\S}$ \\
\midrule
base$\to$SFT (LoRA)   & $\mathbf{0.937}$ & $\mathbf{-15.9}$ & $\le1/16$ \\
base$\to$GRPO         & $0.962$ & $-0.5$ & $0.38$ \\
base$\to$GRPO s1/s3/s4/s5 & $0.959$--$0.964$ & $-2.0$ to $+0.8$ & $0.06$--$0.81$ \\
base$\to$shaped-$\lambda0.1$ & $0.964$ & $+0.8$ & $0.81$ \\
\midrule
\midrule
base$\to$SFT vs.\ \emph{SFT's own} null & $\mathbf{0.937}$ & $\mathbf{-19.9}$ & $\le1/7$ \\
base$\to$SFT s0/s1/s2/s3 (own null) & $0.937$--$0.941$ & $-20.0$ to $-17.0$ & $\le1/7$ \\
\midrule
\multicolumn{4}{l}{\emph{Seed-noise null}: 6 GRPO-family runs, $15$ pairwise; edge Jaccard $0.9626\pm0.0016$, cosine $0.999$.} \\
\multicolumn{4}{l}{\emph{SFT seed null}: 4 LoRA-SFT runs, $6$ pairwise; edge Jaccard $0.9644\pm0.0014$, band $[0.9619,0.9656]$.} \\
\multicolumn{4}{l}{\footnotesize $^{\S}$rank quantile among the $m{+}1$ values; a tail probability only if the contrast is exchangeable with} \\
\multicolumn{4}{l}{\footnotesize the seed pairs, which run exchangeability does not imply (Proposition~\ref{prop:band}: attainable bound $1/k$).} \\
\bottomrule
\end{tabular}
\end{table}

\paragraph{GSM8K domain robustness in full.} A natural objection is that RL would narrow only the features \emph{relevant to its task}, so a single probe distribution could miss a localized narrowing. We therefore re-run the entire battery on the clean Qwen base/SFT/GRPO trio with a second, structurally different probe set --- GSM8K question$+$solution chains ($n{=}500$, ${\sim}32$k tokens). \emph{An honest scope note:} the RL task here is BFCL function calling (Sec.~\ref{sec:method}), so GSM8K is \emph{not} the RL training distribution --- it is a second off-training-distribution probe that engages a structured, step-by-step-reasoning region of the feature basis, and the \emph{on}-distribution (function-calling) probe is the on-task BFCL audit of App.~\ref{app:ontask}. The GSM8K distribution is indeed far more concentrated than pile (base firing-frequency Gini $0.840$ vs.\ $0.683$; mean-activation Gini $0.883$ vs.\ $0.733$; dead fraction $0.36$ vs.\ $0.15$), confirming it genuinely engages a different region of the basis. Yet the base$\to$post deltas remain nil: firing-frequency Gini is $0.840$ throughout ($\Delta\le0.0004$), and every base$\to$GRPO metric has $|\text{rel }\Delta|<1\%$ (the four headline narrowing metrics are $<1\%$ at every stage; SFT's dead fraction rises $+4.7\%$ and its firing top-$1\%$ falls $-1.5\%$, the largest GSM8K moves). The narrowing-metric null is therefore not an artifact of one probe distribution: it holds across two probe domains with very different feature-usage profiles; the probe on the RL task distribution itself is App.~\ref{app:ontask}.

\section{On-task (BFCL) arm and its minimum detectable effect}
\label{app:ontask}

\paragraph{Protocol.} Same estimator, checkpoints, and base-trained SAE as the headline audit (Sec.~\ref{sec:method}); only the probe states change: the RL task's own function-calling prompts (the verl BFCL training parquet, $n{=}500$ prompts for graphs and battery; the $n{=}100$ held-out validation split for the val arm), rendered through the checkpoint tokenizer's chat template with the generation prompt appended, sequence length $1024$ keeping the \emph{last} tokens --- so the graph's last-position state is the function-call decision point, the exact input state verl used for rollouts. The base-trained SAE remains a usable lens on these states (FVE $0.81$--$0.84$). Co-activation graphs (l14$\to$22, $10{,}000$ edges) are built for base, SFT, and the same six GRPO-family runs as the pile arm, giving the same $15$ seed-null pairs.

\paragraph{Outcomes, reported descriptively (Tables~\ref{tab:ontask_rewiring}, \ref{tab:ontask_battery}).} \emph{Displacement:} base$\to$GRPO edge Jaccard $0.746$, $z=-0.23$, inside the seed band $[0.667,0.819]$; the other five base$\to$RL contrasts fall at $z\in[-0.87,+1.83]$, none below the band's more-displaced lower edge (base$\to$s4, $0.829$, sits \emph{above} the band's upper edge at $z=+1.83$; caveat iii). \emph{Levels:} against the base's Gini $0.755$ (top-$1\%$-active $0.217$), all seven post checkpoints have $|\Delta\text{Gini}|\le0.026$ with top-$32$ hub Jaccard $0.829$--$0.882$. \emph{Battery:} the base$\to$GRPO delta is within the five-run seed-pair band on all eight informative metrics (the ninth, mean $L_0$, is TopK-pinned at $50.0$ with a $[0,0]$ band and is excluded, Sec.~\ref{sec:results-battery}); the validation-split trio is flat (firing-freq Gini $0.8532/0.8529/0.8541$ at base/SFT/GRPO over $41{,}612$ identical tokens; FVE $0.81$--$0.84$).

\paragraph{Why the arm is uninformative, quantified.} (i) The on-task seed-noise band is ${\sim}25\times$ wider than pile's (SD $0.0403$ vs.\ $0.0016$; range $0.152$ vs.\ $0.006$): task states are intrinsically higher-variance across runs, so the $2\sigma$ minimum detectable effect is $\Delta\text{Jaccard}=0.0806$ on-task ($10.7\%$ of the statistic) vs.\ $0.0032$ on pile. SFT's actual pile-arm effect is $0.0254$ ($2.6\%$ of that statistic): the on-task arm can only detect effects ${\sim}3\times$ larger in absolute Jaccard (${\sim}4\times$ relative to each arm's statistic) than the one the audit's positive control produces --- it is structurally incapable of adjudicating an effect of the size at issue. (ii) Consistently, the SFT positive control does \emph{not} separate on-task: base$\to$SFT $0.712$, $z=-1.06$ (pile: $z=-15.9$, four times outside the null range). This is mechanistically consistent with the pile localization (App.~\ref{app:localization}) --- SFT's detected displacement prunes boilerplate/function-word edges into the generic-syntax hub, wiring that BFCL's structured task prompts barely exercise --- but it means the on-task arm carries no internal demonstration that it could detect a real signature. (iii) The arm's null band is itself contaminated: base$\to$s4 lands at $z=+1.83$, \emph{above} the band's upper edge --- a base$\to$RL contrast more base-like than the seed pairs are to each other, reflecting the step/recipe mixture of Limitation L7 --- and the hub-Jaccard statistic takes only three distinct values across the $15$ null pairs, too quantized to score. We therefore class the on-task arm as uninformative at the relevant effect size and draw no conclusion from it in either direction; the RL-null claim rests on the pile arm.

\begin{table}[h]
\centering
\caption{\textbf{On-task (BFCL) seed-noise-null displacement vs.\ the pile arm} (\texttt{Qwen3-1.7B}, l14$\to$22, $10{,}000$ edges, edge Jaccard). Reported for completeness, not as a null: the on-task arm's $2\sigma$ MDE ($0.0806$) is ${\sim}3\times$ SFT's actual pile-arm effect ($0.0254$; ${\sim}4\times$ in relative terms), and SFT does not separate on it ($z=-1.06$), so no conclusion is drawn from this table (caveats i--iii).}
\label{tab:ontask_rewiring}
\small
\begin{tabular}{lrr}
\toprule
 & BFCL task states & pile \\
\midrule
seed null, $15$ pairs (mean $\pm$ SD) & $0.7551\pm0.0403$ & $0.9626\pm0.0016$ \\
seed band $[\min,\max]$ & $[0.667, 0.819]$ & $[0.961, 0.966]$ \\
base$\to$GRPO ($z$) & $0.746$ ($-0.23$) & $0.962$ ($-0.5$) \\
base$\to$s1/s3/s4/s5/shaped ($z$ range) & $-0.87$ to $+1.83$ & $-2.0$ to $+0.8$ \\
base$\to$SFT ($z$) & $0.712$ ($-1.06$) & $\mathbf{0.937}$ ($\mathbf{-15.9}$) \\
\bottomrule
\end{tabular}
\end{table}

\begin{table}[h]
\centering
\caption{\textbf{On-task (BFCL) levels and battery.} Left: per-checkpoint graph levels vs.\ base (Gini $0.755$, top-$1\%$-active $0.217$). Right: base$\to$GRPO battery deltas vs.\ the five-run GRPO-family seed-pair band (train split, $n{=}500$); all eight informative metrics within seed noise. The mean-$L_0$ row (greyed) is TopK-pinned at $50.0$ with a $[0,0]$ band and is excluded from the count.}
\label{tab:ontask_battery}
\small
\begin{tabular}{lrr|lrr}
\toprule
Checkpoint & $\Delta$Gini & hub Jacc & Battery metric & $|\Delta_{\text{base}\to\text{GRPO}}|$ & seed max \\
\midrule
SFT     & $-0.026$ & $0.829$ & \textcolor{gray}{mean $L_0$ (pinned)} & \textcolor{gray}{$0.0$} & \textcolor{gray}{$0.0$} \\
GRPO    & $+0.007$ & $0.882$ & dead fraction & $0.0046$ & $0.0070$ \\
s1      & $-0.002$ & $0.882$ & firing-freq Gini & $0.0003$ & $0.0004$ \\
s3      & $+0.006$ & $0.882$ & firing-freq top-$1\%$ & $0.0040$ & $0.0051$ \\
s4      & $+0.013$ & $0.882$ & mean-act Gini & $0.0010$ & $0.0012$ \\
s5      & $+0.017$ & $0.829$ & usage entropy (norm) & $0.0011$ & $0.0012$ \\
shaped-$\lambda0.1$ & $+0.026$ & $0.829$ & act mass / token & $16.9$ & $21.4$ \\
        &          &         & mean act when active & $0.34$ & $0.43$ \\
        &          &         & FVU & $0.0039$ & $0.0069$ \\
\bottomrule
\end{tabular}
\end{table}

\paragraph{Receipts.} \texttt{notes/circuit\_rewiring\_bfcl.json} (displacement + seed null), \texttt{notes/basediff\_bfcl\_levels.json} (levels), \texttt{notes/featstats\_seednull\_bfcl.json} (battery), computed from \texttt{notes/bfcl\_graph\_*\_l14to22.json} ($8$ checkpoints) and \texttt{notes/bfcl\_featstats\_*\_l22.json} / \texttt{*\_l22\_val.json}.

\section{Localization of the SFT signature}
\label{app:localization}

\paragraph{Pruned inbound wiring at the dominant hub.} We define the GRPO-family \emph{consensus} edge set (edges present in all $6$ GRPO-family graphs; $9{,}584$ of $10{,}000$) and isolate edges that base and every GRPO run retain but SFT drops from its top-$10$k graph ($138$ edges, $0.78\%$ of base edge weight). The drops are strikingly localized: $86$ of $138$ ($63\%$ of dropped weight) are inbound edges to a single destination feature --- $\#30329$, the rank-$1$ concentration hub itself (Table~\ref{tab:h1_weights}) --- and the top-$8$ destination features hold $71.7\%$ of all dropped weight. SFT-\emph{added} edges (in SFT, absent from base and all GRPO runs; $124$ edges) are by contrast diffuse (top-$8$ share $24\%$). So the SFT signature is a \emph{localized pruning of inbound wiring at the dominant pretraining hub} ($\approx4\%$ of the hub's in-degree pushed below the graph cutoff) --- consistent with SFT being the only checkpoint whose concentration moves \emph{down} (top-$1\%$ $0.427\to0.416$, Table~\ref{tab:h1_basecontrol}) --- while GRPO leaves the hub's wiring intact. (Caveat: ``dropped'' means fell below the top-$10$k edge cutoff, i.e.\ a real relative weakening, not necessarily to zero.)

\paragraph{What the displaced features are} (max-activating contexts over the pile probe set, base-model encodings). The pruned hub $\#30329$ is a generic \emph{numeric / whitespace / formatting} feature (top activations on digits and leading whitespace before numbers --- ``\texttt{estimated~}\underline{257}'', ``\texttt{93}\underline{5}\texttt{60}'', ``\texttt{ORS~}\underline{192}'' --- and URL/markup tokens; fires on $62\%$ of tokens, mean activation $183$, by far the largest in the graph). The two source features carrying the most dropped weight are likewise high-frequency and generic --- each loses $16$ edges to \emph{other} destinations, while its own edge into $\#30329$ survives in the SFT graph at reduced weight ($3076\to2858$, $3597\to3388$): $\#20227$ fires on boilerplate/function tokens (`` by'', ``:'', `` function'', cookie-notice and code-scaffold text; $55\%$ firing rate) and $\#16308$ is near-ubiquitous ($98\%$ firing rate). So the concrete picture is: \emph{SFT weakens the co-activation edges by which ubiquitous boilerplate/function-word features feed the base model's dominant generic-syntax hub} --- a plausible footprint of instruction-tuning learning to down-weight boilerplate --- and GRPO leaves exactly these edges in place. This turns the audit from a yes/no detector into a lens: when a post-training stage \emph{does} leave a signature, the same machinery names the features it acts on.

\paragraph{Does the displacement carry computation?} The registered causal test of the displaced sources, its pilot and the full sweep are in App.~\ref{app:functional}.

\paragraph{A causal probe keeps this claim honest --- and bounds it.} We tested whether the co-activation edge corresponds to a causal pathway by ablating the source features' decoded contributions at the source layer (subtracting $z_f W_{\text{dec}[f]}$ from the residual, reconstruction error left intact) and reading the hub's activation downstream ($n{=}200$ prompts). The interaction is real and feature-specific --- ablation shifts the hub $3.8$--$15.8\times$ more than a count-matched random-feature control --- but it is \emph{suppressive} (removing the sources \emph{raises} the hub by $4.9$--$5.7\%$; the co-activation edge weight, positive by construction on non-negative codes, is sign-blind), and, decisively, its causal strength is \emph{not weaker in SFT} (hub rise $+5.7\%$) than in base ($+4.9\%$). So the SFT edge-drop that the audit detects is a change in \emph{co-activation statistics}, not (by this probe) a pruned causal coupling. We therefore keep the mechanism claim strictly correlational, and flag the general lesson: co-activation-graph edges can move without the underlying causal interaction moving, one more reason signature claims need causal probes before mechanistic interpretation. (Caveats: the random control is count- but not magnitude-matched, so the specificity ratios are upper bounds; output-level KL under ablation is small, $0.010$--$0.013$.)

\section{The registered functional-transfer sweep}
\label{app:functional}

\paragraph{What was registered.} \citet{bayatmakou2026manycircuits} show that structurally distinct circuits can implement the same computation, so structural displacement, at any number of standard deviations, does not identify a changed mechanism. Before running anything we registered the test they prescribe (\texttt{PREREGISTRATION\_functional\_transfer.md}: faithfulness gate first, branches F1--F4) and, after a power pre-flight showed that the base and SFT top-$k$ graphs share ${\sim}94\%$ of their source features, amended it to the \emph{differential} sources. $D_{\mathrm{SFT}}$ ($D_{\mathrm{base}}$) is the set of source features of edges in the SFT (base) top-$k$ graph and not the other, minus sources in both. For model $M$ and set $S$, $\Delta(M,S)$ is the mean final-token $\mathrm{KL}(p_M\,\|\,p_M^{S})$ when, at layer $14$, the SAE contributions $z_f W_{\mathrm{dec}}[f]$ of $f\in S$ are replaced by those of a resampled prompt (reconstruction error untouched). The gate asks each own-condition set to beat a size- and firing-frequency-matched random set $R$ in its own model; the decisive statistic is the interaction (not the edge Jaccard of Sec.~\ref{sec:method})
$J=[\Delta(\mathrm{SFT},D_{\mathrm{SFT}})-\Delta(\mathrm{SFT},D_{\mathrm{base}})]-[\Delta(\mathrm{base},D_{\mathrm{SFT}})-\Delta(\mathrm{base},D_{\mathrm{base}})]$, with a paired-prompt bootstrap CI and a within-prompt label-permutation $p$ ($B=10{,}000$ each). $J>0$ would mean the SFT-specific sources matter more in SFT than in base, relative to the base-specific ones, so a main effect of ``these features matter everywhere'' cancels.

\paragraph{The pilot, and four defects found before the sweep.} A stop-gated pilot ($k=5{,}000$, $100$ prompts) passed the gate ($D_{\mathrm{SFT}}$ moved the output $0.550$/$0.524$ nats in base/SFT, $65\times$/$90\times$ its random control) and gave $J=-0.027$ (CI $[-0.055,-0.001]$, $p=0.050$). An earlier draft read it as the causal result. Re-reading its inputs for the sweep found four defects. (i) The committed reader ended in a catch-all \texttt{else} that labelled every non-F1 outcome F2, although F2 requires a CI covering zero; the reader now has an exhaustive table (adding F5, reversed: $J<0$, CI below zero, $p<0.05$; and U, discordant: the CI and the permutation test disagree), and under it the pilot reads U. (ii) $D_{\mathrm{SFT}}$ came from the audited SFT run's graph, but the pilot's SFT arm loaded a \emph{different} seeded SFT run's adapter. (iii) The random control was frequency-matched on one $64$-token prompt, so features silent there were tied. (iv) The model was loaded with weight processing, unlike the graph builder. The correction (\texttt{notes/PREREG\_functional\_transfer\_full\_20260924.md}) was committed before any sweep receipt existed. It fixes the pairing (configuration A, the audited graph with its own adapter, is the only one read for the verdict), frequency matching (firing rate over the $500$ graph-building prompts, draws within $\pm200$ ranks, without replacement), the held-out set (every row $5{,}000$--$9{,}999$ of \texttt{pile-10k} with $\ge64$ tokens, $n=4{,}816$, count committed before the main job) and the rule across budgets (one branch at all three $k$, otherwise MIXED).

\begin{table}[h]
\centering
\small
\setlength{\tabcolsep}{2.8pt}
\caption{\textbf{Registered sweep ($n=4{,}816$ held-out prompts; sweep verdict: \textbf{MIXED}).} A is the audited SFT graph with its own adapter (the only configuration read for the verdict); $^{\dagger}$descriptive: B is a second matched SFT run (seed 0), C the pilot's mismatched pairing at full $n$. $\Delta$: $\Delta(\mathrm{SFT},D_{\mathrm{SFT}})$ in nats; ratios are mean $\Delta$ over the matched random control in the SFT cell ($D_{\mathrm{SFT}}$, SFT model) and the base cell ($D_{\mathrm{base}}$, base model); a base-cell ratio below $1$ fails the gate (F3), and then $J$ is not read.}
\label{tab:functional}
\begin{tabular}{llcccccccc}
\toprule
cfg & $k$ & $|D_{\mathrm{SFT}}|/|D_{\mathrm{base}}|$ & $\Delta$ & SFT cell & base cell & $J$ & $95\%$ CI & $p$ & branch \\
\midrule
A & $2{,}000$ & $55/58$ & $0.259$ & $37\times$ & $0.3\times$ & --- & --- & --- & F3 \\
A & $5{,}000$ & $104/97$ & $0.549$ & $22\times$ & $2.0\times$ & $-0.0271$ & $[-0.0326, -0.0217]$ & $0.0001$ & F5 \\
A & $10{,}000$ & $213/206$ & $0.373$ & $12\times$ & $7.9\times$ & $-0.0164$ & $[-0.0217, -0.0113]$ & $0.0001$ & F5 \\
\midrule
B$^{\dagger}$ & $2{,}000$ & $50/57$ & $0.272$ & $27\times$ & $0.5\times$ & --- & --- & --- & F3 \\
B$^{\dagger}$ & $5{,}000$ & $116/106$ & $0.610$ & $14\times$ & $0.5\times$ & --- & --- & --- & F3 \\
B$^{\dagger}$ & $10{,}000$ & $225/212$ & $0.448$ & $6.5\times$ & $5.5\times$ & $-0.0221$ & $[-0.0278, -0.0165]$ & $0.0001$ & F5 \\
C$^{\dagger}$ & $5{,}000$ & $104/97$ & $0.548$ & $22\times$ & $2.0\times$ & $-0.0279$ & $[-0.0336, -0.0225]$ & $0.0001$ & F5 \\
\bottomrule
\end{tabular}
\end{table}

\paragraph{Result.} The SFT-specific sources are load-bearing at every budget and in both models, $12$--$37\times$ their random control in SFT (Table~\ref{tab:functional}). At $10{,}000$ edges the ratio of medians in the SFT cell is $3.5\times$ (the table reports ratios of means). Where both gates pass, the interaction is \emph{negative} and resolved: relative to the base-specific sources, the SFT-specific ones matter slightly less in the SFT model than in the base. At $2{,}000$ edges the $58$ base-specific sources move the base model's output less than their random control, so that budget is F3. The registered rule therefore returns MIXED (F3, F5, F5), and no stage-specificity claim is made in either direction. The second matched SFT run (B) fails the base-cell gate at $2{,}000$ and $5{,}000$ edges and has the same negative sign at $10{,}000$. The pilot's pairing at full $n$ (C) lies within $0.001$ nats of A in every cell ($J=-0.0279$ vs.\ $-0.0271$), so the pairing defect did not create the sign. The drop from the pilot's $65\times$ lies in the control: at $k=5{,}000$ the base-model random control rises from $0.008$ to $0.027$ nats with the corrected frequency matching, loader and prompt set (their shares are not separated), while $\Delta(\mathrm{base},D_{\mathrm{SFT}})$ moves from $0.550$ to $0.581$. \emph{Wording deviation, declared.} For a MIXED verdict the correction's pre-written sentence was ``found the displaced sources load-bearing $\ldots$ but did not resolve stage specificity''. Because two of the three primary budgets are resolved F5 ($p=0.0001$), we state the direction instead (abstract, Introduction, Sec.~\ref{sec:results-rewiring}, Discussion, Conclusion); the claim that follows is unchanged. The verb stays \emph{displaces} under every registered verdict: this source-side intervention corroborates structure but does not validate the co-activation object as an executable circuit.

\section{Tulu operating-point robustness and SAE-family caveats}
\label{app:tulu_robust}

\texttt{Llama-Scope} SAEs are dataset-wise normalized: per-layer average activation norms are $12.1$/$31.6$ at l16/l25, and the JumpReLU threshold lives in normalized space, so a direct \texttt{sae\_lens} \texttt{encode()} call runs unnormalized unless the scale is folded into the weights. We therefore ran the full staircase twice, unnormalized (v1) and normalization-folded (v2). Every conclusion is operating-point robust. The post-SFT stage increments agree to within $0.001$ across the two runs (SFT$\to$DPO $0.796$, DPO$\to$PPO $0.958$, DPO$\to$GRPO $0.906$, PPO$\leftrightarrow$GRPO $0.903$); the base$\to$SFT increment, the largest-displacement and most operating-point-sensitive stage, moves from $0.669$ (v1) to $0.697$ (v2) --- still by far the dominant displacement stage, so the qualitative conclusion is unchanged; the DPO$\to$PPO increment is the one that straddles a rounding boundary (v1 $0.9579$ / v2 $0.9587$), and we quote the v1 value $0.958$ to match the main table; the run-pair null PPO$\leftrightarrow$GRPO is $0.9029$ (v1) / $0.9033$ (v2), reported as $0.903$; at the v2 point level deltas stay $\le0.019$ against a $0.815$ base Gini (v1: $\le0.009$ against $0.780$, Table~\ref{tab:tulu}); the battery verdict is unchanged. The Llama-3.1-8B LoRA-SFT replication (App.~\ref{app:extended}) was run at the normalization-folded (v2) point. Two caveats are grouped here, with full treatment in Limitations. First, \texttt{Llama-Scope} runs far denser than \texttt{qwen-scope} (measured $L_0\approx1.5$k of $131$k at the normalized point, vs.\ $50$), so absolute $L_0$ does not compare across SAE families; we only ever use within-family deltas. Second, we exclude reconstruction-error (FVU) numbers for \texttt{Llama-Scope}: its \texttt{sae\_lens} \emph{decode} path returns implausible error (${\sim}10{,}000\times$ variance, against the ${\sim}90\%$ FVE its authors report), indicating a loader inconsistency, and everything in this paper uses \texttt{encode} only.

\section{H2 / H3 auxiliary-null detail (v0 Pythia-70m panel)}
\label{app:h2h3}

\paragraph{H3 in full: non-collapse, consistent with a barely-perturbed basis.}
H3 is a \emph{binary} non-collapse test: does the per-feature entropy delta clear the $|d|=0.5$ collapse gate (a $|d|=0.5$ threshold we pre-registered ourselves, following Cohen's conventional medium effect; \citet{yue2025,wang2025} report behavioral results and state no feature-level gate, so this bar is ours to defend, not theirs)? Across the committed accuracy-only GRPO runs the answer is \emph{no}: the cross-tokenizer run (24 probes $\times$ 4 continuations, \texttt{h3\_entropy\_cross\_tokenizer\_v0.json}) gives $\Delta H=-0.004$, $d=-0.005$; re-measuring the same baseline checkpoint at three probe${\times}$continuation budgets gives $\Delta H=+0.018$, $d=+0.26$ ($24{\times}4$), $\Delta H=+0.043$, $d=+0.32$ ($24{\times}1$), and $\Delta H=-0.020$, $d=-0.19$ ($96{\times}1$) --- small, sign-unstable, and all far below the gate, so we report only the binary non-collapse. A \emph{reward-shaped} regime (shaped-$\lambda0.1$, which adds an SAE-feature shaping term) shifts the delta to $\Delta H=-0.024$, $d=-0.29$ (still sub-gate): one reward-shaped run moves H3 slightly more than the accuracy-only baseline, within the spread of our probe-budget reruns, so we draw no conclusion about reward structure from it. On the committed cross-tokenizer run the largest per-feature entropy deltas are heavy-tailed and \emph{include} two of the four v0-panel top-attribution features ($\#8286$ and $\#32255$, both $\Delta=+0.11$, among the top-5 gainers, and a third, $\#9666$, tied with them at $+0.110$ and left out of the receipt's top-5 list only by tie order; Table~\ref{tab:h3_delta}); the aggregate delta is nonetheless negligible ($d=-0.005$).

\paragraph{Pretraining-crystallized directions.}
The base-stable concentration admits a pretraining-crystallized reading: \citet{moskvoretskii2026persona} find persona vectors~\citep{chen2025persona_vectors} form within $0.22\%$ of OLMo-3 pretraining and, extracted that early, still steer the post-trained instruct model, although they keep refining through pretraining. Our scale-stable top-$1\%$ concentration and the per-destination-layer single-hub slots of the 8B scan ($\#35239$@22, $\#15675$@26, $\#21438$@30, $\#12735$@32) are compatible with that reading, but the 8B graphs have no SFT or GRPO checkpoint, and identical integer indices across layers are not aligned features. Per \citet{unstable_subspaces2026} the reproducible unit is a \emph{subspace} not a literal index; the two withdrawn v0 coincidence claims are detailed below.

\paragraph{Panel provenance and measurement space.} H2 and H3 audit the real \texttt{Qwen3-1.7B} SFT/GRPO checkpoints, but \emph{indirectly}: each checkpoint generates text, and a frozen external reader --- Pythia-70m with the \texttt{pythia-70m-deduped-res-sm} residual SAE, the v0 pilot pair --- encodes the generated text (protocols in App.~\ref{app:setup}). The $32$-feature panel was preselected by a function-word-outcome screen --- features ranked by association with the next token falling in a closed set of function words, over $2{,}048$ rollouts at sequence length $64$ (positive rate $0.092$); the selection receipt, including all $32$ indices and the screen's definition, is committed at \texttt{notes/panel\_v0\_pythia70m.json} and subsequently evaluated on the v0 pilot graph in this same reader space, so the measurement is well-defined: panel indices (e.g.\ $\#32255, \#9666, \#8286, \#5158$) index the \emph{reader's} SAE dictionary, which is identical across checkpoints because the reader never changes; only the generated text differs between pre and post. These indices are therefore unrelated to the \texttt{qwen-scope} H1 hub indices ($\#30329,\ldots$, Table~\ref{tab:h1_weights}), and H2/H3 probe the checkpoints' \emph{generated prompt-plus-continuation text} through a fixed SAE lens rather than their internal activations --- auxiliary, generation-side corroboration, not an internal-activation null.

\paragraph{H2 (activation delta).} The committed cross-tokenizer receipt (\texttt{h2\_diff\_v0\_cross\_tokenizer.json}) is the shaped-$\lambda0.1$ arm ($n=16$ probes; pre $=$ SFT, post $=$ shaped-reward GRPO). On the v0 panel's top-32 features its per-feature mean-activation delta lies in $[-0.008, +0.006]$ (weighted cosine $0.9998$, Jaccard $1.0$, identical $16/16$ active counts). The four top inbound hubs of the v0 pilot graph ($\#32255, \#9666, \#8286, \#5158$ by inbound weight in \texttt{h1\_summary\_v0.json}; Table~\ref{tab:h2_delta}) all move in the \emph{negative} direction ($-0.0064, -0.0060, -0.0040, -0.0003$ respectively), consistent with this arm's reward explicitly penalizing the panel's activations; no feature is switched off. The cosine-near-1 is not a precision floor: delta SD $0.0029$ is $\sim\!5\%$ of the $0.0518$ mean magnitude (Table~\ref{tab:h2_delta}, Fig.~\ref{fig:h2_delta}). (An earlier accuracy-only baseline-GRPO H2 run is not among the committed receipts, so we report only the shaped arm here.)

\begin{table}[h]
\centering
\caption{\textbf{H2 top-8 per-feature mean-activation deltas (committed shaped-$\lambda0.1$ arm, cross-tokenizer, $n=16$ probes).} Under this reward-shaped arm three of the four v0-graph inbound hubs are among the top-8 movers, all \emph{negative} (the fourth, $\#5158$, moves $-0.0003$); the same $16$ of the $32$ panel features stay active (weighted cosine $0.9998$, Jaccard $1.0$).}
\label{tab:h2_delta}
\small
\begin{tabular}{lrrrl}
\toprule
Feature ID & Mean act.\ (pre) & Mean act.\ (post) & $\Delta$ & In hub set? \\
\midrule
14907 & $0.0248$ & $0.0168$ & $-0.0079$ & no \\
32255 & $0.4790$ & $0.4727$ & $-0.0064$ & yes \\
8232  & $0.0047$ & $0.0110$ & $+0.0063$ & no \\
9666  & $0.4526$ & $0.4466$ & $-0.0060$ & yes \\
32257 & $0.0103$ & $0.0155$ & $+0.0053$ & no \\
12234 & $0.0096$ & $0.0142$ & $+0.0046$ & no \\
27648 & $0.0869$ & $0.0826$ & $-0.0043$ & no \\
8286  & $0.3018$ & $0.2978$ & $-0.0040$ & yes \\
\bottomrule
\end{tabular}
\end{table}

\paragraph{H3 (entropy).} The probe-sensitivity calibration: injecting a known entropy change via sampling temperature ($0.5\to1.5$ gives $d=+0.18$; the extreme $0.3\to2.0$ gives $d=+0.21$) shows the metric is directionally correct; we therefore read H3 only as a binary non-collapse test rather than a calibrated effect-size instrument. A stronger positive control via a \citet{cherrl2026} injected-bias checkpoint is pre-registered. The largest per-feature entropy deltas are heavy-tailed (Table~\ref{tab:h3_delta}); two of the four v0-panel top-attribution features ($\#8286$, $\#32255$) are themselves among the top-5 gainers ($\Delta=+0.11$), and a third ($\#9666$) ties the fifth at $+0.110$, excluded from the top-5 list only by tie order, so the entropy movement is not confined to low-attribution features.

\begin{table}[h]
\centering
\caption{\textbf{H3 per-feature entropy deltas (top-5 gainers + top-5 collapsers), committed cross-tokenizer run, v0 Pythia-70m panel.} Two of the four v0-graph inbound hubs ($\#8286$, $\#32255$) are among the top-5 gainers ($\Delta=+0.11$); a third ($\#9666$) ties the fifth row at $+0.110$ (four features share this value; the receipt's tie order drops it); the aggregate delta is nonetheless negligible ($d=-0.005$, gate not passed).}
\label{tab:h3_delta}
\small
\begin{tabular}{lrrrl}
\toprule
Feature ID & Entropy (pre) & Entropy (post) & $\Delta$ & In hub set? \\
\midrule
14907 & $0.980$ & $1.151$ & $+0.171$ & no \\
22117 & $1.812$ & $1.941$ & $+0.130$ & no \\
8286  & $1.812$ & $1.922$ & $+0.110$ & yes \\
32255 & $1.812$ & $1.922$ & $+0.110$ & yes \\
2180  & $1.812$ & $1.922$ & $+0.110$ & no \\
\midrule
9522  & $0.794$ & $0.616$ & $-0.178$ & no \\
28334 & $0.354$ & $0.260$ & $-0.095$ & no \\
12234 & $0.995$ & $0.659$ & $-0.335$ & no \\
8232  & $0.747$ & $0.345$ & $-0.401$ & no \\
3201  & $1.718$ & $1.205$ & $-0.513$ & no \\
\bottomrule
\end{tabular}
\end{table}

\paragraph{Withdrawn v0 coincidence claims.} On the v0 Pythia-70m panel the four highest-attribution features also had the highest edge in-degree (a v0-pilot small-cardinality observation that does \emph{not} transfer to the qwen-scope H1 graph). We withdraw two earlier-draft claims: (1) the ``entropy-widening'' fourth dimension (the $+0.206$ hub-widening first measured on a smaller probe set is not reproduced on the committed cross-tokenizer run, where three of the four v0 top features move by $\approx+0.11$ and the fourth ($\#5158$) by $+0.06$; Table~\ref{tab:h3_delta}); (2) the ``$4/4$ pass the Steering $\sigma_{\text{KL}}\geq2$ gate'' arm and its $1/35{,}960\approx0.000028$ significance --- a corrected steering analysis, computed on $\sigma_{\text{KL}}$ (not the clamp denominator $\sigma_f$), clears $0/19$. The same-backbone 8B coincidence ($\#35239$; L4, App.~\ref{app:limits}) does not depend on any steering result and is retained.

\paragraph{Base-control visualization.}
Figure~\ref{fig:base-control} plots the base$\to$post Gini change for every post-training checkpoint (Table~\ref{tab:h1_robust}): all five light-recipe Qwen runs sit inside the $|\Delta\text{Gini}|\le0.005$ observed envelope, only heavy \texttt{gemma-2-2b-it} clears it ($+0.035$), and even that is an order of magnitude below the single-checkpoint Gini level.

\begin{figure}[h]
\centering
\includegraphics[width=1.0000\textwidth]{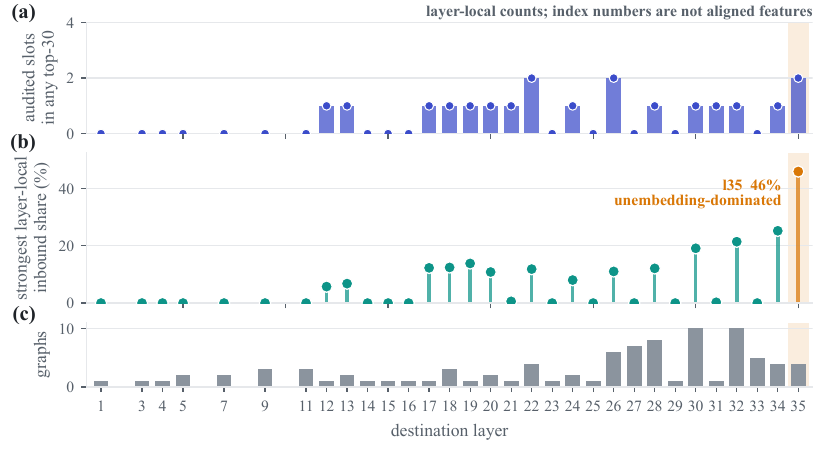}
\caption{\textbf{A layer-local diagnostic of the audited index slots.} (a) Count of the fifteen audited integer slots present in any top-30 set at each destination. (b) Their maximum inbound share within that destination. (c) Number of available source graphs. The SAE dictionaries were trained independently by layer: identical integer indices are not aligned features. This plot therefore reports local slot statistics, not cross-layer feature recurrence or preserved feature identity. Layer 35 is unembedding-dominated.}
\label{fig:backbone-recurrence}
\end{figure}

\begin{figure}[h]
\centering
\includegraphics[width=\textwidth]{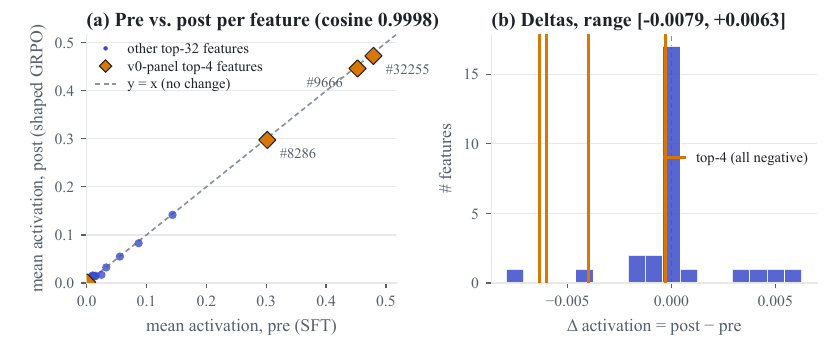}
\caption{\textbf{H2 per-feature delta diagnostic (cross-tokenizer, $n=16$ probes).} (a) Per-feature mean activation pre (SFT) vs.\ post (shaped-reward GRPO, shaped-$\lambda0.1$ arm), $32$ features. v0-panel top features (amber diamonds) sit on the diagonal with small \emph{negative} deltas, consistent with this arm's reward penalizing the panel's activations. (b) Histogram of post$-$pre deltas; the four v0-panel top features are marked. Cosine $0.9998$ is consistent with the delta magnitudes shown ($\sim\!5\%$ of pre-activation mean), not numerical precision.}
\label{fig:h2_delta}
\end{figure}

\paragraph{Layer-local hub-slot diagnostic.}
Figure~\ref{fig:backbone-recurrence} scans a $15$-hub panel across $31$ destination layers of the \texttt{Qwen3-8B} widekt graphs: $13$ of the $15$ appear in at least one graph's top-$30$ inbound set; only three ids recur at more than one destination layer ($\#28040$ at l12--l19, $\#21438$ at l30--l31, $\#58220$ at l21/l34), and because each destination layer has its own independently trained SAE, a cross-layer index match is not feature identity; the terminal-l35 cell is unembedding-dominated; these slot statistics do not establish cross-layer feature identity.

\paragraph{H2 delta diagnostic figure.}
Figure~\ref{fig:h2_delta} shows the per-feature pre- vs.\ post-GRPO mean activations (panel a) and the histogram of post$-$pre deltas (panel b) for the top-$32$ features of the v0 Pythia-70m panel, on the \emph{shaped-reward} (shaped-$\lambda0.1$) arm. The four top-attribution features (highlighted) sit on the $y\!=\!x$ diagonal with small \emph{negative} deltas --- the reward explicitly penalizes this panel's activations --- and are polarity-preserving redistributions, not on-off switches.

\section{Figure and table sources}
\label{app:sources}
Every number in a figure or table is read from a committed receipt in the code release; the files and generators are listed here rather than in the captions.
{\small\begin{longtable}{@{}p{0.13\textwidth}>{\raggedright\arraybackslash}p{0.83\textwidth}@{}}
\toprule Item & Receipts and generator \\ \midrule
\endhead
Fig.~\ref{fig:bands} & Drawn by \texttt{code/analysis/fig1\_r1\_strip.py} from \texttt{results/\{sftband\_20260926,graph\_build\_audit\_20261005\}/}. \\
Table~\ref{tab:main} & \texttt{results/\{sftband\_20260926,graph\_build\_audit\_20261005\}/}, recomputed by \texttt{code/scripts/samearch.py} (App.~\ref{app:build}). \\
Fig.~\ref{fig:base-control} & \texttt{notes/basediff\_\{qwen,gemma\}\_summary.json}, \texttt{h1\_permutation\_null\_1p7b\_l14to22.json}. \\
Table~\ref{tab:tulu} & \texttt{notes/tulu\_*.json} (v1; \texttt{tulu\_v2\_*} normalized). \\
Table~\ref{tab:randsae} & \texttt{notes/circuit\_rewiring\_randsae.json}. \\
Table~\ref{tab:archived} & Recomputed by \texttt{code/scripts/samearch.py} from \texttt{notes/}. \\
Table~\ref{tab:readouts} & \texttt{results/sftband\_20260926/sftband\_\{verdict,readouts\}.json}. \\
Fig.~\ref{fig:h1-concentration} & Source: the 47 committed Qwen3-8B layer-pair receipts (\texttt{notes/attribution\_graph\_v1\_qwen3-8b\_l*to*.json}); generator \texttt{code/analysis/fig7\_concentration\_vs\_depth.py}. \\
Table~\ref{tab:h1_scale} & Qwen-scope rows: \texttt{gini\_width\_control\_qwen\_1p7b\_vs\_8b.json}; transcoder rows: sweep job scripts, no per-run summary JSON. \\
Fig.~\ref{fig:seednull} & (a)~the base, GRPO and audited-SFT graph JSONs in \texttt{results/sftband\_20260926/run/graphs/}, read by \texttt{code/scripts/samearch.py}; (b)~the \texttt{featstats\_*}/\texttt{tulu\_featstats\_*} JSONs; generator \texttt{code/analysis/fig\_seednull\_battery.py}. \\
Table~\ref{tab:functional} & \texttt{notes/functional\_transfer\_full\_\{A,B,C\}\_k*.json}, read only by \texttt{functional\_transfer\_sweep\_read.py}. \\
Fig.~\ref{fig:staircase-data} & \texttt{notes/basediff\_\{qwen,gemma\}\_summary.json}, \texttt{tulu\_basediff\_summary.json}, \texttt{tulu\_stage\_increments.json} (v1); generator \texttt{code/analysis/fig\_staircase\_money.py}. \\
Table~\ref{tab:h2_delta} & Inbound hubs: \texttt{h1\_summary\_v0.json}. \texttt{h2\_diff\_v0\_cross\_tokenizer.json}. \\
Fig.~\ref{fig:backbone-recurrence} & \texttt{notes/fig8\_backbone\_recurrence\_data.json}; generator \texttt{code/analysis/fig8\_layer\_index\_diagnostic.py}. \\
Fig.~\ref{fig:dissociation} & \texttt{notes/bfcl\_behavior\_ci.json}; generator \texttt{code/scripts/bfcl\_behavior\_ci.py}. \\
Table~\ref{tab:h3_delta} & \texttt{h3\_entropy\_cross\_tokenizer\_v0.json}. \\
\bottomrule
\end{longtable}}

\end{document}